\RequirePackage{amsmath} 
\documentclass[a4paper,10pt]{article}

\usepackage{lmodern} 

\usepackage[margin=2.5cm]{geometry}
\usepackage[table,usenames,svgnames]{xcolor}
\definecolor{lightgray}{gray}{0.92}
\newcolumntype{a}{>{\columncolor{lightgray}}r}
\newcolumntype{b}{>{\columncolor{white}}r}
\newcolumntype{e}{>{\columncolor{lightgray}}c}
\newcolumntype{f}{>{\columncolor{white}}c}

\newcommand{\obj}[2]{$\vcenter{\hbox{\protect\includegraphics[height=\baselineskip,origin=c]{scene#1_obj#2.png}}}$}

\newcommand{\seq}[5]{\obj{#1}{#2} $\rightarrow$ \obj{#1}{#3} $\rightarrow$ \obj{#1}{#4} $\rightarrow$ \obj{#1}{#5}}

\newcommand{\meanplot}[4]{
\begin{tikzpicture}
\begin{axis}[
          xmin=0,
          xmax=100,
          ymin=-0.01,
          ymax=0.01,
          ytick={-0.5,0,...,0.5},
          ytick style={draw=none},
          every node near coord/.style={/pgf/number format/.cd,fixed,fixed zerofill,precision=1,/tikz/.cd,font=\tiny\sffamily},
          ticks=none,
          ymajorgrids=true, 
          major grid style={draw opacity=0.9},
          width=#2,
          height=#1,
          scale only axis,
          axis line style={draw opacity=0}
]
\addplot[color=blue,mark=|,line width=0.5pt,nodes near coords={\textcolor{black}{\pgfmathprintnumber[assume math mode=true]\pgfplotspointmeta\%}},point meta=explicit symbolic]
plot [error bars/.cd, x dir = both, x explicit, error bar style={line width=0.5pt}, error mark options={mark size=0}]
table [x index=0,y index=1,x error index=2,meta index=0,row sep=crcr]{%
#3 0 #4\\ 
};
\end{axis}
\end{tikzpicture}
}
\newcommand{\meanplotvalue}[2]{
    \meanplot{0.2cm}{1.0cm}{#1}{#2}
}

\usepackage{floatrow}
\usepackage{algorithm}
\usepackage{algcompatible} 
\usepackage{algpseudocode}
\usepackage{ifthen}
\usepackage{amssymb}
\usepackage{amsmath}
\usepackage{pifont}
\usepackage{graphicx}
\usepackage{subcaption}
\usepackage{stfloats} 
\usepackage{cancel}
\usepackage[tight-spacing=true,separate-uncertainty=true]{siunitx}
\numberwithin{equation}{section}
\usepackage{tikz}
\usetikzlibrary{arrows}
\usetikzlibrary{positioning}
\usetikzlibrary{shadows}
\usepackage{pgfplots}
\usepackage{pgfplotstable}
\pgfplotsset{compat=newest} 
\usepgfplotslibrary{statistics} 
\usepgfplotslibrary{fillbetween} 
\usepgfplotslibrary{groupplots}
\usepackage{environ}
\tikzset{
  treenode/.style = {shape=rectangle, rounded corners,
                     draw, align=center,
                     top color=white, bottom color=blue!20},
  root/.style     = {treenode, font=\Large, bottom color=red!30},
  env/.style      = {treenode, font=\ttfamily\normalsize},
  dummy/.style    = {circle,draw}
}

\newlength\treeheight
\makeatletter
\newsavebox{\measure@tikzpicture}
\NewEnviron{scaletikzpicturetowidth}[1]{%
  \def\tikz@width{#1}%
  \def\tikzscale{1}\begin{lrbox}{\measure@tikzpicture}%
  \BODY
  \end{lrbox}%
  \pgfmathparse{#1/\wd\measure@tikzpicture}%
  \edef\tikzscale{\pgfmathresult}%
  \BODY
}
\makeatother
\pgfdeclarelayer{bg}    
\pgfdeclarelayer{fg}    
\pgfsetlayers{bg,main,fg}  

\PassOptionsToPackage{hyphens}{url}
\usepackage[hidelinks]{hyperref}
\hypersetup{
    colorlinks=true,
    linkcolor=black,
    citecolor=black,
    filecolor=black,
    urlcolor=black,
}

\title{\LARGE \bf Self-Supervised Damage-Avoiding\\Manipulation Strategy Optimization\\via Mental Simulation}

\author{Tobias Doernbach 
	\thanks{Tobias Doernbach was formerly known as Tobias Fromm.\newline $\vcenter{\hbox{\protect\includegraphics[height=\baselineskip,origin=c]{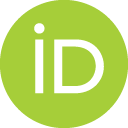}}}~$\url{https://orcid.org/0000-0001-6488-8211}\newline
	The author is with the Robotics Group, Computer Science \& Electrical Engineering, Jacobs University Bremen, Germany; \texttt{research@doernba.ch}. The research leading to the presented results has received funding from the European Union's Seventh Framework program (EU FP7 ICT-2) within the project ``Cognitive Robot for Automation of Logistics Processes'' (RobLog) under grant agreement no.\ 270350.\newline
	\textbf{Supplementary Video:}\newline\url{https://tobias.doernba.ch/research/videos/mental-simulation}\newline
	\textbf{Software:}\newline\url{https://github.com/jacobs-robotics/gazebo-mental-simulation} (see Section~\ref{self_improvement})\newline
	\textbf{Dataset:}\newline\url{https://tobias.doernba.ch/research/datasets/mental-simulation} (see Section~\ref{self_improvement})
	}
}

\begin{document}

\date{}

\maketitle

\def\titlepagesection#1#2{\par\addvspace\medskipamount{\rightskip=0pt plus1cm
\def\and{\ifhmode\unskip\nobreak\fi\ $\cdot$
}\noindent{\bfseries #1}\enspace\ignorespaces#2\par}}

\begin{abstract}
Everyday robotics are challenged to deal with autonomous product handling in applications like logistics or retail, possibly causing damage on the items during manipulation. Traditionally, most approaches try to minimize physical interaction with goods. However, this paper proposes to take into account any unintended object motion and to learn damage-minimizing manipulation strategies in a self-supervised way.
The presented approach consists of a simulation-based planning method for an optimal manipulation sequence with respect to possible damage. The planned manipulation sequences are generalized to new, unseen scenes in the same application scenario using machine learning.
This learned manipulation strategy is continuously refined in a self-supervised, simulation-in-the-loop optimization cycle during load-free times of the system, commonly known as \emph{mental simulation}.
In parallel, the generated manipulation strategies can be deployed in near-real time in an anytime fashion.
The approach is validated on an industrial container-unloading scenario and on a retail shelf-replenishment scenario.
\end{abstract}

\definecolor{burntorange}{cmyk}{0,0.52,1,0}
\def\teal{teal!60}
\def\gray{gray!30}
\def\darkgray{gray}

\tikzstyle{normalnode}=[rectangle, rounded corners, thin, inner sep=0.1cm,
                     bottom color=\gray, top color=white, drop shadow,
                     text=black, text width=1.55cm, font=\sffamily\scriptsize, align=center]
\tikzstyle{supernode}=[rectangle, rounded corners, thin,
                     bottom color=\teal, top color=white, drop shadow,
                     text=black, text width=2cm, font=\sffamily\small, align=center]
\tikzstyle{centralnode}=[centralnodefg, bottom color=\darkgray, top color=\gray, drop shadow]
\tikzstyle{centralnodefg}=[circle, thin, text=black, inner sep=0cm, text width=1.65cm,
                     font=\sffamily\scriptsize, align=center]
\tikzstyle{markednode}=[draw=burntorange, bottom color=burntorange!60, top color=white]
\tikzstyle{arrow}=[draw=\teal,ultra thick,shorten >=5pt,shorten <=5pt]
\tikzstyle{thinarrow}=[draw=\teal,thick,shorten >=5pt,shorten <=5pt]
\tikzstyle{connection}=[draw=\darkgray,dashed]
\tikzstyle{legend}=[rectangle, rounded corners, thin, minimum width=3cm, font=\sffamily\normalsize,
                    draw=burntorange, right color=burntorange!60, left color=white, rotate=90]

\def\nodedist{0.5cm}

\newcommand{\process}[2]{%
\def\markedsupernode{#1}
\def\markednode{#2}
\ifthenelse{\equal{\markedsupernode}{} \OR \equal{\markedsupernode}{all}}{%
\begin{figure*}[th]
}{
\begin{minipage}{\textwidth}
}
\resizebox{\linewidth}{!}{
\centering
\begin{tikzpicture}[auto, thick]
  \foreach \place/\name/\jumplabel/\caption in {%
        {(-3.5,3.5)/a/scene_representation/Scenario Definition \scriptsize (Section \ref{scene_representation})},
        {(0,3.5)/b/random_scenes/Training Scene Generation \scriptsize (Section \ref{random_scenes})},
        {(4,3.5)/c/planning/Mental Simulation \scriptsize (Section \ref{planning})},
        {(2,0)/d/learning/Manipulation Strategy Generation \scriptsize (Section \ref{learning})},
        {(-3.5,0)/e/self_improvement/Manipulation Strategy Anytime Deployment \scriptsize (Section \ref{self_improvement})}}
        \ifthenelse{\equal{\markedsupernode}{\name}}{\node[supernode, markednode] (\name) at \place {\hyperref[\jumplabel]{\caption}}}{\node[supernode] (\name) at \place {\hyperref[\jumplabel]{\caption}}};
    \draw [->] (a) edge[arrow] (b);
    \draw [->] (b) edge[arrow,bend left] (c);
    \draw [->] (c) edge[arrow,bend left] (d);
    \draw [->] (d) edge[arrow,bend left] (b);
    \draw [->] (d) edge[arrow] (e);

    \begin{pgfonlayer}{bg}    
        \ifthenelse{\equal{\markedsupernode}{e} \OR \equal{\markedsupernode}{all}}{\node[centralnode, markednode] (impbg) at (2,2.25) {}}{\node[centralnode] (impbg) at (2,2.25) {}};
    \end{pgfonlayer}
        \ifthenelse{\equal{\markedsupernode}{e} \OR \equal{\markedsupernode}{all}}{\draw[<-,draw=gray!20,line width=2.5pt] ([shift={(impbg)}]-75:0.5) arc[radius=0.5, start angle=-75, end angle=250]}{\draw[<-,draw=white,line width=2.5pt] ([shift={(impbg)}]-75:0.5) arc[radius=0.5, start angle=-75, end angle=250]};
    \begin{pgfonlayer}{fg}    
        \node[centralnodefg] (impfg) at (2,2.25) {\hyperref[self_improvement]{Manipulation Strategy \mbox{Optimization}}};
    \end{pgfonlayer}
    \ifthenelse{\equal{\markedsupernode}{a} \OR \equal{\markedsupernode}{all}}{%
       \ifthenelse{\equal{\markednode}{1}}{\node[normalnode, markednode, left=\nodedist of a] (a1) {\hyperref[object_representation]{Object \mbox{Representation}}}}{\node[normalnode, left=\nodedist of a] (a1) {\hyperref[object_representation]{Object \mbox{Representation}}}};
       \ifthenelse{\equal{\markednode}{2}}{\node[normalnode, markednode, below left=\nodedist of a] (a2) {\hyperref[robot_control]{Robot Control}}}{\node[normalnode, below left=\nodedist of a] (a2) {\hyperref[robot_control]{Robot Control}}};
       \path (a) edge[connection] (a1);
       \path (a) edge[connection] (a2);
    }{}
    \ifthenelse{\equal{\markedsupernode}{c} \OR \equal{\markedsupernode}{all}}{%
       \ifthenelse{\equal{\markednode}{1}}{\node[normalnode, markednode, right=\nodedist of c] (c1) {\hyperref[cost_estimation]{Manipulation Cost \mbox{Estimation}}}}{\node[normalnode, right=\nodedist of c] (c1) {\hyperref[cost_estimation]{Manipulation Cost \mbox{Estimation}}}};
       \ifthenelse{\equal{\markednode}{2}}{\node[normalnode, markednode, below right=\nodedist of c] (c2) {\hyperref[sequence_planning]{Manipulation Sequence Planning}}}{\node[normalnode, below right=\nodedist of c] (c2) {\hyperref[sequence_planning]{Manipulation Sequence Planning}}};
       \path (c) edge[connection] (c1);
       \path (c) edge[connection] (c2);
    }{}
    \ifthenelse{\equal{\markedsupernode}{d} \OR \equal{\markedsupernode}{all}}{%
       \ifthenelse{\equal{\markednode}{1}}{\node[normalnode, markednode, right=\nodedist of d] (d1) {\hyperref[lr]{Label Ranking}}}{\node[normalnode, right=\nodedist of d] (d1) {\hyperref[lr]{Label Ranking}}};
       \ifthenelse{\equal{\markednode}{2}}{\node[normalnode, markednode, below right=\nodedist of d] (d2) {\hyperref[features]{Scene Features}}}{\node[normalnode, below right=\nodedist of d] (d2) {\hyperref[features]{Scene Features}}};
       \ifthenelse{\equal{\markednode}{3}}{\node[normalnode, markednode, below=\nodedist of d] (d3) {\hyperref[modeling_preference_patterns]{Subconscious Preference Patterns}}}{\node[normalnode, below=\nodedist of d] (d3) {\hyperref[modeling_preference_patterns]{Subconscious Preference Patterns}}};
       \ifthenelse{\equal{\markednode}{4}}{\node[normalnode, markednode, below left=\nodedist of d, text width=1.8cm] (d4) {\hyperref[ranking_weighted]{Preference Weights for Label Ranking}}}{\node[normalnode, below left=\nodedist of d, text width=1.8cm] (d4) {\hyperref[ranking_weighted]{Preference Weights for Label Ranking}}};
       \path (d) edge[connection] (d1);
       \path (d) edge[connection] (d2);
       \path (d) edge[connection] (d3);
       \path (d) edge[connection] (d4);
    }{}

\end{tikzpicture}
}
\ifthenelse{\equal{\markedsupernode}{} \OR \equal{\markedsupernode}{all}}{%
\caption{Overview of the proposed method}
\floatfoot{Nodes are clickable and linked to the respective sections in this paper.}
\label{fig:components}
\end{figure*}
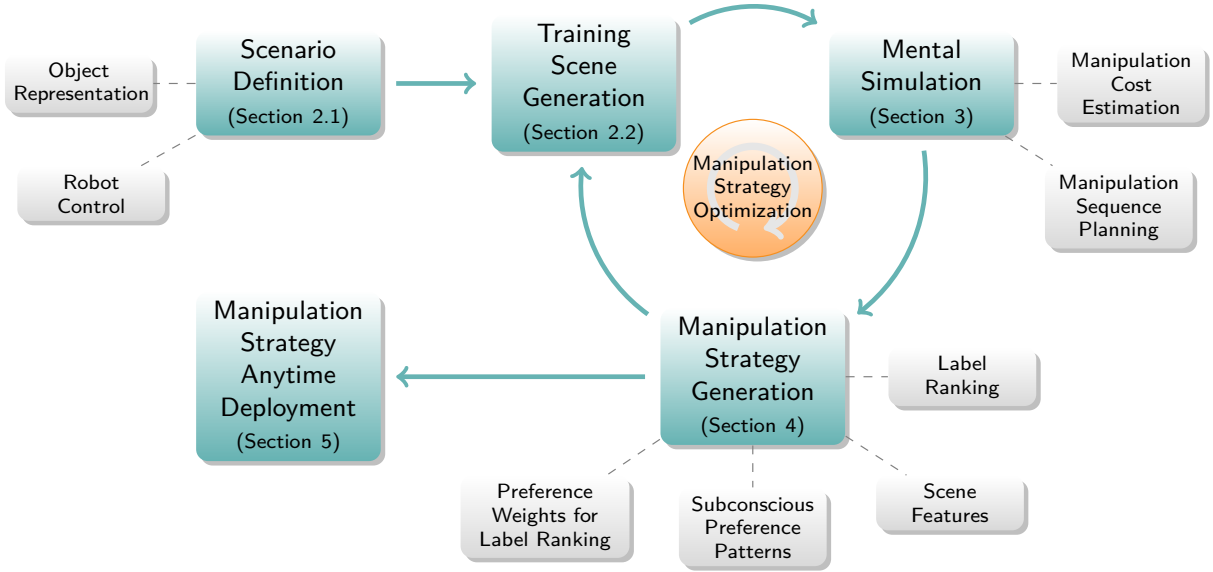
}{
\vspace{0.5cm}
\end{minipage}
}
}

\section{Introduction}

Since common perception and motion planning ap\-proa\-ches have to deal with noisy sensor readings and cluttered environments, autonomously handling fragile goods can lead to damage of items or robots.
The bottom-up way to deal with this problem is physical compliance of grasping systems or enhanced perception and manipulation algorithms (e.g.\ \cite{Stoyanov2016,Stilman2007,Katz2013,Eppner2016,Correll2018,Pavlichenko2018}, amongst others). These methods commonly rely on obstacle avoidance as to not provoke any margin-violating situation in advance. In contrast, this article proposes a top-down approach to tackle the problem in the sense that changing the environment is taken into account during the manipulation planning stage.

\process{all}{}
This work presents a \emph{manipulation strategy optimization} method that selects a sequence of objects in a given configuration to be unloaded or otherwise removed from a container, i.e.\ a supermarket shelf or a shipping container.
In the first place, we define a \emph{manipulation strategy} as a decision process which determines an optimal sequence in which to manipulate the scene objects with respect to application-specific optimization criteria.
This optimization criterion, in the case of the presented method, is defined as the damage possibly caused to other objects which may be touched and shifted during the manipulation process.
The generated strategies can be optimized autonomously by the robot during load-free times. This is achieved by generating and processing training data for learning preferences and physical constraints between the objects in a self-supervised way. A simulation-in-the-loop setup is exploited which utilizes a physics simulation to generate arbitrary amounts of training data, hence no user interaction is needed throughout the whole process as opposed to classical supervised learning approaches.

\subsection{Related work}
The proposed method relies on \emph{mental simulation} of the movements of all objects present in the scene during manipulation, a term which originates from a psychology context with a first mention in \cite{Kahneman1981}.
In general, reasoning and drawing conclusions from a simulation process has been investigated in the literature under different terms, but altogether considering the dynamics of a scene in a human-like way.
Battaglia et al.\ \cite{Battaglia2013} introduced the term \emph{intuitive physics engine} which describes the human anticipation capabilities from a cognition-scientific viewpoint. They find the quantitative evaluations of human reasoning capabilities to be surprisingly similar to the results obtained from physics simulations.
Other well-known terminology is \emph{temporal projection} \cite{Kunze2017}, \emph{physics-based reasoning} \cite{Akbari2015} and \emph{physical reasoning} \cite{Weitnauer2010} which deal with predicting real-world behavior using knowledge inferred from physics simulation.
In terms of repetitively performing actions which improve environment manipulation strategies, \emph{robotic playing} \cite{Hangl2016} is another related bio-inspired technique which compares trial-and-error behavior while accumulating environment knowledge with a children's way of exploring the world.
In the case of the proposed method, we can add the term \emph{dreaming robot} to this list of synonyms for mental simulation because the manipulation strategy optimization happens during load-free times, for instance at night when the robot is otherwise unused.
Additionally, since the manipulation strategies generated inside this approach can be transferred to a real-world scenario, \emph{transfer learning} is another applicable high-level term.

In general, simulation-in-the-loop architectures have not been studied extensively yet in the robotics literature, but have recently been on the rise for problems in the physical reasoning domain, with most authors using the term \emph{mental simulation} like in this article.
Bozcuoglu and Beetz~\cite{Bozcuoglu2017}, for instance, recently proposed a generic simulation setup for knowledge generation and reasoning through mental simulation. However, they do not yet close the loop of propagating the gained knowledge into an optimization cycle. Haidu and Beetz~\cite{Haidu2016} utilize this setup for recognizing and interpreting actions from simulation in order to collect sufficient knowledge about the task for replaying it on a real robot.
On the other hand, Levine et al.~\cite{Levine2016} use up to 14 real manipulators in parallel to collect data for grasp learning. They use visual features combined with a deep learning technique in a parallelized way, making use of massive hardware. This is similar to the proposed approach in a way that parallel randomized experiments are performed in order to generate training data, but different in a way that the presented method does not rely on hardware other than computing capacity and tries to make use of idle times as far as possible.

\subsection{Method overview}

Fig.~\ref{fig:components} shows an overview of the proposed method. In order to obtain and optimize manipulation strategies, the scene dynamics are tracked by a physics simulator. The simulated environment is synchronized with the perceived scene which includes the robot, the environment and a number of objects to manipulate, defined domain-specificly depending on the application (Fig.~\ref{fig:components} top left). In order to allow for generalization of these mental simulation capabilities, training scenes are generated automatically (top center) which, after determining the respective optimal manipulation sequence in terms of minimizing undesired object motion \cite{Fromm2016}, serve as training examples for a machine learning process (top right). Once a number of training samples has been acquired, a classifier is built and updated iteratively on new incoming samples, yielding an updated manipulation strategy (bottom right).
Subsequently, the resulting classifier can, over the course of many simulated manipulation procedures, be deployed to predict manipulation sequences with respect to the learned strategy from any new scene (bottom left).

The possibilities are manifold: Firstly, the constraints on the motion planning search space are relaxed in comparison to approaches where other objects are considered as obstacles. As in previous work \cite{Stoyanov2016} which uses a conventional motion planning method without sophisticated task-level planning, some scene configurations do not allow for collision-free manipulation at all. The presented method, however, works one level higher than motion planning and allows for generating manipulation strategies which facilitate the motion planning itself. Eventually, this allows the planner to successfully determine a viable manipulation sequence in most cases.

Secondly, because the necessary training data can be generated and consumed in a self-supervised manner by running simulations of manipulation actions, the resulting strategy can be optimized during load-free times. Parallelizing the simulation process makes the actual manipulation very efficient because the strategy improves quickly and can be applied instantly, without the need for any more simulation or other processing.
Additionally, since strategy optimization happens in the background without dependencies on the real scene, the currently active strategy can be deployed \emph{anytime} in \emph{near-real time} without the need to synchronize the learning process with physical robot behavior.

The main motivation of the presented approach is to \emph{avoid damage to the possibly heavy or fragile goods as well as the robot itself} which may occur if an object is shifted or dropped unintendedly.
However, detecting actual physical damage on handled goods has to be regarded as an own field of research related to object recognition and classification. In the scope of this work, we hence define damage avoidance as an implicit, proactive way of minimizing unintended motion of objects during the manipulation process.

The remainder of this introductory section will describe the main contributions as well as exemplary applications of the presented method which will guide the reader through the paper. Section~\ref{prerequisites} gives a brief overview about some prerequisites necessary for working with and deploying the method. Section~\ref{planning} explains how the mental simulation creates manipulation sequences which then, in Section~\ref{learning}, are compiled into manipulation strategies. Section~\ref{self_improvement} wraps up the overall process of self-supervised manipulation strategy optimization. Section~\ref{evaluation} presents a thorough evalution of the proposed approach which is concluded in Section~\ref{conclusions} with a discussion and outlook. For orientation throughout the paper, the reader is additionally referred to Fig.~\ref{fig:components} which allows for easy navigation between the different parts of the method.

\subsection{Contributions}
\begin{enumerate}
\item In any case, although the presented method aims on optimizing autonomous robot behavior using the measure of anticipated dynamics, \emph{no imposition of any explicit logical or spatial dependencies between objects} is its first main contribution. In addition, no priors are included about the type and size of objects, degrees of freedom of the robot, type of manipulator or other application domain-specific parameters. This means that the method is domain-independent and can be deployed on new application scenarios within minimal time, given that a working simulation of the manipulation procedures to be sequenced has been established already (see Section~\ref{prerequisites}).\label{contribution1}

\item The second contribution of this work is the \emph{integration of simulation into the processing loop} which is essential for the presented self-supervised mental simulation approach. Another example where simulation in the loop may play an important role is system integration which this way can be conducted as a continuous process. Hence, parts of the system can be tested individually using simulated components and gradually replaced by their real-world counterparts. Such a continuous system integration \cite{Fromm2017} technique uses simulation in the same way, embedded into a closed loop and as a full-featured component which seamlessly integrates into the processing pipeline.\label{contribution2}

\item Finally, the third main contribution of the proposed approach is the \emph{generation of manipulation strategies} from the manipulation sequences planned within the simulation processing loop.\label{contribution3}
\end{enumerate}

\subsection{Application scenarios}
\begin{figure}[tbp]
  \centering
  \begin{subfigure}[b]{0.48\textwidth}
    \includegraphics[width=\linewidth]{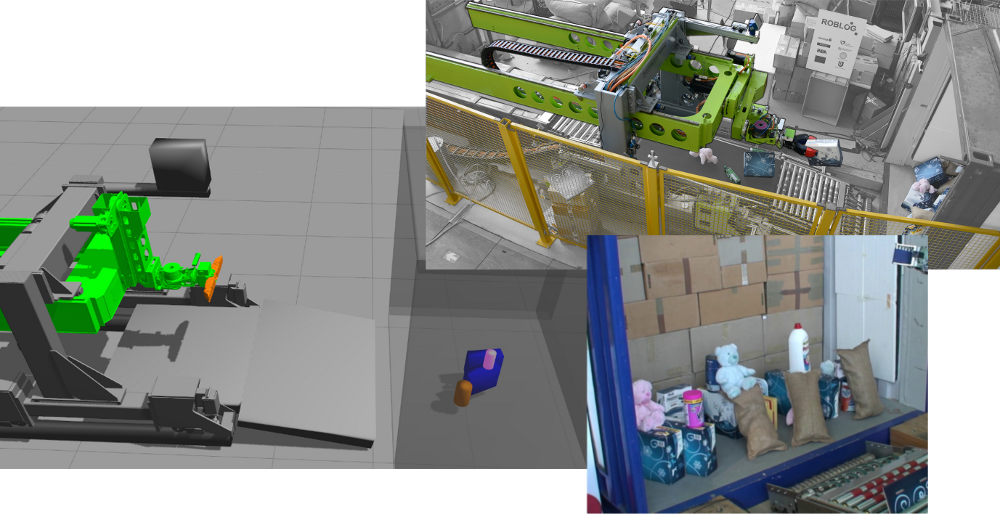} 
    \caption{Logistics (container unloading) \cite{Stoyanov2016}}
    \label{fig:intro_scenario_roblog}
  \end{subfigure}
  \hfill
  \begin{subfigure}[b]{0.48\textwidth}
   \includegraphics[width=\linewidth]{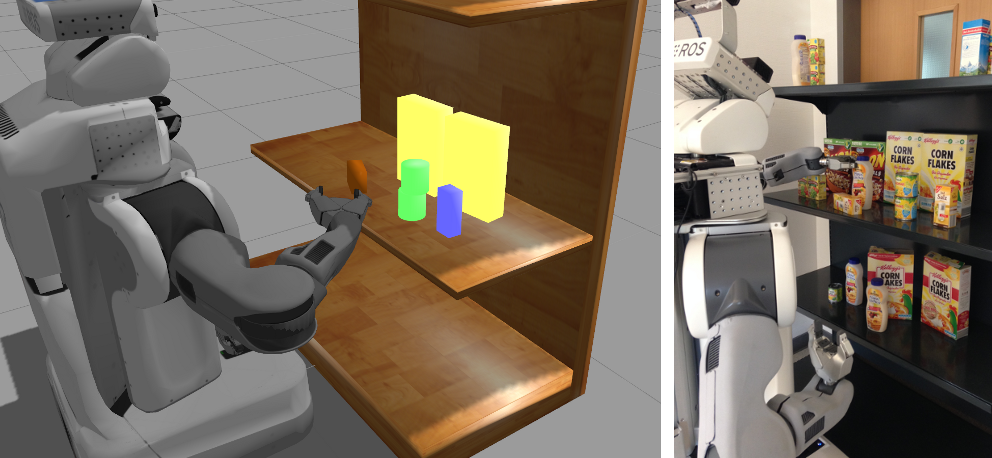} 
   \caption{Supermarket (shelf replenishment) \cite{Winkler2016}}
   \label{fig:intro_scenario_shopping}
  \end{subfigure}
  \caption{Typical application scenarios for autonomous manipulation}
  \label{fig:intro_scenarios}
\end{figure}

All of the different terminology mentioned so far share the common ground of mentally simulating robot interaction in dynamic environments, but have been applied to a multitude of different domains. In order to demonstrate the generality of the approach which is not limited to a particular domain, two different everyday scenarios are used (see Figure~\ref{fig:intro_scenarios}): \emph{logistics}, with autonomous container unloading in the EU project ``Cognitive Robot for Automation of Logistics Processes'' (RobLog)\footnote{\url{http://roblog.eu}} \cite{Stoyanov2016}, and in a \emph{supermarket} environment for autonomous shelf replenishment \cite{Winkler2016}.

The first of these scenarios typically contains a number of bulky, heavy, fragile or otherwise damage-prone goods. Especially in terms of relieving human workers from health-endangering labor, robotic applications have been flourishing in this domain for several years, as in previous work on shipping container unloading.

Secondly, the presented method can be applied as well on typical service robots in retail and domestic environments, interleaved with logistics in terms of hardware and software transitioning from one field to the other, but still facing different challenges. As a concrete scenario, a supermarket featuring a PR2 robot is used as a second application example.

\section{Prerequisites}\label{prerequisites}
First of all, before explaining the presented method in detail, this section gives a short overview about certain prerequisites that have to be considered as given when deploying the method. This includes the initial definition of the scene with respect to the particular usage scenario as well as how to auto-generate a large number of scenes used within the manipulation strategy optimization cycle.

\subsection{Scenario definition}\label{scene_representation}
As possible application scenarios are different, the degree of abstraction with respect to simulation of objects and robots needs to be adapted to the desired use and available development and computing capabilities. Modeling robot behavior and object properties in detail in a simulation engine may not be efficient and even redundant.

The presented method was designed to be used in integrated scenarios that use object recognition systems along with a physics simulation on the one hand and a real-world robot on the other hand which executes the planned manipulation strategies.
However, the focus of this method is on manipulation strategy planning and not the enhancement of perception systems or motion planning/execution. Hence, noise and other inaccuracies within the perception and execution pipeline will not be addressed in this publication. Instead, this subsection briefly explains the approach taken for representing objects and the robot within our mental simulation.

\subsubsection{Object representation}\label{object_representation}

\begin{figure*}[t!]
    \centering
    
    \def\objectsheight{1.25cm}
    
    \begin{subfigure}{0.48\linewidth}
        \centering
        \setlength{\tabcolsep}{.4em} 
        \begin{tabular}{ccrr}
            \multicolumn{2}{c}{\includegraphics[height=\objectsheight]{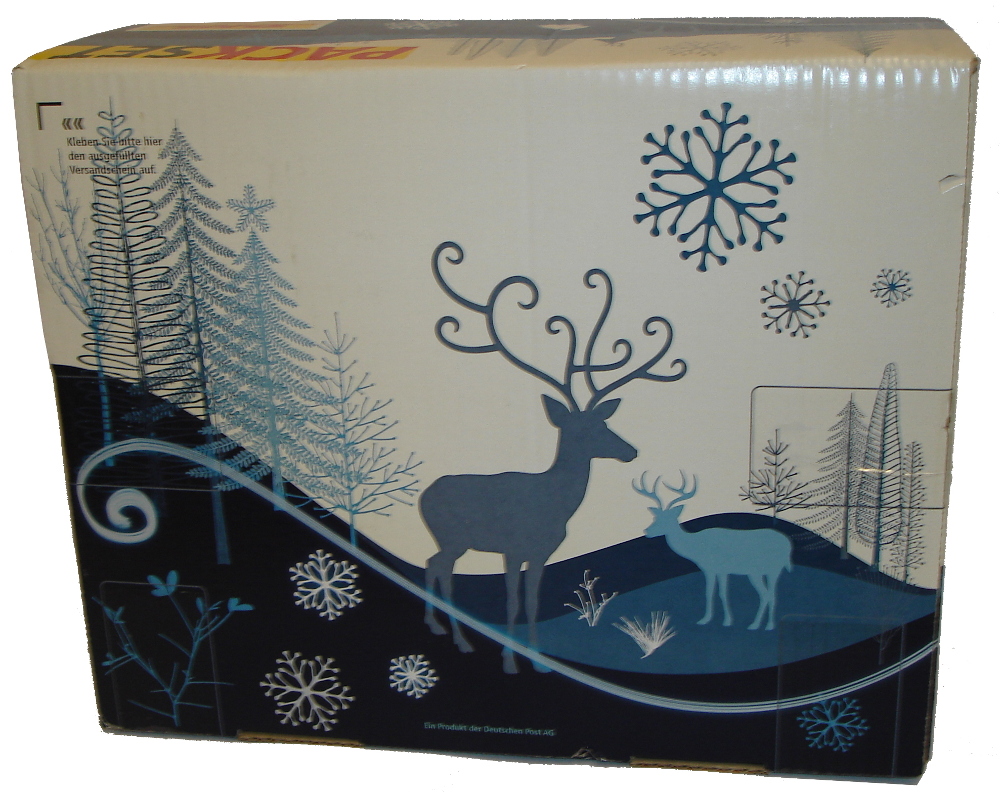}}&
            \includegraphics[height=\objectsheight]{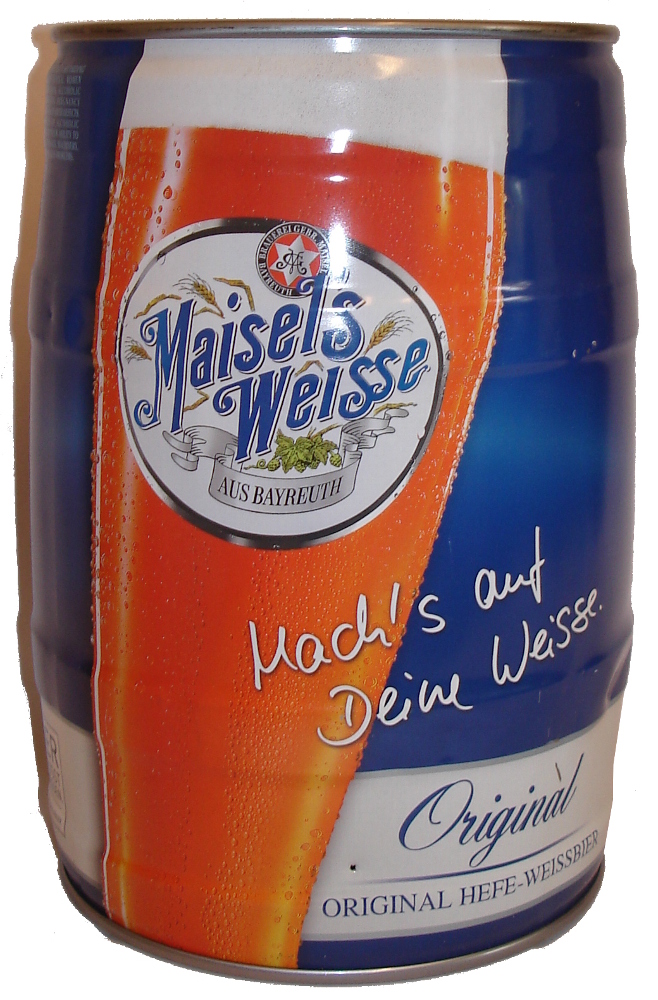}&
            \includegraphics[height=\objectsheight]{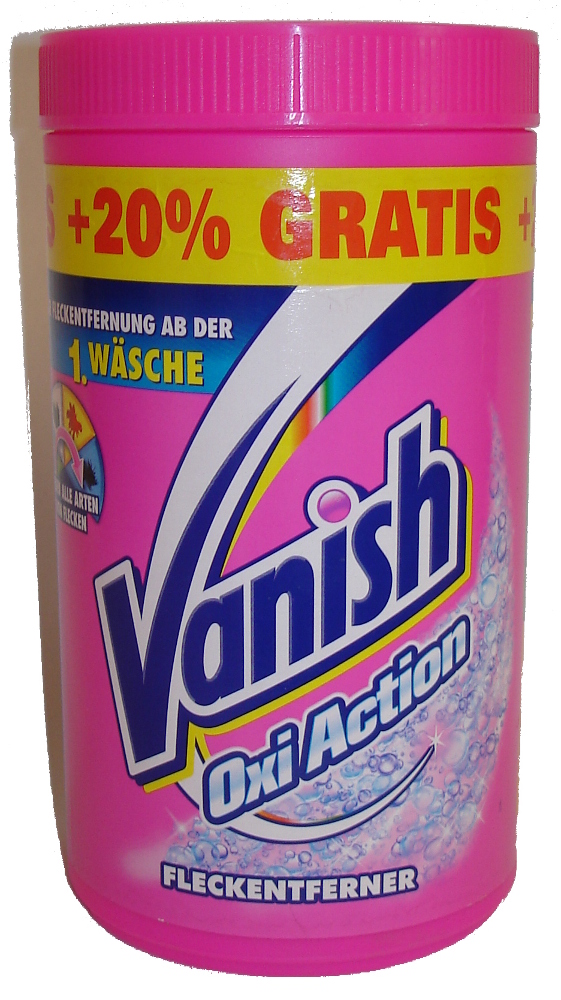}\\
            
            \includegraphics[height=\objectsheight]{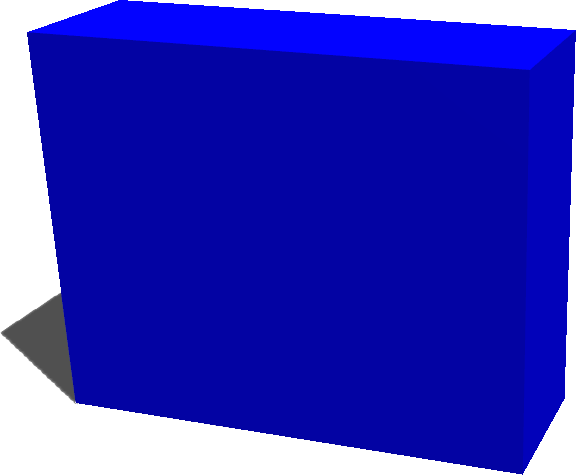}&
            \includegraphics[height=\objectsheight]{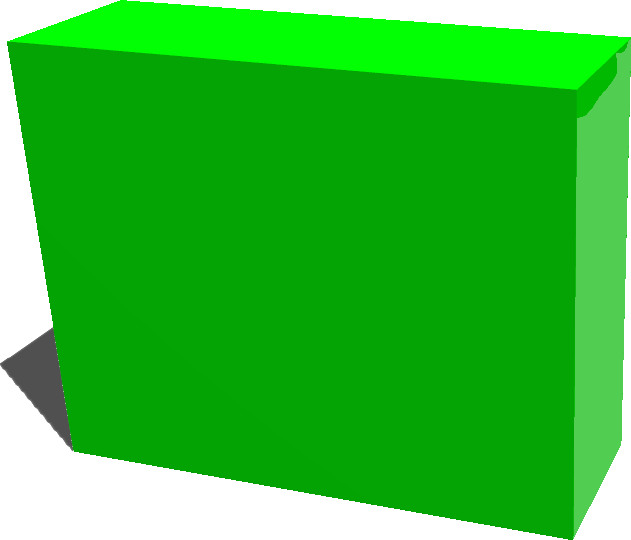}&
            \includegraphics[height=\objectsheight]{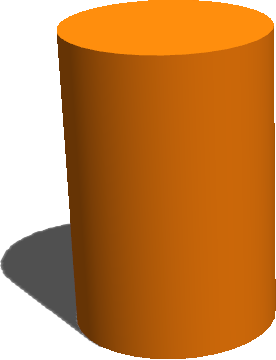}&
            \includegraphics[height=\objectsheight]{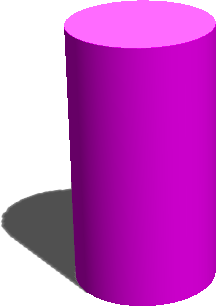}\\
        \end{tabular}
        \caption{Logistics scenario}\label{fig:simulation_objects:logistics}
    \end{subfigure}
    \begin{subfigure}{0.48\linewidth}
        \centering
        \setlength{\tabcolsep}{.4em} 
        \begin{tabular}{ccrr}
            \multicolumn{2}{c}{\includegraphics[height=\objectsheight]{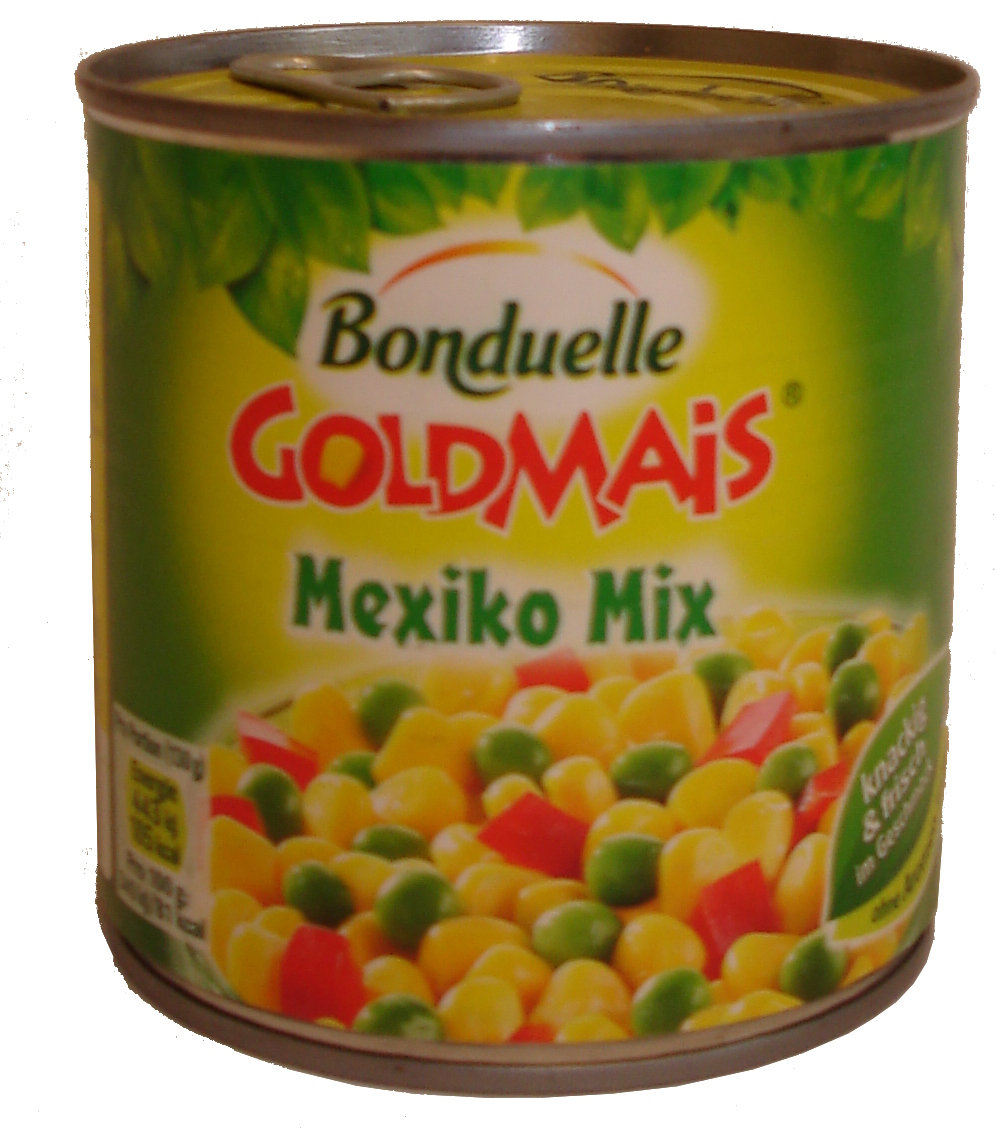}}&
            \includegraphics[height=\objectsheight]{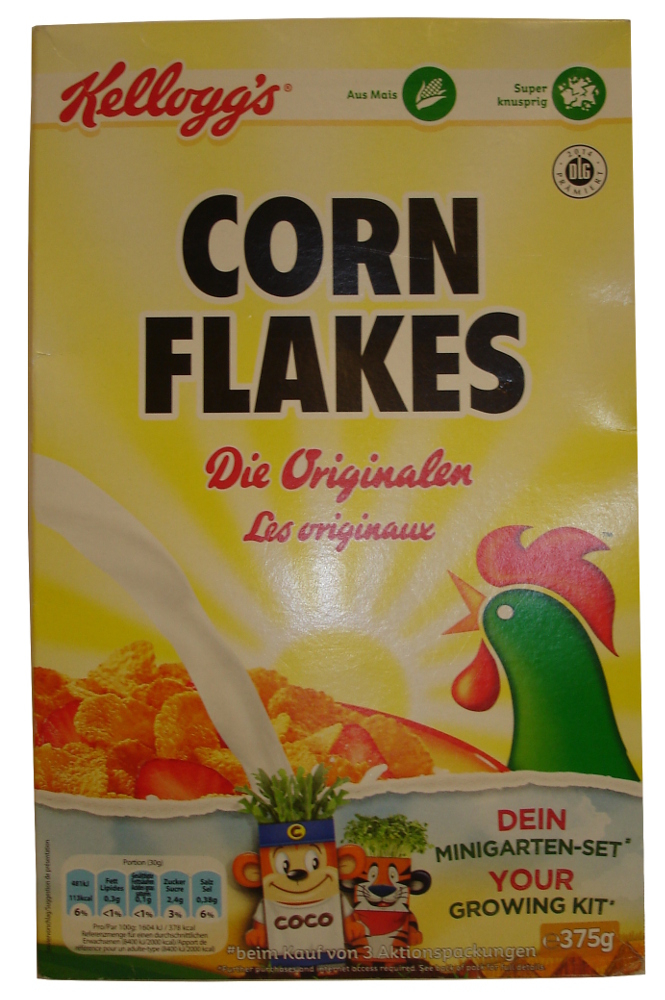}&
            \includegraphics[height=\objectsheight]{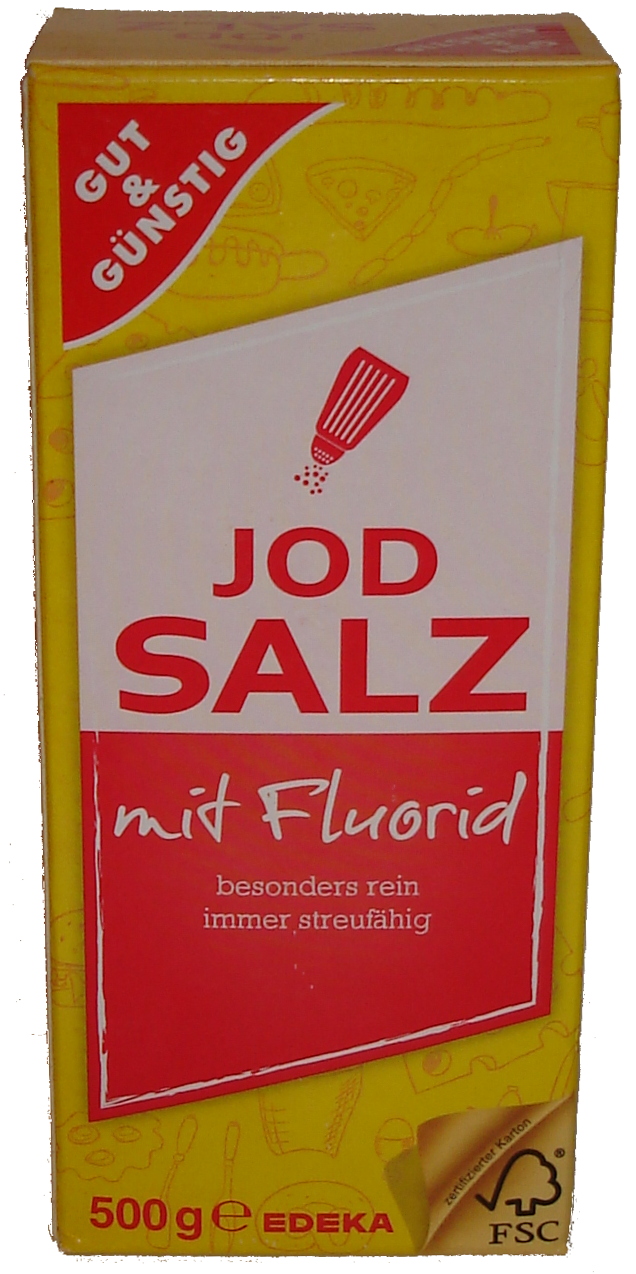}\\
            
            \includegraphics[height=\objectsheight]{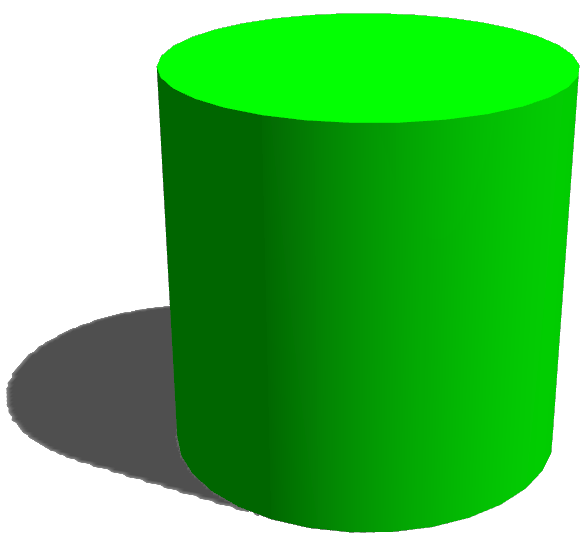}&
            \includegraphics[height=\objectsheight]{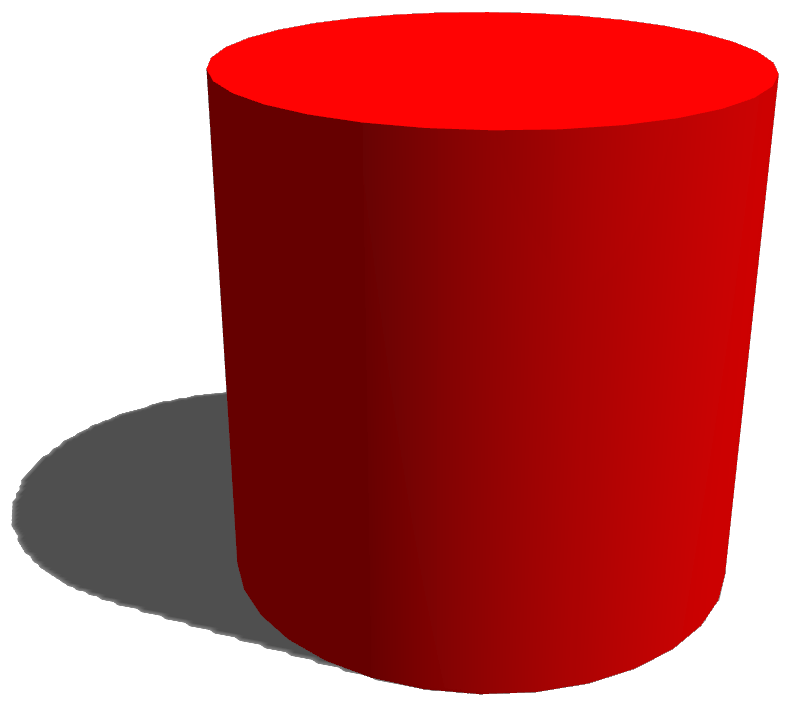}&
            \includegraphics[height=\objectsheight]{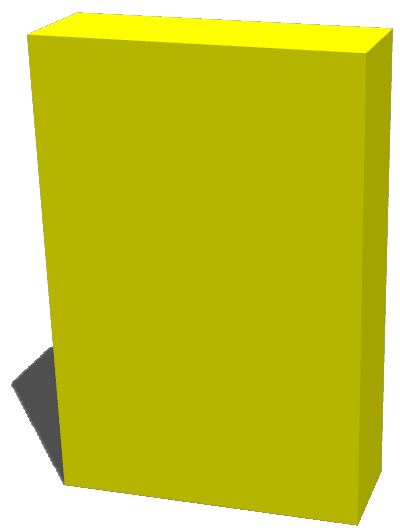}&
            \includegraphics[height=\objectsheight]{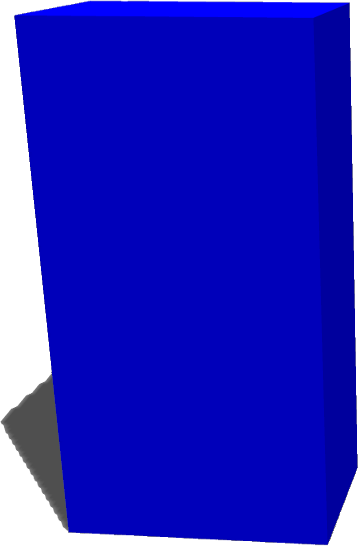}\\
        \end{tabular}
        \caption{Supermarket scenario}\label{fig:simulation_objects:supermarket}
    \end{subfigure}
    
    \caption{Objects and their simulation representations (not drawn to scale)}
    \label{fig:simulation_objects}
\end{figure*}

As for representing the scene objects in simulation, it is generally desired to obtain detailed models of them in order to achieve realistic dynamic behaviors. 
However, detailed object models generated from a real object's point cloud introduce the inherent disadvantage of possibly generating instable initial scenes in the physics simulation due to modelling inaccuracies. This may lead to some objects shifting immediately after spawning them in the physics engine.
This has to be avoided for mentally simulating manipulation actions since initially stable scenes in the physics simulation represent reality where, during and after object recognition, objects most likely are not transposed either if not physically influenced. In order to preserve stable initial configurations, the proposed method therefore uses abstract object representations as shown in Fig.~\ref{fig:simulation_objects} in order to avoid modeling errors.

\begin{figure*}[b!h]
	\centering
	\includegraphics[width=\linewidth]{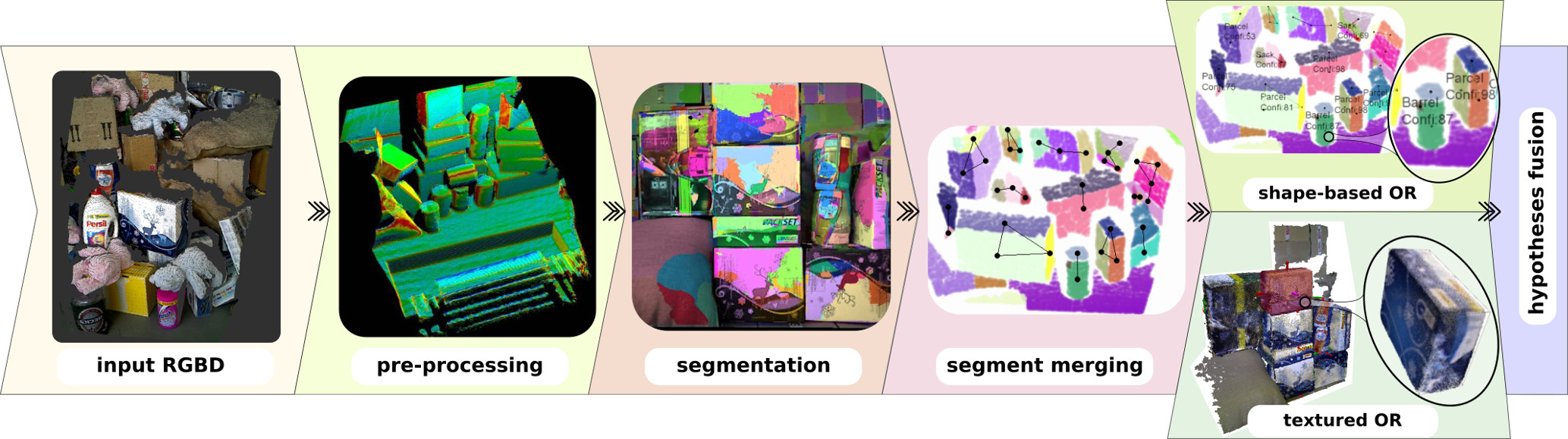}
	\caption{Perception pipeline \cite{Stoyanov2016}, consisting of the respective processing steps from data acquisition to the fusion of hypotheses from a texture-based \cite{Vaskevicius2012} and a shape-based recognition module \cite{Mueller2014}}
	\label{fig:pipeline}
\end{figure*}

During productive use in a concrete application scenario, the respective scene configuration needs to be established in the physics simulation.
This requires to recognize and localize objects occuring in the scene; for the given use cases the perception system originally developed for the logistics scenario \cite{Stoyanov2016,Vaskevicius2014} is utilized. Therein, different segmentation and filtering steps are combined into a pipeline which feeds the pre-processed sensor data into several object recognition modules. Amongst these are a graph-based shape model object recognition module \cite{Mueller2014} and a feature-based textured object recognition module \cite{Vaskevicius2012}, as illustrated in Fig.~\ref{fig:pipeline}.

In this exemplary perception pipeline, as a first step, RGBD data is acquired and immediately preprocessed for noise reduction. Next, segments are generated that are homogeneous with respect to geometric and/or color-based criteria in order to provide a low-level description of the scene objects. These segments are then merged according to some application-dependent heuristics such as convexity. From these segments, the shape-based and texture-based object recognition modules identify the object candidates. Their hypotheses are then fused heuristically as a final step to broadcast the canonical object configuration to be projected into the simulation engine.

\subsubsection{Robot control}\label{robot_control}
Motion execution, on the simulated as well as the real-world robot, requires grasp planning taking into account the perceived objects.
Grasp planning itself depends on the scenario for the reason of differing sizes and degrees of freedom of the grippers, also the dimensions, weight and texture of the objects has influence on the grasping policy. In the presented work, the grasping step is aimed to be kept as generic as possible by placing grasping configurations in a predefined distance around the principal axes of known object models like the ones in Fig.~\ref{fig:simulation_objects}. This means that, whatever object is added to the scenario, respective grasping configurations can be generated with identical properties like the existing ones of other objects. Hence the method imposes minimal constraints on the grasping process and is ready to be reused in many possible settings where, on demand, more sophisticated grasping policies can be developed.

For physics simulation, the Gazebo simulator\footnote{\url{http://gazebosim.org}}~\cite{Koenig2004} is used, including kinematics and dynamics of the robot along with simulated controllers. The container unloading robot of the logistics scenario (Fig.~\ref{fig:intro_scenario_roblog}) uses a custom set of low-level and high-level controllers whereas in the supermarket scenario (Fig.~\ref{fig:intro_scenario_shopping}), the simulation model shipped by default with the PR2 robot is used. A standard motion planner like the ones integrated in the Open Motion Planning Library\footnote{\url{http://ompl.kavrakilab.org/}}~\cite{Sucan2012} eventually allows for planning approach, grasp and retract motions.

Using the defined scenarios, the physics simulation can now be utilized to perform plan manipulation sequences via mental simulation. This, together with a methodology how to take care of possible damage, will be explained in the following section.

\subsection{Training scene generation}\label{random_scenes}
One important part of the proposed method includes the robot using load-free times to optimize its manipulation strategies. This includes re-training the respective machine learning algorithms, but also the generation of additional training data. In order to be ready for unloading many possible combinations of objects in different spatial relationships, we generate training scenes with configurations where objects are likely to collide with or obstruct other objects.
This happens by spawning the objects in clusters defined by their $x,y,z$ size in the predefined workspace (e.g.\ inside a container, on a specific shelf level). Random cluster centroids are then drawn from the workspace volume around which the objects are located with random 6-D poses.

\begin{figure*}[t!b]
	\centering
	\def\randomscenesheight{0.1\linewidth}
	\begin{subfigure}{\linewidth}
		\centering
		\includegraphics[height=\randomscenesheight]{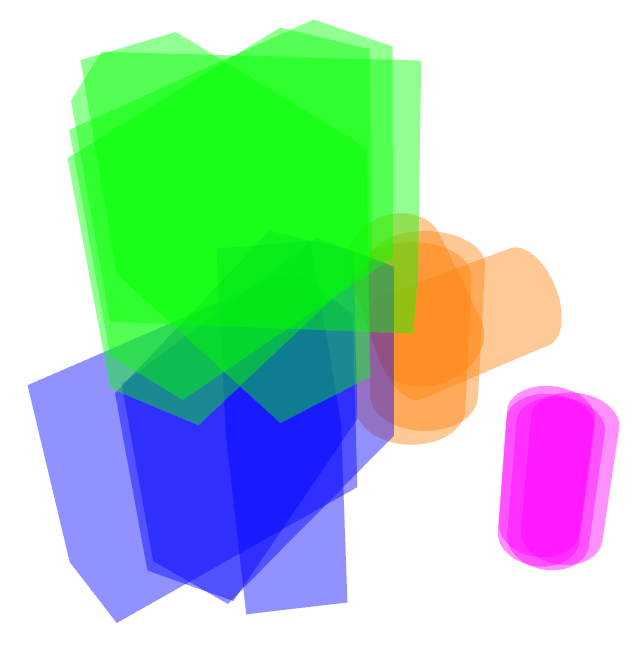}
		\includegraphics[height=\randomscenesheight]{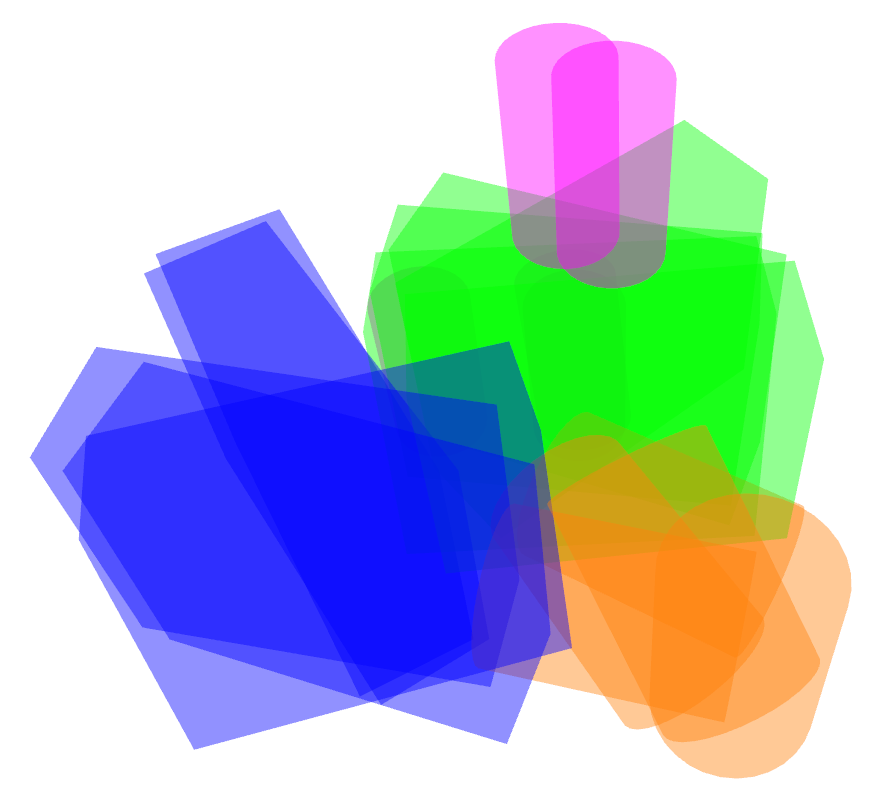}
		\includegraphics[height=\randomscenesheight]{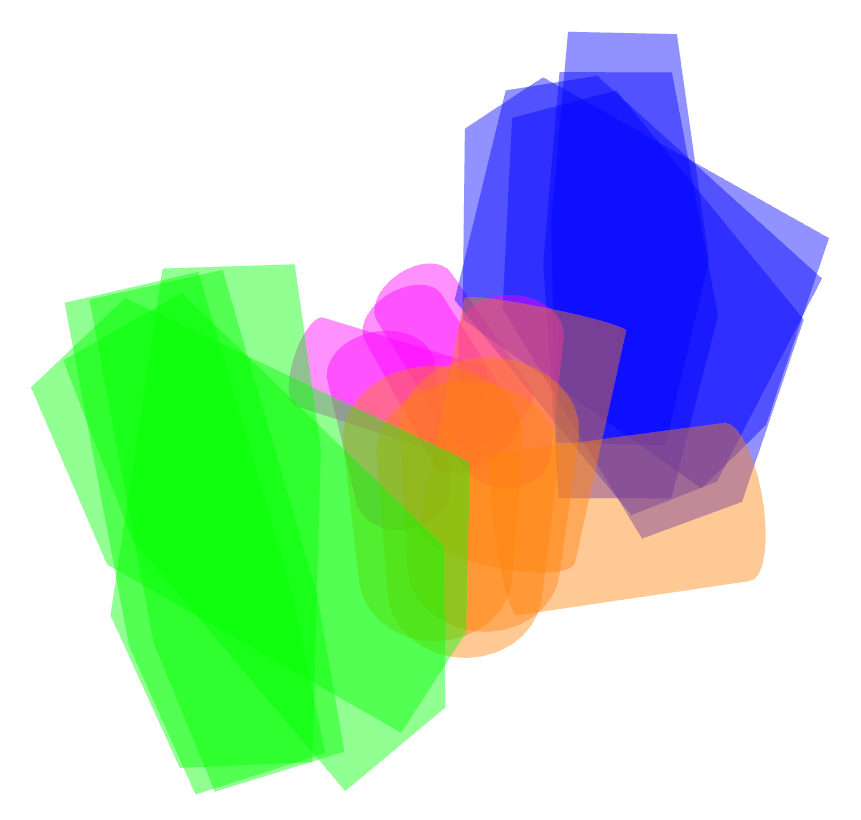}
		\includegraphics[height=\randomscenesheight]{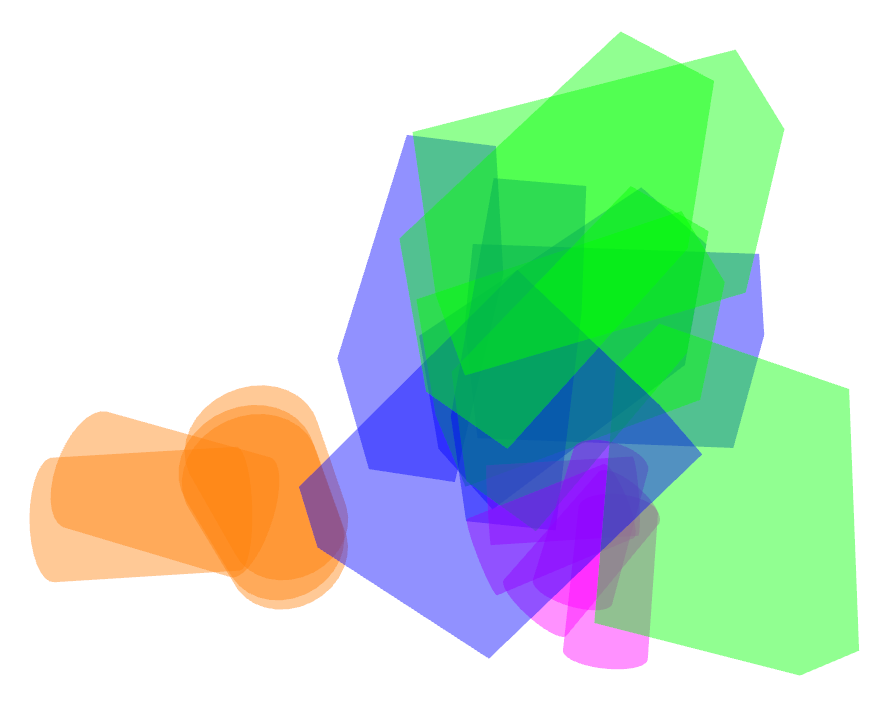}
		\includegraphics[height=\randomscenesheight]{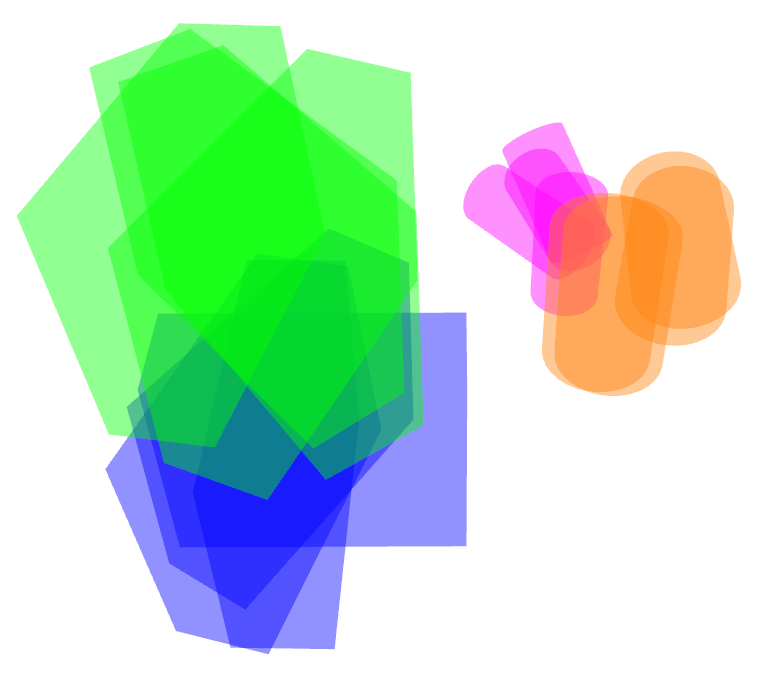}
		\includegraphics[height=\randomscenesheight]{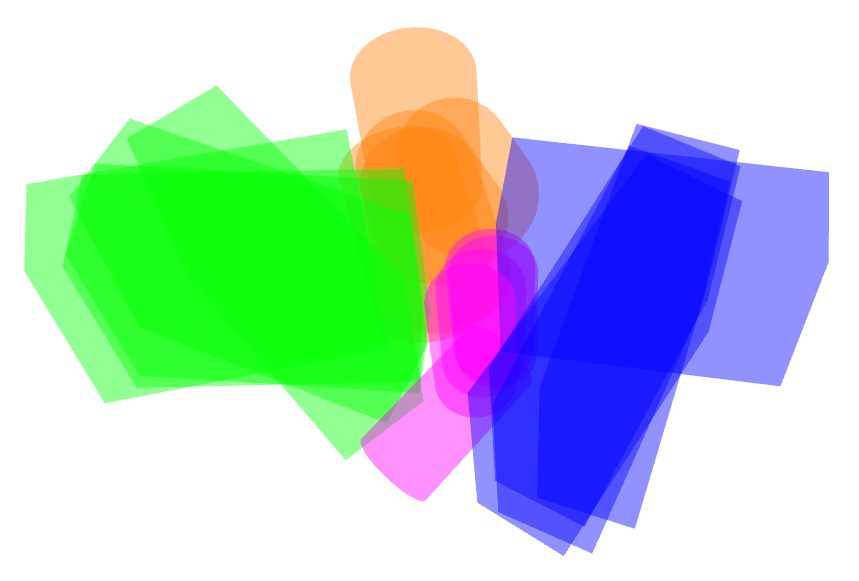}
		\caption{Logistics scenario}
	\end{subfigure}
	\hfill
	\begin{subfigure}{\linewidth}
		\centering
		\includegraphics[height=\randomscenesheight]{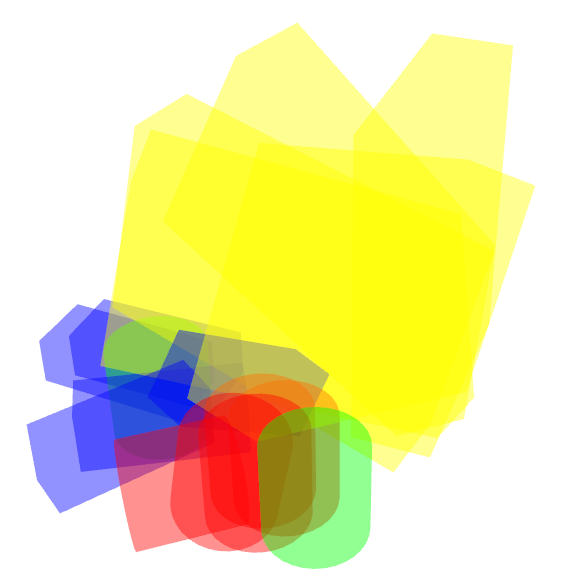}
		\includegraphics[height=\randomscenesheight]{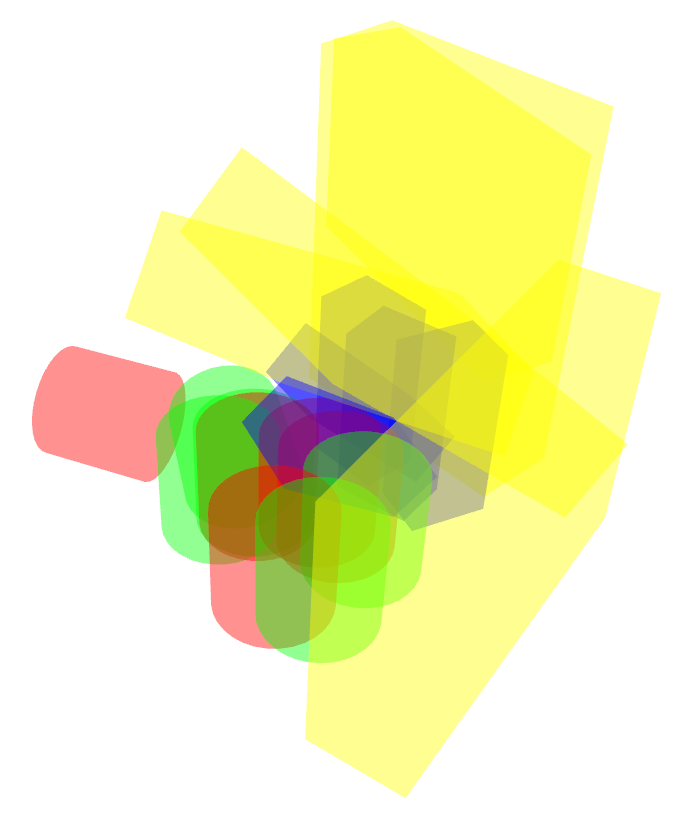}
		\includegraphics[height=\randomscenesheight]{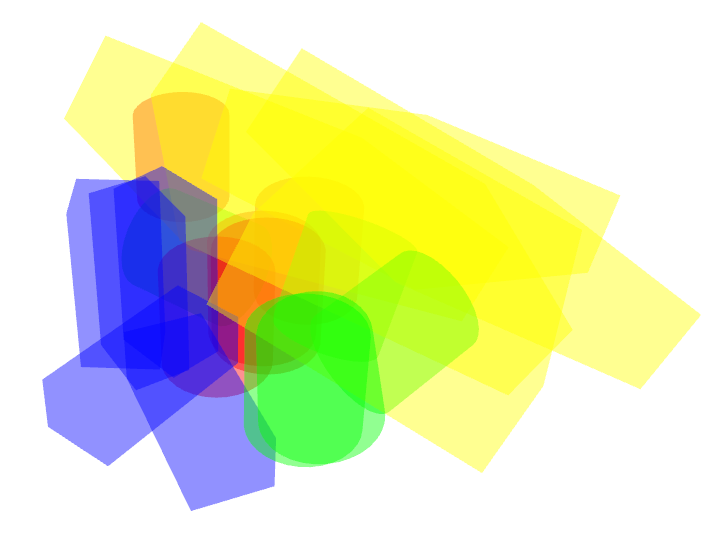}
		\includegraphics[height=\randomscenesheight]{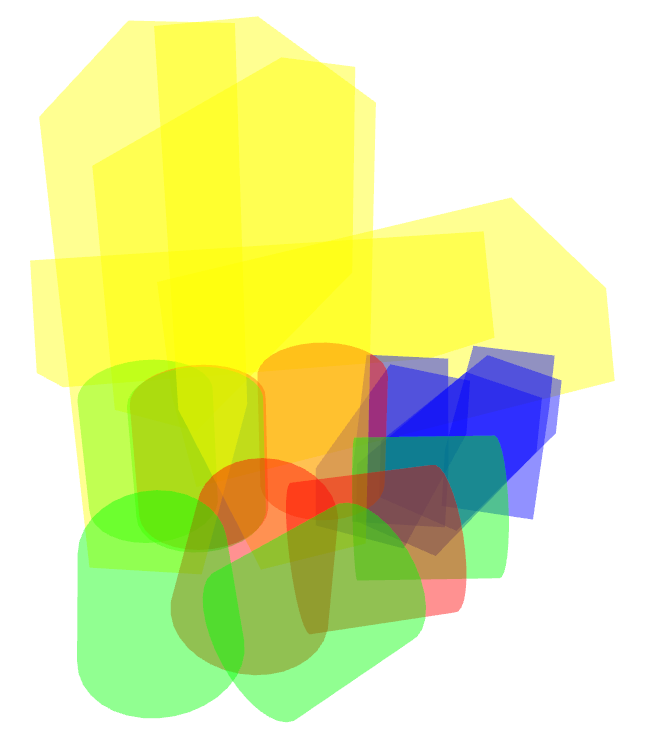}
		\includegraphics[height=\randomscenesheight]{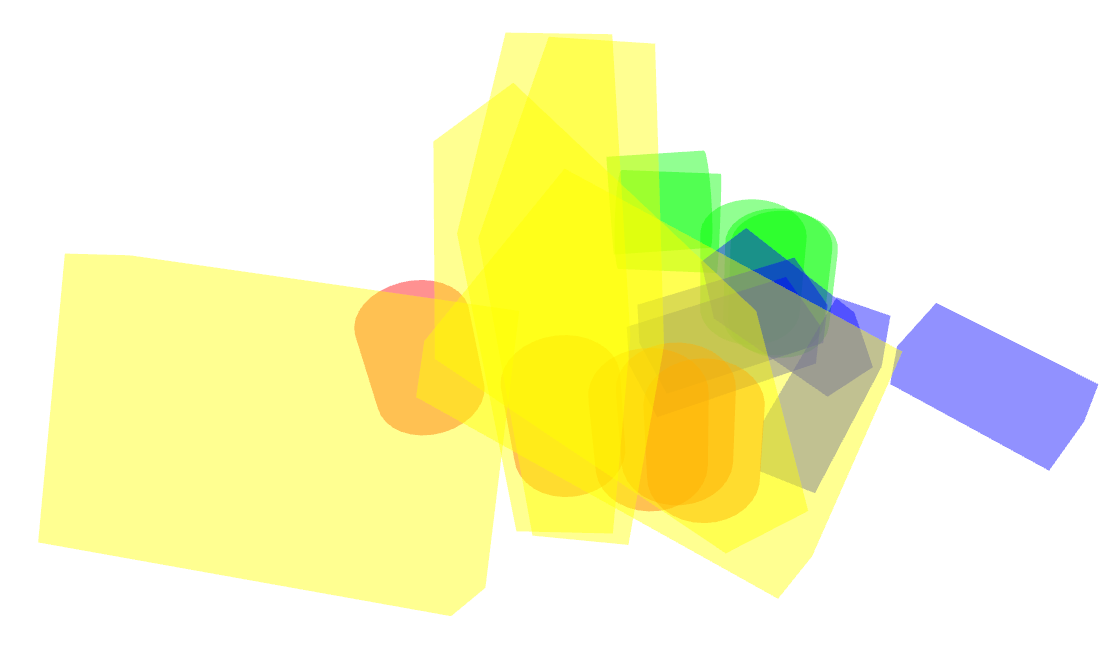}
		\includegraphics[height=\randomscenesheight]{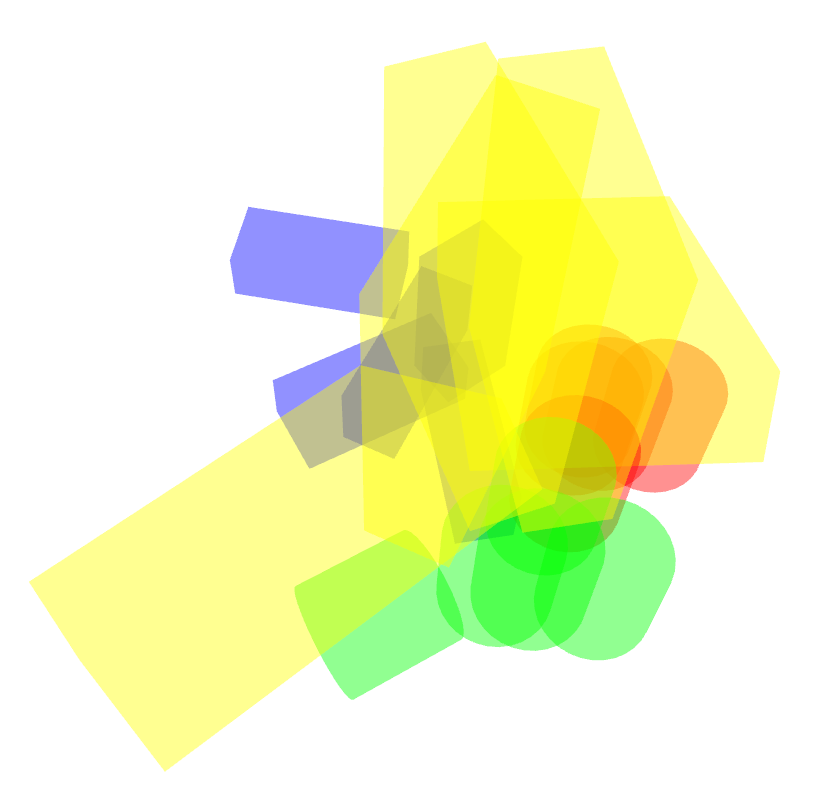}
		\caption{Supermarket scenario}
	\end{subfigure}
	\caption{Training scene examples}
	\floatfoot{Five superimposed collision-free variations of each scene with stochastic object pose noise.}
	\label{fig:random_scenes}
\end{figure*}

For refining the object poses and removing any interpenetrations that may have occured, the presented method uses \textsc{Promts}\footnote{\url{https://github.com/Rasoul77/promts}} \cite{Mojtahedzadeh2015a} which translates any given object configuration in space to a collision-free configuration.
Battaglia et al.\ \cite{Battaglia2013} use the same principle that helps generating scenes which incorporate a certain amount of noise, but still are conceivable as per human intuition and the physics simulation. 

Figure~\ref{fig:random_scenes} shows some examples for both of the used scenarios. In fact, we create several variations of each scene with stochastic noise added to the object poses which are superimposed in the examples. Whenever new training scenes have been generated, they are provided as input for the training data extraction process as described in the next section.

\section{Mental simulation: manipulation sequence planning}\label{planning}
Before advancing to the self-supervised acquisition of manipulation strategies, this section explains the underlying planning method which has been already presented and evaluated in recent work \cite{Fromm2016}.
It validates and selects the best sequence to remove or unload a number of objects from a scene with respect to possible damage. This happens entirely as a mental simulation without physical robot interaction. The result of such a planning action is taken as a training sample for the manipulation strategy generation method explained in Section~\ref{learning}.

The proposed planner relies on mentally simulating interaction with the scene, hence an\-ti\-ci\-pa\-ting its dynamics using a physics simulation. Instead of using plain motion planning as in classical approaches, while planning manipulation of a specific object we explicitely disregard the presence of all other movable objects in the scene. However, the physics simulation takes into account possible motion of all objects which, after manipulation, is considered as a whole in order to retrospectively assess the success of the manipulation process.

In contrast to traditional motion planning ap\-proa\-ches, the presented one does not per se consider other movable objects as obstacles.
For this reason, the constraints on the planning process are significantly relaxed by decreasing the size of the planning search space.
Additionally, the presented method aims to avoid damaging the handled goods or even the robot itself. This could happen when dropping or shifting an object through the manipulation process.
Summarized, this manipulation sequence planning approach optimizes autonomous robotic behavior while regarding the anticipated dynamics of the perceived scene which is projected into the robot's mental simulation.

In the literature, an effective example of motion planning for a humanoid robot was presented by Okada et al.\ \cite{Okada2004}. In that work, any occuring obstacle is actively cleared from the envisioned walking trajectory, hence taking into account necessary actions which have not been planned explicitely when formulating the high-level goal.
In recent work in a different scenario \cite{Winkler2016}, a similar approach has been taken. In contrast, the work proposed in this publication uses a dynamics simulation for validating manipulation actions in addition to purely spatial knowledge. 

Kitaev et al.\ \cite{Kitaev2015} and Dogar et al.\ \cite{Dogar2012} work on grasp planning through clutter. They take into account shifting objects and explicitly manipulating these. The objective of these works is partly similar to the presented approach, though having the goal of removing obstacles from planned manipulation paths and not for determining optimal sequences.
On the other hand, Stilman et al.\ \cite{Stilman2007} explicitly plan which objects to move away and where to move them in order to reach a target object. In contrast, the proposed collision-agnostic approach mitigates excessive motion planning times which occur in the context of collision avoidance.

In the following, our planning method will be explained in detail. For this, let $\mathcal{O}$ be the set of objects present in the scene, $\alpha \in \mathcal{O}$ the \emph{active object} (that will be manipulated), $\Phi = \mathcal{O} \setminus \alpha$ the set of \emph{passive objects} (that will not be manipulated) and $\phi \in \Phi$ one member of this set.

\subsection{Manipulation cost estimation}\label{cost_estimation}
In order to determine suitable and efficient solutions to planning problems, cost functions have been used for a long time (e.g.\ in \cite{Goldberg1990,Okada2004}). Applying a cost function to a specific problem, however, requires a certain amount of domain knowledge for creating the optimization targets of the problem. The presented approach aims to keep the amount of injected knowledge as low as possible. Nevertheless, a cost function needs to be defined that regards spatial modifications of the scene in order to minimize unintended motions of passive objects.

\begin{figure}[tb]
    \centering
    \begin{subfigure}[t]{0.35\textwidth}
        \centering
        \includegraphics[width=\linewidth]{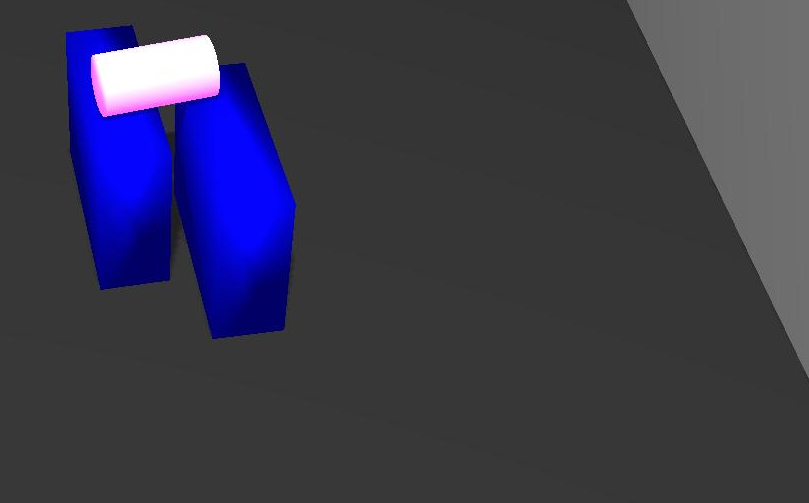} 
        \caption{pre-manipulation}
        \label{fig:mscv1}
    \end{subfigure}
    \begin{subfigure}[t]{0.35\textwidth}
        \centering
        \includegraphics[width=\linewidth]{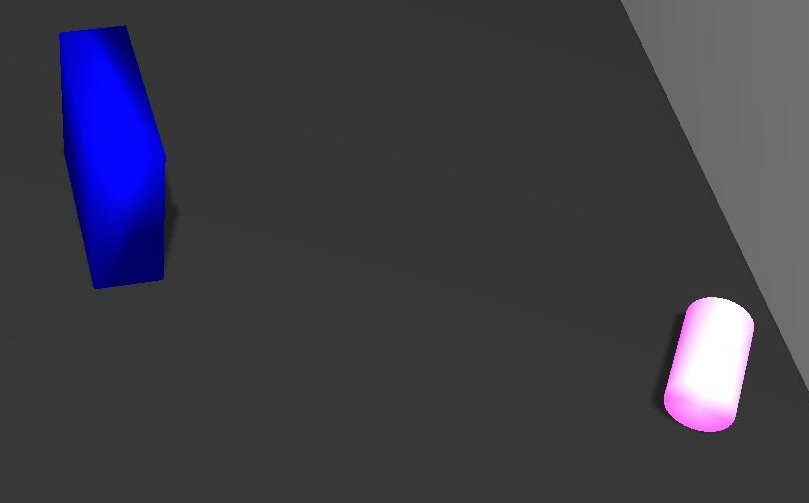} 
        \caption{post-manipulation}
        \label{fig:mscv2}
    \end{subfigure}
    \begin{subfigure}[t]{0.7\textwidth}
        \centering
        \includegraphics[width=\linewidth]{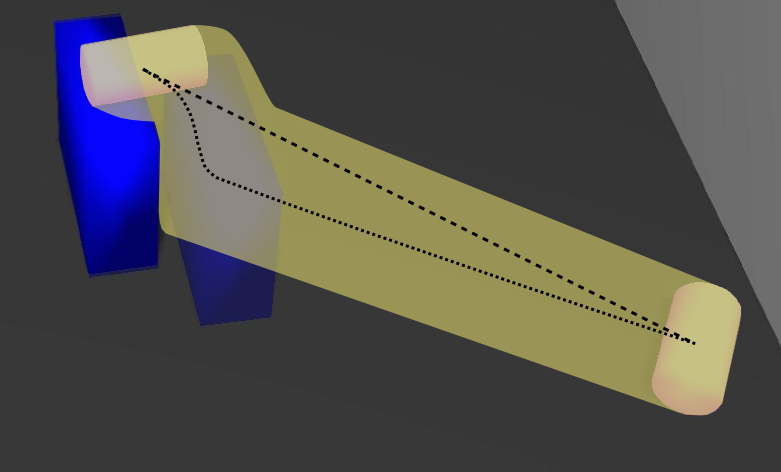} 
        \caption{trajectory (dotted line), Euclidean distance (dashed line) and swept convex volume (yellow)}
        \label{fig:mscv3}
    \end{subfigure}
    \caption{Visualization of the trajectory and swept convex volume covered by a passive object during manipulation. After the robot extracted the right \obj{1}{blue}, the cylindrical \obj{1}{pink} dropped on the floor and rolled away following the dotted trajectory.}
    \label{fig:mscv}
\end{figure}

One may first think of simple, generic Euclidean distance-based or trajectory length-based cost functions, however, in order to consider complex motion paths which may include translation and rotation, these do not deliver accurate cost estimates. 
In order to visualize this, Fig.~\ref{fig:mscv3} shows the movement of an object following a curved trajectory. On such a path, a Euclidean distance-based cost function, even if combined with rotational difference, does not take into account the whole volume (marked yellow in the figure) with all its curves and convexities. Such volumes are often covered when objects roll off uncontrolledly. Imagine an object rolling from the middle of a tabletop to its edge, then falling down, bumping off a wall and rolling back to the middle of the table. Such a complex motion cannot be covered by a distance-based cost function.
Trajectory length-based cost functions, however, more accurately model this behavior, but still do not reflect when an object with a high side length ratio (e.g.\ the salt container in Fig.~\ref{fig:simulation_objects:supermarket}) spins around all its axes. For low-fidelity recovery of movement costs this may be sufficient.

However, for the motions encountered in the targeted scenarios, a \emph{swept volume}-based representation gives a more accurate estimate of the object motion costs. Generally, a common application field of swept volume estimations is collision detection \cite{Baeuml2011} along with space occupancy estimation \cite{VonDziegielewski2012}. A mathematical formulation is given in the survey of Abdel-Malek et al.\ \cite{Abdel-Malek2006}.
Swept volume-based approaches use a \emph{generator} which creates the \emph{swept volume} by following a \emph{trajectory}.
In the case of the presented method, the generator is defined as the object surface. This follows a set of poses covered while moving during simulation runtime with the result of the outer object boundary during motion.
In contrast to \cite{Abdel-Malek2006}, this is a simplification in the sense that it is voxel-based instead of continuous and hence can be utilized in discrete-generator and discrete-trajectory scenarios.

The generated volume in most applications coincides with the concave hull around the spatial locations touched by the object. However, the effort required for the hull computation may exceed the provided capabilities \cite{VonDziegielewski2012}. Additionally, the concave hull of a number of points is not generally well-defined, which possibly creates ambiguities for different kinds of objects.
For the mitigation of these effects, in the proposed approach the concave hull is replaced with the convex hull, normalized by object volume, which is easy to compute and well-defined. The result is called \emph{swept convex volume} (for a visualization, see Figure~\ref{fig:mscv}).

Since all scene objects shall be regarded when computing manipulation costs, the \emph{maximum swept convex volume} $V_{max}$ is computed, which is the maximum over all scene objects' swept convex volumes:
\begin{equation}\label{eq:mscv}
	V_{max} = \max_{\phi \in \Phi}{V_s(\phi)}
\end{equation}
where $V_s(\phi)$ is the swept convex volume of the $\phi \in \Phi$ as computed in Algorithm~\ref{alg:scv}. 

\begin{algorithm}[tb]
\centering
\caption{Swept convex volume calculation}
\label{alg:scv}
\begin{algorithmic}[1]
    \State \textbf{input}: object mesh $\mathcal{M}(\phi)$, object poses $\mathbf{p}_{0..n}(\phi)$ covered during simulation
    \State create point cloud $\mathcal{C}(\phi)$ from $\mathcal{M}(\phi)$ at $\mathbf{p}_{0}(\phi)$
    \ForAll {$\mathbf{p}_{i}(\phi)$}
        \State $\mathcal{C}_{i}(\phi) \gets \mathcal{C}(\phi)$ transformed from $\mathbf{p}_{0}(\phi)$ to $\mathbf{p}_{i}(\phi)$\label{alg:scv:weight}
        \State $\mathcal{C}(\phi) \gets \mathcal{C}(\phi) \cup \mathcal{C}_{i}(\phi)$
    \EndFor
    \State $H(\phi) \gets \mathrm{convhull}(\mathcal{C}(\phi))$
    \State $H_{0}(\phi) \gets \mathrm{convhull}(\mathcal{C}_{0}(\phi))$
    \State $V_s(\phi) \gets \mathrm{volume}(H(\phi))~/~\mathrm{volume}(H_{0}(\phi))$
    \State \textbf{output}: swept convex volume $V(\phi)$
\end{algorithmic}
\end{algorithm}

As stated previously, the main focus of the presented method is to defend vulnerable goods from damage. Therefore, additional weights can be employed on each component of the 6-D pose to $V_{max}$, calling the resulting cost function the \emph{maximum weighted swept convex volume} $V_{w}$.
These weights are adaptable depending on the usage domain, e.g. to sanction objects dropping vertically off a shelf. Additional scenarios may include objects placed on a running conveyor belt where lateral influence may push an items off the belt. Employing rotational weights may help in scenarios where liquid-containing objects are prone to spill when tipped over.

The parameterization of said weights has influence on Algorithm~\ref{alg:scv} where, in Line \ref{alg:scv:weight}, $\mathbf{p}_{i}(\phi)$ has to be replaced with its weighted version $\mathbf{p}^{w}_{i}(\phi)$:
\begin{equation}
\mathbf{p}^{w}_{i}(\phi) = \mathbf{p}_{0}(\phi) + \mathrm{diag}(\mathbf{w})\cdot(\mathbf{p}_{i}(\phi)-\mathbf{p}_{0}(\phi))
\end{equation}
where $\mathbf{w}=\begin{bmatrix}w_{x}&w_{y}&w_{z}&w_{\varphi}&w_{\theta}&w_{\psi}\end{bmatrix}^{\intercal}$
are do\-main-de\-pen\-dent weights for each of the components of the 6-D object pose. The employment of weights $w_{\{x,y,z\}}>1$ in the translation domain has the effect of stretching the convex volume in the respective direction which results in enhanced costs.
In the rotation domain, $w_{\{\varphi,\theta,\psi\}}>1$ generates increased rotation of the object volume at a specific covered object pose. Assuming that the object does not perform a 360$^\circ$ turn on the respective axis during zero translation, this again increases the cumulative convex volume.

In general, the presented planning approach can be utilized for domains with other foci than damage avoidance by employing respective weights. For the envisioned application scenarios, however, we set the weights on
\begin{equation}
\mathbf{w} = \begin{bmatrix}1&1&2&1&1&1\end{bmatrix}^{\intercal}
\end{equation}
which puts emphasis on damage-prone vertical motion. This is of special importance in domains e.g.\ like the presented supermarket scenario where objects dropped off a shelf would break.


\subsection{Damage-avoiding manipulation sequence planning}\label{sequence_planning}
Using mental simulation of a robot's interaction with a scene as a validation method for planned actions is beneficiary additionally in time-critical scenarios as well as a second stage for a manipulation order planning algorithm.
Mojtahedzadeh et al.~\cite{Mojtahedzadeh2015}, for instance, present a planner that uses static equilibrium calculations for discovering physical support between objects. That planner consequently prefers objects not supporting any others.
Implicitly, the manipulation sequences produced by the proposed approach will often be similar. However, additionally, dynamic events are taken into account which occur during the manipulation action.
Consequently, using the proposed method for generating and validating plans on scenarios with dynamical and unpredictable content may enhance and enrich other planning algorithms.

In general, mentally simulated interaction can be used as a validation method which estimates the costs for manipulating a particular scene configuration. In order to deduce the next action for the robot to be performed, however, the distinction between positively and negatively validated actions, respectively, has to be drawn in some way. Naively, this can be achieved by modeling specific thresholds for the manipulation costs.
Common physics-based validation methods like the one of Rockel et al.~\cite{Rockel2015} determine such thresholds heuristically which signal whether or not some object is currently regarded as toppling. Pastor et al.~\cite{Pastor2011}, on the other hand, use statistical methods to determine if a particular motion coincides with a predefined spatial envelope. The drawback of these methods is the necessity to adjust said parameters whenever deploying the method in a new scenario.
However, in the vast majority of deployment settings, classification between actions causing positive and negative results in the respective context has to happen automatically. The user should not have to predefine thresholds and parameters in order to be adaptable to changing scenarios and environments. In contrary to the mentioned approaches, the main constribution of the method presented in this paper is that it works \emph{prior-free} in this respect (see \emph{Contribution~\ref{contribution1}} on p.~\pageref{contribution1}).

In the usage example of avoiding damage by unintendedly moving objects, the presented method provides manipulation sequences taking into account exactly these side-effects. Additionally, obstacle-avoiding motion planning is prone to fail in certain scenarios due to an overly complex motion planning problem in confined spaces. Stoyanov et al.\ sketch the difficulties of classical planning when applied in heavily confined spaces where passive objects are regarded as obstacles: \textit{"Most of the unloading failures were due to failures in finding collision-free grasping trajectories for objects [...] tightly packed with other objects."} \cite[p.~11]{Stoyanov2016}
However, the proposed planner tries to minimize the motion of passive objects during manipulation although this is explicitely permitted. Consequently, the likelihood of ending up with no viable manipulation plan is decreased significantly when using the presented method as a high-level manipulation sequence planner.

\begin{figure*}[tb]
    \centering
    \input{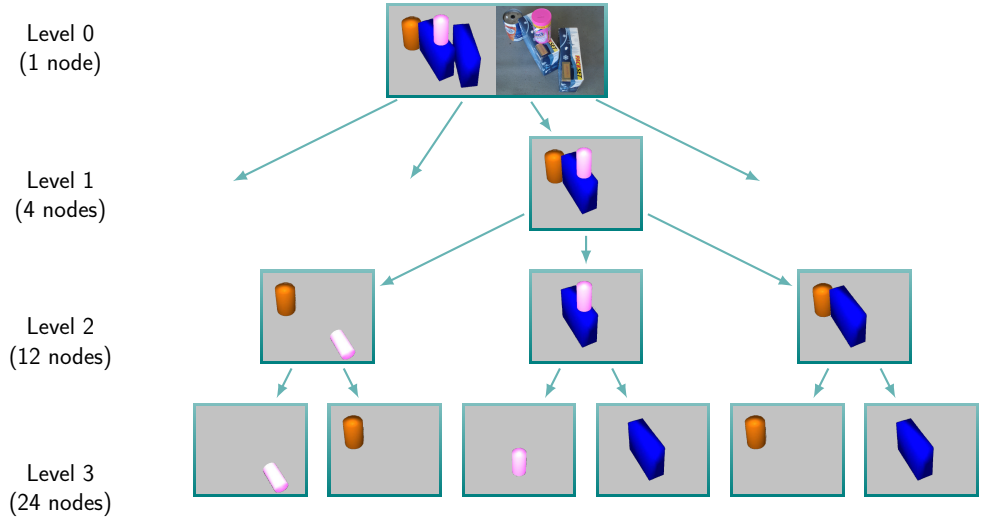}
	\caption{Example search tree}
    \floatfoot{Some branches cropped for increased visibility. Each node shows its initial configuration, i.e.\ prior to manipulation.}
	\label{fig:tree_example}
\end{figure*}

\begin{algorithm}[bt]
\centering
\caption{Search tree generation for manipulation sequence planning}
\begin{algorithmic}[1]
  \State initialize search tree $\mathcal{S}_{0} \gets \emptyset$
  \State initialize set of objects $\mathcal{O}_{0}$ with current scene
	\Procedure{createNode}{$\mathcal{S}_{i}$, $\mathcal{O}_{i}$}
	  \If{$|\mathcal{O}_{i}|\le 1$, i.e.\ this is a leaf node}
		\State\Return{}
	  \Else
		\State create new tree node $N_{i}$ containing object set $\mathcal{O}_{i}$ 
		\State $\mathcal{S}_{i+1} \gets \mathcal{S}_{i} \cup N_{i}$
		\State determine new active object $\alpha_{i} \in \mathcal{O}_{i}$
		\State$\mathcal{O}_{i+1} \gets \mathcal{O}_{i} \setminus \alpha_{i}$
		\State\Return{\Call{createNode}{$\mathcal{S}_{i+1}$, $\mathcal{O}_{i+1}$}}
	  \EndIf
	\EndProcedure
 \State \textbf{output}: filled search tree $\mathcal{S}$
\end{algorithmic}\label{alg:planning_tree_generation}
\end{algorithm}

The solution presented in this publication includes a \emph{search tree} containing all object configurations occuring during the sequenced manipulation of scene objects. Figure~\ref{fig:tree_example} shows an example of such a tree with the initial configuration appearing in the root (top). Traversing the tree towards its branches, the manipulation sequence is performed by removing one object each from the initial configuration. In the example, this causes extended costs for the \obj{1}{pink} which falls off the \obj{1}{blue} when the latter is manipulated (as shown in the leftmost depicted tree node on Level 2). In the tree leaves, only one object remains for direct manipulation without regarding the scene dynamics because no passive object is to be paid attention to.

\begin{algorithm}[tb]
\centering
\caption{Manipulation sequence planning for mental simulation}
\label{alg:planning}
\begin{algorithmic}[1]
    \State \textbf{input}: scene objects $\mathcal{O}$, search tree $\mathcal{S}$ generated from $\mathcal{O}$ ($\rightarrow$ Algorithm~\ref{alg:planning_tree_generation})
    \ForAll {nodes $N_{i} \in \mathcal{S}$}
        \State spawn all objects in $N_{i}$ in simulation
        \State plan approach $T_{1}^{\alpha}$ and extract $T_{2}^{\alpha}$ trajectory
        \State move simulated robot on $T_{1}^{\alpha}$
        \State grasp active object $\alpha$
        \State move simulated robot on $T_{2}^{\alpha}$
        \State release active object $\alpha$
        \State determine manipulation costs $V_{w,i}$ ($\rightarrow$ Section~\ref{cost_estimation})
	\EndFor
    \ForAll {leaf nodes $N_{j}^{\mathrm{leaf}} \in \mathcal{S}$}
		\State $V_{w,j} \gets$ summed-up costs of $N_{j}^{\mathrm{leaf}}$'s parents
		\State $V_{w,\mathrm{min}} \gets \mathrm{min}(V_{w,\mathrm{min}}, V_{w,j})$
	\EndFor
    \State \textbf{output}: node $N_{j}$ with lowest manipulation costs $V_{w,\mathrm{min}}$
\end{algorithmic}\label{alg:planning_anticipation}
\end{algorithm}

Algorithm~\ref{alg:planning_tree_generation} shows how a search tree $\mathcal{S}$ is generated for a particular scene given its initial object configuration. This works in a recursive way, starting from the root node (Level 0) which includes the initial configuration, down to the leaves where only one object is left in the scene. Each node $N_{i} \in \mathcal{S}$ contains the respectively assigned object set $\mathcal{O}_{i}$ as seen in Figure~\ref{fig:tree_example}, but not the poses of the objects.
After filling the tree, this is traversed like in Algorithm~\ref{alg:planning_anticipation}, using a depth-first search-like technique, with the objective of finding the node with the minimum manipulation costs $V_{w,\mathrm{min}}$. During traversal, the final state of the mental simulation of each node is propagated into its child nodes. This determines the initial object poses for the respective child node prior to running mental simulation on it.

Using depth-first search may seem inefficient at first glance. Standard search algorithms like $A^{*}$, however, require an admissible heuristic which needs to be adapted to the given problem domain.
Since damage avoidance is one of the main contributions of this article, the maximum weighted swept convex volume $V_w$ is used as a manipulation cost function which considers complex trajectories and changes of movement direction and spin. 
However, defining an admissible heuristic is not possible for this problem because the distance (measured in terms of the used cost function) to the target configuration cannot be determined before actually having searched the tree until its leaves.

Section~\ref{evaluation} shows the general feasibility and efficiency of the presented planning method. However, the whole mental simulation process may take extended time, especially for a growing number of scene objects. Therefore, the next section explains how to increase the overall efficiency by generating manipulation strategies which eventually replace time-consuming planning in the long run.

\section{Manipulation strategy generation}\label{learning}
One of the main contributions of the proposed approach is the \emph{generation of ma\-ni\-pu\-la\-tion strategies} from previously-planned manipulation sequences. Such a strategy comprises of a machine learning classifier which allows to \emph{predict a sequence of manipulation actions} while applying some prior knowledge.

In the given case, this prior knowledge consists of planned manipulation sequences on many different scenes where the prediction yields a desirable manipulation sequence for a new, previously unseen scene. All scenes can be characterized using certain distinct features; all features of a particular scene together with the anticipated optimal manipulation sequence form one training sample. \emph{Label ranking}, a machine learning technique, then takes the part of predicting new manipulation sequences from the known training data given a set of scene features.

Learning-based approaches in a manipulation context exist in a large diversity of applications and used techniques, including learning the support order of piles and the geometric relations between objects therein \cite{Mojtahedzadeh2015,Panda2013,Sjoo2011} or object affordances \cite{Katz2013}. However, manipulation strategy generation benefits from this only if it is able to take into account relational predicates like \texttt{in}, \texttt{on} and \texttt{behind} or support and containment relations. Additionally, assessing a scene for these predicates may yield ambiguous results, especially for geometry-based predicates. Finally, since we want to take the surrounding workspace into account as well, only describing relations between objects may not be sufficient for anticipating dynamic behaviors.

Another established field of research is grasp learning using different kinds of features and measures \cite{Fischinger2015,Kappler2015}. This works in a similar way as the presented method with respect to the generation of a classifier which allows for predicting grasps for new situations, but on grasp level. Other learning-based approaches tackle problems like predicting physical effects on objects using visual features \cite{Li2017,Mottaghi2016}. However, none of these works deal with high-level sequencing which is the main contribution of the proposed method.

\subsection{Label ranking using ranking by pairwise comparison}\label{lr}
A label ranking classifier $C$ basically solves the problem of \emph{ordering a set of abstract labels $\mathcal{L}$ with size $n=|\mathcal{L}|$ into a sequence $\pi \subseteq \mathcal{L}$} using a feature vector $\mathbf{x} \in \mathbf{X}$ such that
$C: \mathbf{X} \to \mathcal{L}, \mathbf{x} \mapsto \pi$.
The resulting sequence $\pi$ is a permutation of all occuring labels where label $\pi(i)$ is ranked higher than label $\pi(j)$. Generally, this \emph{pairwise preference of $\pi(i)$ over $\pi(j)$} is denoted as $\pi(i) \succ \pi(j)$.

In the literature, many different label ranking methods exist, based on different established machine learning algorithms. Amongst the most popular ones are Decision Tree-based \cite{DeSa2017,Cheng2009} and Gaussian Mixture Model-based methods \cite{Zhou2014,Grbovic2012} as well as \emph{ranking by pairwise comparison} (RPC) \cite{Huellermeier2008} which is used in the proposed approach.

RPC uses an ensemble of classifiers to predict pairwise preferences and afterwards employs a voting scheme to combine the atomic classifiers' outputs into a common prediction.
Because of this, one important parameter of RPC is the way of how to combine classifier outputs into a prediction. There are two major voting schemes which have be evaluated for this purporse so far \cite{Huellermeier2004}: \emph{binary voting} and \emph{soft voting}. The former bases on the result of a binary classifier which emits whether or not the input corresponds with its learned preference. Soft voting, on the other hand, outputs a continuous value in $[0,1]$ which can be interpreted as the confidence of the classifier about the compliance with the learned preference.

RPC proves advantageous for the targeted purpose since the used atomic classifiers correspond to pairwise preferences which play a major role in the overall method. Section~\ref{modeling_preference_patterns} will go into detail on this where pairwise preferences and their importance in human-like intuitive manipulation is explained.
One more major advantage of RPC in the regarded use case is the fact that it qualifies as an \emph{eager learning} method which uses a considerable amount of time for training the ensemble of classifiers, but prediction happens in near real-time. Since the whole processing circle of training scene generation, mental simulation and manipulation strategy optimization (which includes classifier training) is deferred into load-free times, no delays are inflicted on productive use.

\subsection{Distinction from related methods}
In contrast to label ranking, there are several methods which target problems overlapping with the presented one. However, these cannot be used in a similar way as the proposed method, but are important to be isolated from it because they produce structurally and logically different results:

\paragraph{Structured prediction}
Instead of ranked labels, this me\-thod outputs binary flags which mark if a particular label is relevant to the input sample or not. This can be used to classify if an image falls into a category like "nature" or "architecture". Examples of packages using this technique are $\mathrm{SVM^{struct}}$ \cite{Joachims2002} and conditional random fields \cite{Sutton2011}.

\paragraph{Learning to rank}
Although sounding similar in the first moment, this method provides solutions to a different problem, namely creating rankings of training samples and not rankings of abstract labels. One common usage is recording click-through preferences; the labels used as inputs into the algorithm are simply the sample indices in the training set. There is no possibility in this method to provide a ranking of abstract labels as training input. Prominent examples for learning to rank are RankNet \cite{Burges2005} and ranking SVMs like $\mathrm{SVM^{rank}}$ \cite{Joachims2002,Joachims2006}.

\paragraph{Sequence classification}
This term is usually used where only the sequences themselves, but no external training features exist, for instance when natural speech as a sequence of words is to be annotated with the respective word classes. Therefore this method only makes sense within context-rich environments where the exploitation of the latter satisfies the required generalization capabilities of some classifier. A survey of approaches using this technique can be found in \cite{Xing2010}.

\subsection{Scene features}\label{features}
Ranking by pairwise comparison relies on atomic classifiers which are based on multinomial logistic regression \cite{Huellermeier2008,LeCessie1992}. These classifiers are used to predict a confidence for a pairwise preference $\pi(i) \succ \pi(j)$ from a given feature vector $\mathbf{x}$. Such features have to be descriptive for the respective scene in order to allow for reliable manipulation scene prediction in a new, unknown scene. This subsection explains the feature vector that is calculated from every training scene as an input to the manipulation strategy generation process.

\subsubsection{Attentional vector sum}
As a measure for semantic spatial relations between scene objects, the \emph{attentional vector sum} (AVS) model is employed which is used in language comprehension research and was introduced by Regier and Carlson \cite{Regier2001}. It has been extended in several refined models (e.g.\ by Kluth \cite{Kluth2016}) and used for similar problems in robotics (e.g.\ by Sj\"o\"o and Jensfelt \cite{Sjoo2011}).

The AVS model generally evaluates spatial prepositions like \texttt{behind}, \texttt{above} and \texttt{left of} with respect to two distinct objects and provides an acceptability rating for the respective predicate in the shape of
\begin{equation*}
\mathtt{above(object1, object2)} = 0.1~\mathrm{where}~\mathtt{above(\cdot)} \in [0,1].
\end{equation*}
In the presented application, the AVS is evaluated for each of the prepositions \texttt{in front of}, \texttt{behind}, \texttt{above}, \texttt{below}, \texttt{left of} and \texttt{right of} from the robot's point of view for each permutation of two objects.

\subsubsection{Visibility}
Another important feature of a scene with respect to the manipulation order is object visibility which intuitively correlates with the manipulation difficulty level.
Hence, the visibility ratios $r_{\mathrm{vis}}$ of all objects $o \in \mathcal{O}$ is computed as follows:
\begin{equation}
r_{\mathrm{vis}}(o) = \frac{V(H(\mathcal{C}(o))}{V(H(\mathcal{M}(o))}
\end{equation}
where $\mathcal{C}(o)$ is the 2.5-D point cloud of $o$ visible in the scene, $\mathcal{M}(o)$ is $o$'s previously known 3-D object model, $H(\cdot)$ denotes a convex hull and $V(\cdot)$ the volume of such a hull.

\pgfplotstableread{roblog-dhl-box-blue-simple.gmm.csv}\gmmbox
\pgfplotstableread{roblog-dhl-box-blue-simple.frequencies.csv}\visibilitybox
\pgfplotstableread{roblog-maisels-barrel-simple.gmm.csv}\gmmmaisels
\pgfplotstableread{roblog-maisels-barrel-simple.frequencies.csv}\visibilitymaisels
\pgfplotstableread{roblog-vanish-detergent-simple.gmm.csv}\gmmvanish
\pgfplotstableread{roblog-vanish-detergent-simple.frequencies.csv}\visibilityvanish

\pgfplotstableread{shopping-mexico-mix-can.gmm.csv}\gmmmexico
\pgfplotstableread{shopping-mexico-mix-can.frequencies.csv}\visibilitymexico
\pgfplotstableread{shopping-cornflakes-box.gmm.csv}\gmmcornflakes
\pgfplotstableread{shopping-cornflakes-box.frequencies.csv}\visibilitycornflakes
\pgfplotstableread{shopping-jodsalz-salt-box.gmm.csv}\gmmjodsalz
\pgfplotstableread{shopping-jodsalz-salt-box.frequencies.csv}\visibilityjodsalz

\begin{figure*}[btp]
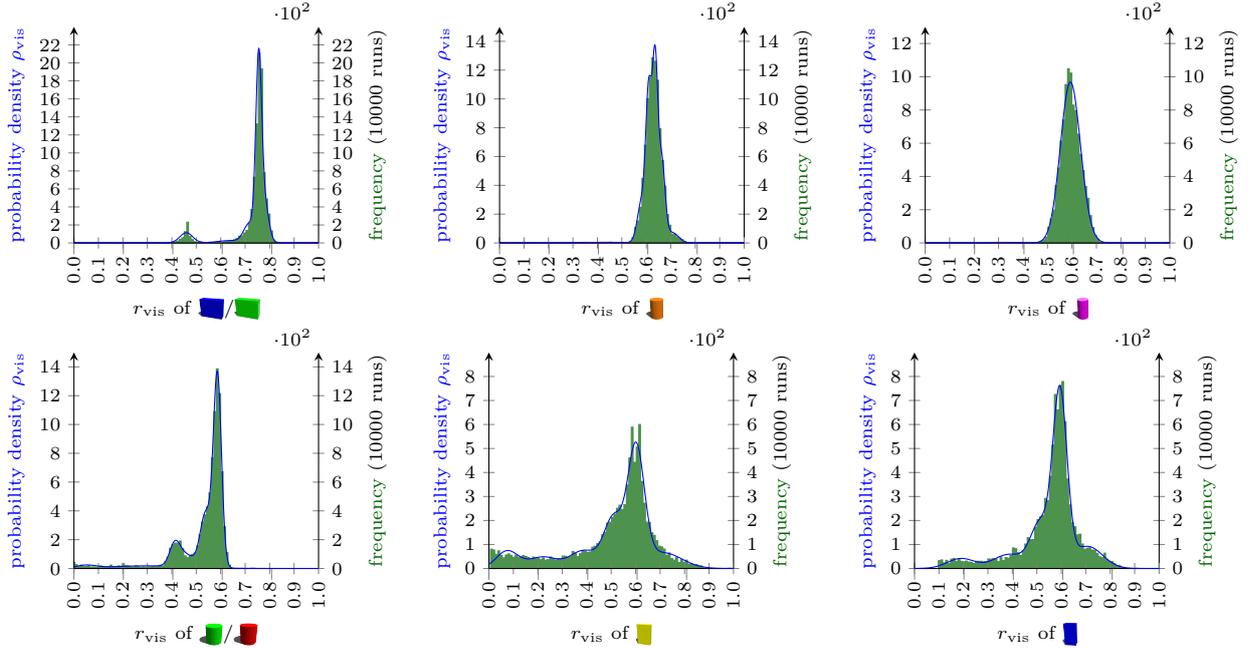

  \centering
    \scriptsize
    \pgfplotsset{every non boxed x axis/.append style={x axis line style=-}} 
  \begin{subfigure}{0.3\textwidth}
  \centering
    \begin{tikzpicture}[font=\scriptsize]

    \begin{axis}[
          axis y line=right,
          axis x line=bottom,
          ybar,
          bar width=0.008\textwidth,
          ylabel={\textcolor{DarkGreen}{frequency} (10000 runs)},
          ymin=0,
          ymax=2400,
          ytick={0,200,...,2300},
          xticklabels={,,},
          scaled y ticks=base 10:-2,
	      ylabel near ticks,
	      width=\textwidth,
          height=0.18\textheight
        ]
        \addplot[draw=none, fill=DarkGreen!70] table[x expr=\coordindex, y=Frequency] {\visibilitybox};
    \end{axis}

    \begin{axis}[
          ylabel={\textcolor{blue}{probability density $\rho_{\mathrm{vis}}$}},
          ymin=0,
          ymax=24,
          ytick={0,2,...,23},
          xmax=1000,
          xtick={0,100,...,1000},
          scaled x ticks=manual:{}{\pgfmathparse{#1/1000}},
	      xticklabel style={rotate=90,
            /pgf/number format/.cd,
            fixed,
            fixed zerofill,
            precision=1,
          /tikz/.cd},
          xlabel={$r_{\mathrm{vis}}$ of $\vcenter{\hbox{\includegraphics[height=\baselineskip]{scene1_obj0.png}}} / \vcenter{\hbox{\includegraphics[height=\baselineskip]{scene1_obj1.png}}}$},
          axis x line=bottom,
          axis y line=left,
	      ylabel near ticks,
	      width=\textwidth,
          height=0.18\textheight,
          tick label style={font=\sffamily}
        ]
        \addplot[color=blue] table[x expr=\coordindex, y=GMM_Score] {\gmmbox};
    \end{axis}

    \end{tikzpicture}
  \end{subfigure}
\hfill
  \begin{subfigure}{0.3\textwidth}
  \centering
    \begin{tikzpicture}[font=\scriptsize]

    \begin{axis}[
          axis y line=right,
          axis x line=bottom,
          ybar,
          bar width=0.008\textwidth,
          ylabel={\textcolor{DarkGreen}{frequency} (10000 runs)},
          ymin=0,
          ymax=1500,
          ytick={0,200,...,1400},
          xticklabels={,,},
          scaled y ticks=base 10:-2,
	      ylabel near ticks,
	      width=\textwidth,
          height=0.18\textheight,
          tick label style={font=\sffamily}
        ]
        \addplot[draw=none, fill=DarkGreen!70] table[x expr=\coordindex, y=Frequency] {\visibilitymaisels};
    \end{axis}

    \begin{axis}[
          ylabel={\textcolor{blue}{probability density $\rho_{\mathrm{vis}}$}},
          ymin=0,
          ymax=15,
          ytick={0,2,...,14},
          xmax=1000,
          xtick={0,100,...,1000},
          scaled x ticks=manual:{}{\pgfmathparse{#1/1000}},
	      xticklabel style={rotate=90,
            /pgf/number format/.cd,
            fixed,
            fixed zerofill,
            precision=1,
          /tikz/.cd},
          xlabel={$r_{\mathrm{vis}}$ of $\vcenter{\hbox{\includegraphics[height=\baselineskip]{scene1_obj2.png}}}$},
          axis x line=bottom,
          axis y line=left,
	      ylabel near ticks,
	      width=\textwidth,
          height=0.18\textheight,
          tick label style={font=\sffamily}
        ]
        \addplot[color=blue] table[x expr=\coordindex, y=GMM_Score] {\gmmmaisels};
    \end{axis}

    \end{tikzpicture}
  \end{subfigure}
\hfill
  \begin{subfigure}{0.3\textwidth}
  \centering
    \begin{tikzpicture}[font=\scriptsize]

    \begin{axis}[
          axis y line=right,
          axis x line=bottom,
          ybar,
          bar width=0.008\textwidth,
          ylabel={\textcolor{DarkGreen}{frequency} (10000 runs)},
          ymin=0,
          ymax=1300,
          ytick={0,200,...,1200},
          xticklabels={,,},
          scaled y ticks=base 10:-2,
	      ylabel near ticks,
	      width=\textwidth,
          height=0.18\textheight,
          tick label style={font=\sffamily}
        ]
        \addplot[draw=none, fill=DarkGreen!70] table[x expr=\coordindex, y=Frequency] {\visibilityvanish};
    \end{axis}

    \begin{axis}[
          ylabel={\textcolor{blue}{probability density $\rho_{\mathrm{vis}}$}},
          ymin=0,
          ymax=13,
          ytick={0,2,...,12},
          xmax=1000,
          xtick={0,100,...,1000},
          scaled x ticks=manual:{}{\pgfmathparse{#1/1000}},
	      xticklabel style={rotate=90,
            /pgf/number format/.cd,
            fixed,
            fixed zerofill,
            precision=1,
          /tikz/.cd},
          xlabel={$r_{\mathrm{vis}}$ of $\vcenter{\hbox{\includegraphics[height=\baselineskip]{scene1_obj3.png}}}$},
          axis x line=bottom,
          axis y line=left,
	      ylabel near ticks,
	      width=\textwidth,
          height=0.18\textheight,
          tick label style={font=\sffamily}
        ]
        \addplot[color=blue] table[x expr=\coordindex, y=GMM_Score] {\gmmvanish};
    \end{axis}

    \end{tikzpicture}
  \end{subfigure}
  \begin{subfigure}{0.3\textwidth}
  \centering
    \begin{tikzpicture}[font=\scriptsize]

    \begin{axis}[
          axis y line=right,
          axis x line=bottom,
          ybar,
          bar width=0.008\textwidth,
          ylabel={\textcolor{DarkGreen}{frequency} (10000 runs)},
          ymin=0,
          ymax=1500,
          ytick={0,200,...,1400},
          xticklabels={,,},
          scaled y ticks=base 10:-2,
	      ylabel near ticks,
	      width=\textwidth,
          height=0.18\textheight,
          tick label style={font=\sffamily}
        ]
        \addplot[draw=none, fill=DarkGreen!70] table[x expr=\coordindex, y=Frequency] {\visibilitymexico};
    \end{axis}

    \begin{axis}[
          ylabel={\textcolor{blue}{probability density $\rho_{\mathrm{vis}}$}},
          ymin=0,
          ymax=15,
          ytick={0,2,...,14},
          xmax=1000,
          xtick={0,100,...,1000},
          scaled x ticks=manual:{}{\pgfmathparse{#1/1000}},
	      xticklabel style={rotate=90,
            /pgf/number format/.cd,
            fixed,
            fixed zerofill,
            precision=1,
          /tikz/.cd},
          xlabel={$r_{\mathrm{vis}}$ of $\vcenter{\hbox{\includegraphics[height=\baselineskip]{scene3_obj0.png}}} / \vcenter{\hbox{\includegraphics[height=\baselineskip]{scene3_obj1.png}}}$},
          axis x line=bottom,
          axis y line=left,
	      ylabel near ticks,
	      width=\textwidth,
          height=0.18\textheight,
          tick label style={font=\sffamily}
        ]
        \addplot[color=blue] table[x expr=\coordindex, y=GMM_Score] {\gmmmexico};
    \end{axis}

    \end{tikzpicture}
  \end{subfigure}
\hfill
  \begin{subfigure}{0.3\textwidth}
  \centering
    \begin{tikzpicture}[font=\scriptsize]

    \begin{axis}[
          axis y line=right,
          axis x line=bottom,
          ybar,
          bar width=0.008\textwidth,
          ylabel={\textcolor{DarkGreen}{frequency} (10000 runs)},
          ymin=0,
          ymax=900,
          ytick={0,100,...,800},
          xticklabels={,,},
          scaled y ticks=base 10:-2,
	      ylabel near ticks,
	      width=\textwidth,
          height=0.18\textheight,
          tick label style={font=\sffamily},
        ]
        \addplot[draw=none, fill=DarkGreen!70] table[x expr=\coordindex, y=Frequency] {\visibilitycornflakes};
    \end{axis}

    \begin{axis}[
          ylabel={\textcolor{blue}{probability density $\rho_{\mathrm{vis}}$}},
          ymin=0,
          ymax=9,
          ytick={0,1,...,8},
          xmax=1000,
          xtick={0,100,...,1000},
          scaled x ticks=manual:{}{\pgfmathparse{#1/1000}},
	      xticklabel style={rotate=90,
            /pgf/number format/.cd,
            fixed,
            fixed zerofill,
            precision=1,
          /tikz/.cd},
          xlabel={$r_{\mathrm{vis}}$ of $\vcenter{\hbox{\includegraphics[height=\baselineskip]{scene3_obj2.png}}}$},
          axis x line=bottom,
          axis y line=left,
	      ylabel near ticks,
	      width=\textwidth,
          height=0.18\textheight,
          tick label style={font=\sffamily}
        ]
        \addplot[color=blue] table[x expr=\coordindex, y=GMM_Score] {\gmmcornflakes};
    \end{axis}

    \end{tikzpicture}
  \end{subfigure}
\hfill
  \begin{subfigure}{0.3\textwidth}
  \centering
    \begin{tikzpicture}[font=\scriptsize]

    \begin{axis}[
          axis y line=right,
          axis x line=bottom,
          ybar,
          bar width=0.008\textwidth,
          ylabel={\textcolor{DarkGreen}{frequency} (10000 runs)},
          ymin=0,
          ymax=900,
          ytick={0,100,...,800},
          xticklabels={,,},
          scaled y ticks=base 10:-2,
	      ylabel near ticks,
	      width=\textwidth,
          height=0.18\textheight,
          tick label style={font=\sffamily}
        ]
        \addplot[draw=none, fill=DarkGreen!70] table[x expr=\coordindex, y=Frequency] {\visibilityjodsalz};
    \end{axis}

    \begin{axis}[
          ylabel={\textcolor{blue}{probability density $\rho_{\mathrm{vis}}$}},
          ymin=0,
          ymax=9,
          ytick={0,1,...,8},
          xmax=1000,
          xtick={0,100,...,1000},
          scaled x ticks=manual:{}{\pgfmathparse{#1/1000}},
	      xticklabel style={rotate=90,
            /pgf/number format/.cd,
            fixed,
            fixed zerofill,
            precision=1,
          /tikz/.cd},
          xlabel={$r_{\mathrm{vis}}$ of $\vcenter{\hbox{\includegraphics[height=\baselineskip]{scene3_obj3.png}}}$},
          axis x line=bottom,
          axis y line=left,
	      ylabel near ticks,
	      width=\textwidth,
          height=0.18\textheight,
          tick label style={font=\sffamily}
        ]
        \addplot[color=blue] table[x expr=\coordindex, y=GMM_Score] {\gmmjodsalz};
    \end{axis}

    \end{tikzpicture}
  \end{subfigure}
\caption{Gaussian mixture model probability densities $\rho_{\mathrm{vis}}$ for visibility ratios $r_{\mathrm{vis}}$ of different objects in the logistics (top) and supermarket (bottom) scenarios}
\label{fig:gmm}
\end{figure*}

In order to abstract from object-dependent probability distributions for the visibility ratio which result from self-occlusion, 10000 scenes per object were generated with the object randomly placed and rotated in the workspace of the respective scenario.
Afterwards, a \emph{Gaussian mixture model} (GMM) has been fitted on the \textcolor{DarkGreen}{histograms of frequencies} of visibility ratios $r_{\mathrm{vis}}$ occuring in each run (see Figure~\ref{fig:gmm}).
The number of GMM components as well as the covariance type were optimized by minimizing the Bayesian Information Criterion \cite{Schwarz1978}.

The \textcolor{blue}{probability density $\rho_{\mathrm{vis}}$} of the GMM then is taken as the final visibility score which is still an object-specific value, but generalizes away from multimodal distibutions of the visibility ratios $r_{\mathrm{vis}}$ like for \obj{1}{0} and \obj{1}{1} as can be seen in Figure~\ref{fig:gmm}.

Note that the visibility scores of the objects in the supermarket scenario (Fig.~\ref{fig:gmm} bottom row) are less distinct than the objects of the logistics scenario (Fig.~\ref{fig:gmm} top row). This is caused by the high-altitude camera position which requires a strong tilt in order to have the whole shelf in view. Hence, the sensor data on the margins of the field of view may sometimes be cropped, causing objects' visibility ratios to drop when randomly spawned close to the shelf margins.
Please also note that $r_{\mathrm{vis}}$ usually does not take values close to $0$ because some part of the object is always visible, otherwise it would not have been recognized by the perception system. Additionally, $r_{\mathrm{vis}}$ values close to $1$ do not occur because the 2.5-D input point clouds are subject to self-occlusion, hence the back part of the objects is cropped.

\subsubsection{Feature vector}
In addition to the more complex AVS and visibility features, several object-specific and object-relational features are computed as shown in Table~\ref{table:features}.

The final set consists of $23n+17p$ features with the number of objects in the scene being $n=\lvert\mathcal{O}\rvert$ and the number of all possible object pair combinations being $p=\binom{n}{2}$. For a scene comprising four objects, this would end up in a $194$-dimensional feature vector $\mathbf{x}$. One concrete instance of $\mathbf{x}$, together with the respective manipulation sequence $\pi$ as determined from mental simulation, eventually forms one training sample for $C$ with $\mathbf{x} \mapsto \pi$.

\begin{table}[hbt]
    \rowcolors{1}{}{lightgray}
    \centering
    \begin{tabular}{p{6cm}c}
        \textbf{Feature (object-specific)} & \textbf{Dimensionality} \\
        Pose within the scene & $6n$ \\
        Distance vector to workspace bottom & $3n$ \\
        Distance vector to workspace back & $3n$ \\
        Distance vector to initial gripper position & $3n$ \\
        Visibility $\rho_{\mathrm{vis}}$ & $n$ \\
        Axis-aligned bounding box size & $3n$ \\
        Oriented bounding box size & $3n$ \\
        \parbox[t]{6cm}{Free space around object\\ \scriptsize{(= Euclidean distance to closest object's surface)}} & $n$ \\
        \textbf{Feature (object-relational)} & \textbf{Dimensionality} \\
        Distance vector per object pair & $3p$ \\
        Euclidean distance per object pair & $p$ \\
        Contacts per object pair (point, normal and force) & $7p$ \\
        AVS for 6 prepositions per object pair & $6p$ \\
        \textbf{Total} & $\mathbf{23n+17p}$\\
    \end{tabular}
    \caption{Features and dimensions depending on the number of objects $n=\lvert\mathcal{O}\rvert$ and the number of object pair combinations $p=\binom{n}{2}$}
    \label{table:features}
\end{table}

\subsection{Subconscious preference patterns}\label{modeling_preference_patterns}
Whenever a human needs to decide between two alternatives like "Should I take A or B?", they use some inherent classification scheme from their prior knowledge to come up with a decision. In a situation that requires motoric interaction on two objects, the decision which one to manipulate is driven by the concrete scene and objects, but also by physical properties of the human itself.

Imagine, for instance, a right-handed user who, whenever possible with respect to occlusions, reachable work\-space and other kinetic limitations, always grasps with their right hand. For a setting where two objects are similarly reachable, but one needs to be grasped with the right and the other one with the left hand, a right-handed human, minimizing physical effort, would naturally know which one to grasp best.

Now imagine a scenario with a robot for which grasping object A is easier than grasping object B in a specific scene because of
\begin{itemize}
    \item limited workspace due to manipulator size, position on the robot base and number of degrees of freedom
    \item scene obstacles preventing from reaching object B
    \item uneven probability distributions in the stochastic motion planner which make it more likely to obtain solutions for object A
    \item the grasping pose of object B, which is sampled from the noisy object representation, being located outside of the manipulator workspace.
\end{itemize}

\begin{figure}[tb]
    \centering
    \includegraphics[width=0.7\linewidth]{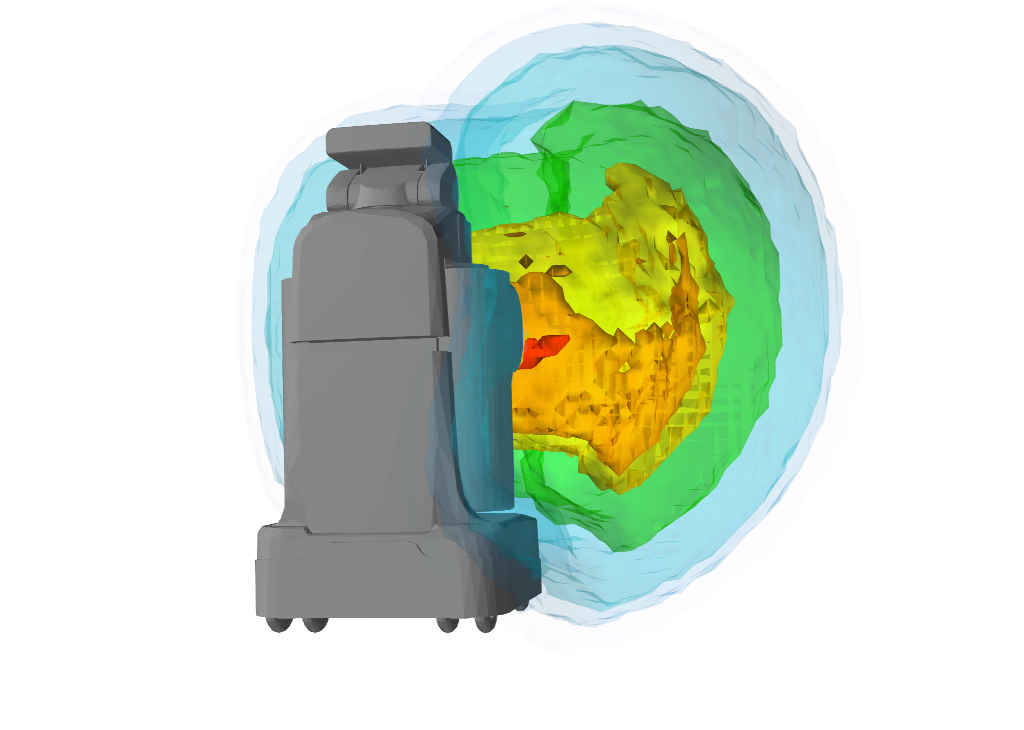}
    \caption{Kinematic reachability of the PR2's right arm\\ \tiny{adapted from \url{http://openrave.org/docs/0.6.6/\_images/tutorial\_inversereachability\_back.png}, $\copyright$ 2006-2018 OpenRAVE (CC BY 3.0)}}
    \label{fig:pr2_workspace}
\end{figure}

One good example for such a scenario is the PR2 robot in the supermarket (Fig.~\ref{fig:intro_scenario_shopping}) which, for the ease of obtaining viable grasping configurations, is used in a right handed-only way, i.e.\ the left arm is tucked up while grasping only with the right one. Consequently, a tendency to grasp objects on the right first can be observed in scenarios using this robot configuration -- see the robot arm's kinematic reachability in Fig.~\ref{fig:pr2_workspace}.

Therefore, noise on object pose and grasping pose level plays an important role when considering which object to prefer for grasping. Many publications deal with circumventing such noise and finding manipulation plans nevertheless, e.g.\ \cite{Battaglia2013,Vaskevicius2014a,Mojtahedzadeh2015a,Weisz2012}, amongst many others.

However, humans deal with such noise intuitively, implicitely taking into account all mentioned constraints. Over many observations, \emph{subconscious preference patterns} become apparent as shown in experiments. The proposed approach integrates these patterns in a way that, as for determining a manipulation sequence for an unknown scene, the robot behaves humanlike in a way that certain manipulation actions are more likely than others for the reasons stated above.
Since many underlying sources are encoded into the presented method's output, changing properties like the used motion planner or the robot model will get reflected in the manipulation sequence predictions similar to human long-term knowledge.

The evaluation (see Section~\ref{evaluation}) shows how such preference patterns emerge from the considered application scenarios. As described above, \emph{ranking by pairwise comparison} serves as a label ranking method which is particularly useful because it employs the mentioned preference patterns to create a classifier which can derive a ranking. The next subsection describes how this method is applied in the course of the overall approach.

\subsection{Preference-weighted label ranking}\label{ranking_weighted}
Using the preference patterns as described above and the feature vector $\mathbf{x}$, the original \emph{ranking by pairwise comparison} (RPC) method combines them into a final ranking by calculating the sum of votes like described in \cite{Huellermeier2008}:
\begin{equation}
s = \displaystyle\sum_{\pi(i) \succ \pi(j)} C_{ij}
\end{equation}
where $C_{ij}: \mathbf{X} \to \mathcal{L}, \mathbf{x} \mapsto \pi$ is the output of the base classifier for $\pi(i) \succ \pi(j)$ and $n$ labels. In case of using \emph{soft voting} (see Section~\ref{learning}), $C_{ij} \in [0,1]$, whereas for \emph{binary voting} $C_{ij} \in \{ 0,1 \}$, i.e.\ the input does or does not belong to the respective class.

Since the goal is to regard the label ranking process from a preference-pattern viewpoint, the following enhancement allows for taking into account \emph{preference weights} $w_{ij}$ when computing the sum of votes:
\begin{equation}
s = \displaystyle\sum_{\pi(i) \succ \pi(j)} v_{ij}
\end{equation}
where
\begin{equation}
    v_{ij} =
\begin{cases}
    w_{ij} \cdot C_{ij} & \text{if } C_{ij} > 0.5,\\
    0 & \text{otherwise,}
\end{cases}
\end{equation}
\begin{equation}\label{eq:preference_weights}
    w_{ij} = \frac{|\{(i,j)~|~\pi(i) \succ \pi(j)\}|}{m},
\end{equation}
and $m$ is the number of training samples.

The weights $w_{ij}$ are used to balance a certain preference $\pi(i) \succ \pi(j)$ versus its inverse $\pi(j) \succ \pi(i)$. Given the training data with 100 samples per scene to minimize the influence of noise in the physics simulation and motion planner, they are formed by the proportion of a particular preference in relation to the whole training data. This way, the resulting ratio indicates how likely a particular object will be preferred over another in this particular scene configuration.

In the practical process of generating manipulation strategies, now a classifier is trained using the described preconditions and additions to RPC. The resulting classifier, spawned from the training data which was extracted from auto-generated scenes, forms the current \emph{manipulation strategy} of the robot in the given environment. This strategy can be optimized in a simulation-in-the-loop cycle as explained in detail in Section~\ref{self_improvement}.

However, experimental evaluation shows that the integration of preference weights into the classifier training process did not show a significant increase in prediction fidelity for the given use cases compared to the original RPC method (see Section~\ref{results_rpc}).
Nevertheless, we use the preference weights to meaningfully compare two manipulation strategies.
In order to do this, a loss function $l(\pi,\pi')$ which gives the difference between two rankings $\pi$ and $\pi'$ has to be defined.
In the presented approach the number of pairwise preferences is counted which appear in inverse order (\emph{discordant preferences}) compared to the training data, i.e.
\begin{equation}
l(\pi,\pi') = |\{(i,j)~|~\pi(i) \prec \pi(j) \land \pi'(i) \succ \pi'(j)\}|.
\label{eq:measure_kendalls_loss}
\end{equation}

Since, for evaluation of the presented method, a similarity measure is needed in order to make sound statements about the resulting rankings, we use the established \emph{Kendall rank correlation coefficient} \cite{Kendall1938}, commonly denoted as \emph{Kendall's tau coefficient} $\tau: \mathcal{L} \times \mathcal{L} \to \mathbb{Q}, (\pi,\pi') \mapsto \tau$ with
\begin{equation}
\tau(\pi,\pi')=1-\frac{4\,l(\pi,\pi')}{n(n-1)}
\label{eq:measure_kendalls_tau}
\end{equation}
where $n=|\mathcal{L}|$ is the number of labels to appear in the ranking. In the use case of manipulation sequences, this equals with the number of objects in the scene.

As a side note, in the context of label ranking, many approaches use some form of weights to allow for adaptation of their algorithms to the respective domain. Nevertheless, the definition of these weights usually does not match the preference weights defined above, like in \cite{Shalev-Shwartz2006} where the authors assign per-label relevance weights different in every sample.
The same applies for the labelwise weight variant of $\tau$ in Kumar and Vassilvitskii's work \cite{Kumar2010} which only depend on the label itself, but not on its relation to other labels.

Since there is the necessity for preference weights which consider a pairwise permutation of labels, though, the \emph{preference-weighted Kendall's tau} rank correlation measure $\tau_{w}: \mathcal{L} \times \mathcal{L} \to \mathbb{R}, (\pi,\pi') \mapsto \tau_w$ is proposed with
\begin{equation}\label{eq:preference_weighted_kendalls_tau}
    \tau_w(\pi,\pi') = 1-\frac{4\,l_{w}(\pi,\pi')}{n(n-1)},
\end{equation}
\begin{equation}\label{eq:preference_weighted_kendalls_loss}
    l_{w}(\pi,\pi') = \displaystyle\sum_{\pi(i) \succ \pi(j)} d_{ij},
\end{equation}
\begin{equation}
    d_{ij} =
\begin{cases}
        \max{(w_{ij}, w_{ji})} - 0.5 &
    \begin{aligned}[t]
         \text{if } & \pi(i) \prec \pi(j) \\
         & \land \pi'(i) \succ \pi'(j)
    \end{aligned}\\
        0 & \text{otherwise}
\end{cases}
\end{equation}

where $w_{ij}$ is the preference weight of the respective pairwise preference $\pi(i) \succ \pi(j)$ computed from the noisy training data of one scene as in Eq.~\ref{eq:preference_weights} and $n=|\mathcal{L}|$ is the number of labels to appear in the ranking.

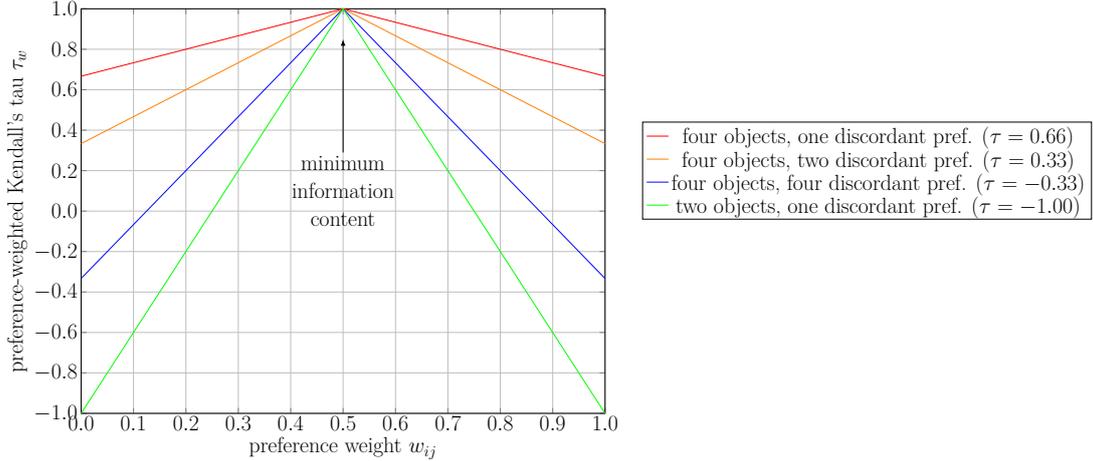
\begin{figure}[bt]
\centering
\resizebox{0.9\linewidth}{!}{
\begin{tikzpicture}[font=\LARGE]
\begin{axis}[
      xtick={0.0,0.1,...,1.1},
      x tick label style={/pgf/number format/.cd,fixed,fixed zerofill,precision=1,/tikz/.cd},
      xmin=0.0,
      xmax=1.0,
      ytick={-1.0,-0.8,...,1.1},
      y tick label style={/pgf/number format/.cd,fixed,fixed zerofill,precision=1,/tikz/.cd},
      ymin=-1.0,
      ymax=1.0,
	  xmajorgrids=true,
	  ymajorgrids=true,
      xlabel={preference weight $w_{ij}$},
      ylabel={preference-weighted Kendall's tau $\tau_{w}$},
	  width=\linewidth,
	  height=0.8\linewidth,
      legend style={at={(1.5,0.6)},anchor=center}
    ]
	\addplot+[mark=, color=red]file{relevance_weights_vs_weighted_tau_4obj_1dis.csv}; \addlegendentry{four objects, one discordant pref. ($\tau=0.66$)}
	\addplot+[mark=, color=orange]file{relevance_weights_vs_weighted_tau_4obj_2dis.csv}; \addlegendentry{four objects, two discordant pref. ($\tau=0.33$)}
	\addplot+[mark=, color=blue]file{relevance_weights_vs_weighted_tau_4obj_4dis.csv}; \addlegendentry{four objects, four discordant pref. ($\tau=-0.33$)}
	\addplot+[mark=, color=green]file{relevance_weights_vs_weighted_tau_2obj.csv}; \addlegendentry{two objects, one discordant pref. ($\tau=-1.00$)}
    \node(note) at (axis cs:0.5,0.1) [anchor=center, align=center] {minimum\\information\\content};
    \draw[-latex, thick] (note) to[out=90, in=270] (axis cs:0.5,0.85);
\end{axis}
\end{tikzpicture}
}
\caption{Preference-weighted Kendall's tau $\tau_{w}$ relative to preference weights $w_{ij}$ with respect to different numbers of discordant preferences and objects}
\label{fig:weighted_kendalls_tau}
\end{figure}

This way, wherever the distance between two rankings is to be calculated using $\tau_{w}$, every preference that appears only in one of the rankings gets weighted with the respective preference weight. Figure~\ref{fig:weighted_kendalls_tau} shows an example of $\tau_w$ scaling between 1.0 and the respective $\tau$ when the weights scale between 0.5 (completely random, no preference, minimum information content) and 1.0 (strong preference, maximum information content).
The resulting $\tau_{w}$ allows for continuous values with an image cardinality of $\lvert \tau_{w}[\pi,\pi'] \rvert = \lvert \mathbb{R} \rvert$ as opposed to $\tau$ which has a discrete image with a cardinality of only $\lvert \tau[\pi,\pi'] \rvert = 2n-1$ for $n$ labels. This bears the advantage that the more fine-grain $\tau_{w}$ can distinguish pairs of sequences according to preference relations where the coarse-grain $\tau$ provides the same measure for these pairs. The evaluation section shows examples why and how this is useful for the proposed method.

Summarized, this section described how to generate manipulation strategies from mentally simulated manipulation sequences. These can now be deployed in a real application scenario and continuously optimized as described in the following.

\section{Wrapping everything up: Self-supervised free-time manipulation strategy optimization for anytime deployment}\label{self_improvement}
Having defined the full procedure of auto-generating training scenes and extracting the respective training data, we are now able to generate manipulation strategies using the described ranking by pairwise comparison classifier and optimize them in a self-supervised manner. This happens during load-free times of the robot which typically occur during night hours or weekends or even, guarded by a task scheduler, during regular operation while the current CPU load permits.

\algblockdefx{WhileP}{EndWhileP}[1]%
  {\textbf{while }{#1} \textbf{do in parallel}}%
  {\textbf{end while}}
\algblockdefx{WhileSP}{EndWhileSP}[1]%
  {\textbf{while }{#1} \textbf{do in several \emph{Docker} containers in parallel}}%
  {\textbf{end while}}
\begin{algorithm*}[bth]
\centering
\caption{Self-supervised manipulation strategy optimization}
\label{alg:optimization}
\begin{algorithmic}[1]
    \State generate initial set of training scenes\footnotemark
    \State extract samples from training scenes using mental simulation
    \State add new samples to training and testing set with a 2:1 ratio
    \WhileSP{CPU load low and not interrupted}
        \WhileP{not interrupted}\label{alg:optimization:loop_begin}
            \State update classifier
                \State\hspace{\algorithmicindent} -- retrain classifier with new training set
                \State\hspace{\algorithmicindent} -- evaluate classifier on testing set, discard new samples if $\tau_{w}$ decreased
            \State generate more training samples
                \State\hspace{\algorithmicindent} -- generate training scenes
                \State\hspace{\algorithmicindent} -- extract samples from training scenes using mental simulation
                \State\hspace{\algorithmicindent} -- add new samples to training and testing set with a 2:1 ratio
        \EndWhileP\label{alg:optimization:loop_end}
    \EndWhileSP
    \State deploy classifier anytime to predict a manipulation sequence for a real scene
\end{algorithmic}
\end{algorithm*}
\footnotetext{The initial training set size should depend on the maximum tolerable time until the classifier is required for the first time.}

Algorithm~\ref{alg:optimization} describes the individual steps of the strategy optimization cycle. Depending on the free computing capacities on the robot, one iteration of this loop varies in runtime, hence the per-time utility improvement of the respective classifier is subject to other tasks being executed at the same time. Therefore a manipulation strategy may take a long time to converge\footnote{Convergence of the optimization method strongly depends on the concrete implementation and application scenario and is hard to define generically as explained in the evaluation section.} if not operated during load-free runtimes. The prediction of manipulation sequences on real scenes, however, happens instantly in near-real time with the currently built classifier. The presented method can be classified as an \emph{anytime algorithm} because it always delivers a valid result, even when interrupted, and improves upon its solutions the longer it keeps running.

The next section shows how, through multiple strategy optimization cycles, the overall prediction accuracy increases and the manipulation strategy adheres more and more to the discovered preference patterns. Nevertheless, if the user decides to accept the currently active strategy for deployment, they can do so anytime. However, as soon as the next load-free time slot appears, the manipulation strategy optimization cycle can be continued where it was interrupted.
As for bootstrapping the proposed method on a newly deployed robot, the latter initially incorporates no knowledge about manipulation strategies, but the environment and object models have to be known beforehand. The method does not put any semantic constraints on the objects or the combinations into which they are grouped to train an individual classifier. Hence, the order of object combinations to be learned can be purely application-driven, e.g.\ commencing with classifiers for combinations of low numbers of objects in order to quickly converge towards viable classifier performance. Later during robot lifetime, training of larger combinations can be performed which is more computationally intensive, but the robot can meanwhile continue working with the classifiers trained up to that point.

One major advantage of the proposed self-supervised strategy optimization meth\-od is that it allows for effortless parallelization. All experiments were performed using a setup of multiple similar \emph{Docker}\footnote{\url{https://docker.com}}~\cite{Merkel2014} containers executed in parallel, increasing the overall system efficiency even more. Even more, the generation of training samples can run in parallel to updating the classifier with the existing set of samples, so this increases the system efficiency even more.
A generic version of the integrated parallelized container setup is provided online\footnote{\url{https://github.com/jacobs-robotics/gazebo-mental-simulation}} together with some usage examples and Gazebo models. Additionally, the extracted features from all scenes as used in the experiments in the next section are publicly available along with the trained classifiers\footnote{\url{https://tobias.doernba.ch/research/datasets/mental-simulation}}.

Summarized, the presented \emph{simulation-in-the-loop} approach allows for continuous self-su\-per\-vised optimization of the current manipulation strategy, leading to more accurate predictions with respect to reality over robot lifetime.

\section{Evaluation}\label{evaluation}
In order to evaluate the presented approach, the coherency with respect to human behavior as well as efficient strategy optimization cycles play an important role. 
In this section, the worst-case complexity of different parts of the approach is estimated. Additionally, several measures are explained which were taken to improve the overall efficiency. Afterwards, several experiments on different application scenarios show the performance of the presented method within the specific steps and, finally, the performance of the full self-supervised strategy optimization cycle.

\subsection{Efficiency considerations}\label{efficiency}
\subsubsection{Training data generation}
Section~\ref{ranking_weighted} described an idea of how to integrate preference weights $w_{ij}$ into the RPC classifier. However, as shown in the experimental evaluation (Section~\ref{experiments}), the weights do not significantly improve the performance of the classifier if integrated into the prediction process. Nevertheless, preference weights are used to calculate $\tau_{w}$ as a performance measure for evaluation.

On the other hand, this means that the manipulation strategy generation part of the method experiences a speedup of 100 with respect to integrating the preference weights into classifier training because they do not need to be calculated during self-supervised fully-autonomous operation. Hence, it is not necessary to plan manipulation sequences for a specific scene more than once.

\subsubsection{Planning efficiency}
The described planning method, in contrast to the mentioned $A^{*}$ algorithms or other commonly used search methods like discretized Rapidly-Exploring Random Trees (RRT) or Rapidly-Exploring Random Leafy Trees (RRLT) \cite{Morgan2004}, is critical with respect to the efficiency of estimating the cost function for each node. The search process itself is trivial and happens quickly because of the relatively low number of nodes compared to, for instance, a motion planning problem. Therefore, the efficiency of the method depends mainly on the problem of cost estimation and not the worst-case tree size $|\mathcal{S}_O|$ which, for a number of objects $n$, constructed from all possible permutations of the object set like in Algorithm~\ref{alg:planning_tree_generation}, is
\begin{equation}\label{eq:tree_size}
	|\mathcal{S}_O| = \sum_{i=0}^{n} \frac{n!}{(n-i)!} = 1+\frac{n!}{(n-1)!}+...+\frac{n!}{2}+n!
\end{equation}
where $\frac{n!}{(n-i)!}$ is the number of objects for the respective tree level $i$, $i=0$ representing the top level (i.e.\ root), $i=n$ the bottom level (i.e.\ leaves); see the example in Figure~\ref{fig:tree_example}. 
With respect to the time complexity of manipulation sequence planning, since the used depth-first search possesses a worst-case time complexity of $O(|V| + |E|)$ with $|V|$ being the number of vertices and $|E|$ being the number of edges in the search tree \cite{Reif1985}, Algorithm~\ref{alg:planning} reaches
\begin{equation}\label{eq:complexity}
	O(|\mathcal{S}_O| + (|\mathcal{S}_O|-1)) = O\left(2\sum_{i=0}^{n} \frac{n!}{(n-i)!}-1\right)
	= O\left(1+\frac{2n!}{(n-1)!}+...+n!+2n!\right) \implies O(n!).
\end{equation}

On the other hand, best-case time complexity can be achieved if the costs of the very first leaf node are lower than the costs of any other node considered later in the process. In this case, all children of these following nodes can be disregarded. The minimal number of nodes to consider hence amounts to
\begin{equation}\label{eq:tree_size_omega}
	|\mathcal{S}_{\Omega}| = \sum_{i=1}^{n-1} (n-i) = 1+2+...+(n-1)
	= \frac{n(n-1)}{2}-n = \frac{n(n-3)}{2}
\end{equation}
which results in a best-case time complexity of
\begin{equation}\label{eq:complexity_omega}
	\Omega(|\mathcal{S}_{\Omega}| + (|\mathcal{S}_{\Omega}|-1))
	= \Omega\left(n(n-3)-1\right) \implies \Omega(n).
\end{equation}
In practice, best-case complexity is not as unlikely to achieve as it may seem because during many manipulation actions the passive objects are not moved at all, hence causing zero costs.
Either way, the experiments in the next subsection show that the average-case time complexity is favorable in practice due to different possibilities of pruning the tree during planning.

\subsubsection{Scene clustering}
Equation~\ref{eq:tree_size} showed a nonlinear increase in search tree size with a rising number of scene objects. As a remedy, it is suggested to break down scenes with many objects into smaller clusters of $n \leq 4$ objects which are then treated individually. Practically, as regarded in Section~\ref{modeling_preference_patterns}, the robot will anyway not be able to physically access more than a low number of objects from a certain point of view due to space and dexterity constraints.
However, the exact way of clustering scene objects strongly depends on the scenario and a generic consideration of this subproblem does not fit the scope of this publication and has to be treated in future work.
Most importantly, however, is the fact that during deployment of a trained classifier in a productive setting the per-cluster application of an individual classifier instead of one single classifier for the whole scene creates no tangible efficiency loss since the prediction runtimes of RPC lie in the sub-second range.

In any case, using smaller object clusters mitigates the issue that the reachability of scene objects strongly depends on the robot kinematics and hence often is limited by a large extent.
In the course of complex manipulation procedures like used in the proposed approach, the higher the number of objects in a cluttered scene grows, the less likely any valid grasping configurations can be generated for each individual object. To stay with the PR2 supermarket example, a right-handed robot may have severe difficulties manipulating objects on its left-hand side. In this case it is reasonable to handle clusters to the left of the robot with lower priority if only the right arm is used for manipulation.
Nevertheless, even when only using a part of the scene as active objects, the remaining objects should still be included in the simulated scene. This way, if any passive object is moved accidently, even if it is not active anywhere in the search tree, it will still account for the cost function.

\subsubsection{Search tree optimization}
Clustering the scene into smaller parts dramatically increases the method's efficiency, but additionally, it is desired to keep the search tree as non-redundant as possible any time. For instance, tree branches where there is no possibility to represent an optimal solution can be pruned. Several measures are presented in this subsection which keep the search tree as small as possible. Section~\ref{experiments} shows that the presented method is able to prune a typical search tree down to \SI{48.7}{\%} of its original size using the following strategies.

\paragraph{Implicit search tree pruning:}
As shown by an empiric consideration in the Experiments section, in every scene there is a number of configurations for which the simulation of a manipulation action is pointless, e.g.\ touching a passive object during grasp approach which in turn pushes the active object into unreachable distance.
Another possibility is that all configurations include collisions of the robot with the environment. In this work, the low-level motion planning process is abstracted away since it represents an own branch of research. Hence, such configurations have to be imposed with infinite costs and therefore can be removed from the tree of viable configurations.

\paragraph{Explicit search tree pruning:}
Since the planner uses depth-first search as explained in Algorithm~\ref{alg:planning}, some tree bran\-ches can be pruned because the costs accrued in the currently active branch already exceed the total costs of any other branch that has been traversed to its leaf. Exploration of the current branch in this case can be cancelled immediately and infinite costs are assigned to the whole branch.

\paragraph{Reuse of similar subconfigurations:}
Especially for scenes where objects do not physically interact a lot when being manipulated, similar configurations may appear in several nodes of the tree with respect to object instances and poses. In this case, the subconfiguration of these nodes behave similarly and can be replaced with one another. Therefore, after one of these similar configurations has been processed, the results are directly projected into the other ones without running simulation on those. Peshkin and Sanderson~\cite{Peshkin1987} already described this policy as beneficial.

\subsection{Experiments}\label{experiments}
In order to evaluate the presented method with respect to real-life usage, it is important to note that no ground truth exists other than human intuition which the results can be compared with. The only way of providing ground truth manipulation sequences is human assessment with the human built-in mental simulation capabilities being prone to abstracting and simplifying complex dynamics just as a physics engine. Hence, the following results have to be judged using common-sense intuition since the author is not aware of any comparable approaches presented so far which would provide a baseline dataset.

\subsubsection{Application scenarios}
The experimental evaluation of the presented method is performed on the two scenarios described in the introduction with typical scenes shown in Figure~\ref{fig:intro_scenarios}.

Figure~\ref{fig:histograms_parcelrobot} shows a typical \emph{logistics} scenario with Scene 2 being more challenging because some items provide support to each other. This increases the likelihood of moving a passive object whenever attempting a manipulation action.

In the \emph{supermarket} scenario in Figure~\ref{fig:histograms_pr2}, the tall shelf equipped with cans, rolling away if dropped, provides a hostile environment to any damage-prone items. In such a retail environment, items additionally often tip over like in Scene 4 when customers pull out goods from the back.

When a traditional, collision avoidance-based motion planning algorithm is confronted with this kind of scenarios, it is not very likely to find any solution to clear the scene completely because of objects touching and blocking each other. In all these scenes, touching object configurations can be resolved only heuristically when using classical methods. In contrary, the presented task-level planning method allows for prior-free resolution with less difficulties in clearing all objects than, for example, the approaches presented in \cite{Stoyanov2016} and \cite{Winkler2016} operating on the same scenarios. Both of their evaluations report enhanced complexity in planning and execution for obstructions. Instead, the presented approach avoids these pitfalls by initially selecting a feasible high-level manipulation order for which the motion planning complexity itself is reduced significantly.

\subsubsection{Cost-benefit comparison with existing approaches}
\pgfplotstableread{scene1_frequencies_graphics.csv}\parcelrobotsceneonefrequencies
\pgfplotstableread{scene2_frequencies_graphics.csv}\parcelrobotscenetwofrequencies

\begin{figure*}[btp]
  \centering
  \begin{subfigure}{\textwidth}
  \centering
  \begin{minipage}[r]{.45\textwidth}
    \pgfplotsset{every non boxed y axis/.append style={y axis line style=-}} 
    \resizebox{\linewidth}{!}{
    \begin{tikzpicture}[font=\footnotesize]
    \begin{axis}[
          xbar,
          bar width=0.03\textwidth,
	      xmajorgrids=true,
          xlabel={frequency},
          xmin=0,
          xmax=65,
          xtick={0,10,...,60},
          ytick=data,
          ytick style={draw=none},
          y dir=reverse,
          yticklabels from table={\parcelrobotsceneonefrequencies}{Ranking},
          ylabel={first-ranked sequence $\pi$},
          axis x line=bottom,
          axis y line=left,
	      ylabel near ticks,
          enlarge y limits=0.07,
	      width=0.8\linewidth,
          height=0.22\textheight,
          tick label style={font=\sffamily},
        ]
        \addplot[draw=none, fill=blue!30!white] table[y expr=\coordindex, x=Frequency] {\parcelrobotsceneonefrequencies};
    \end{axis}
    \node at (-2.25,0.53\linewidth) {\footnotesize{$\tau_{w}~~~\tau$}};
    \end{tikzpicture}
    }
  \end{minipage}
  \begin{minipage}[l]{.53\textwidth}
    \includegraphics[width=\textwidth]{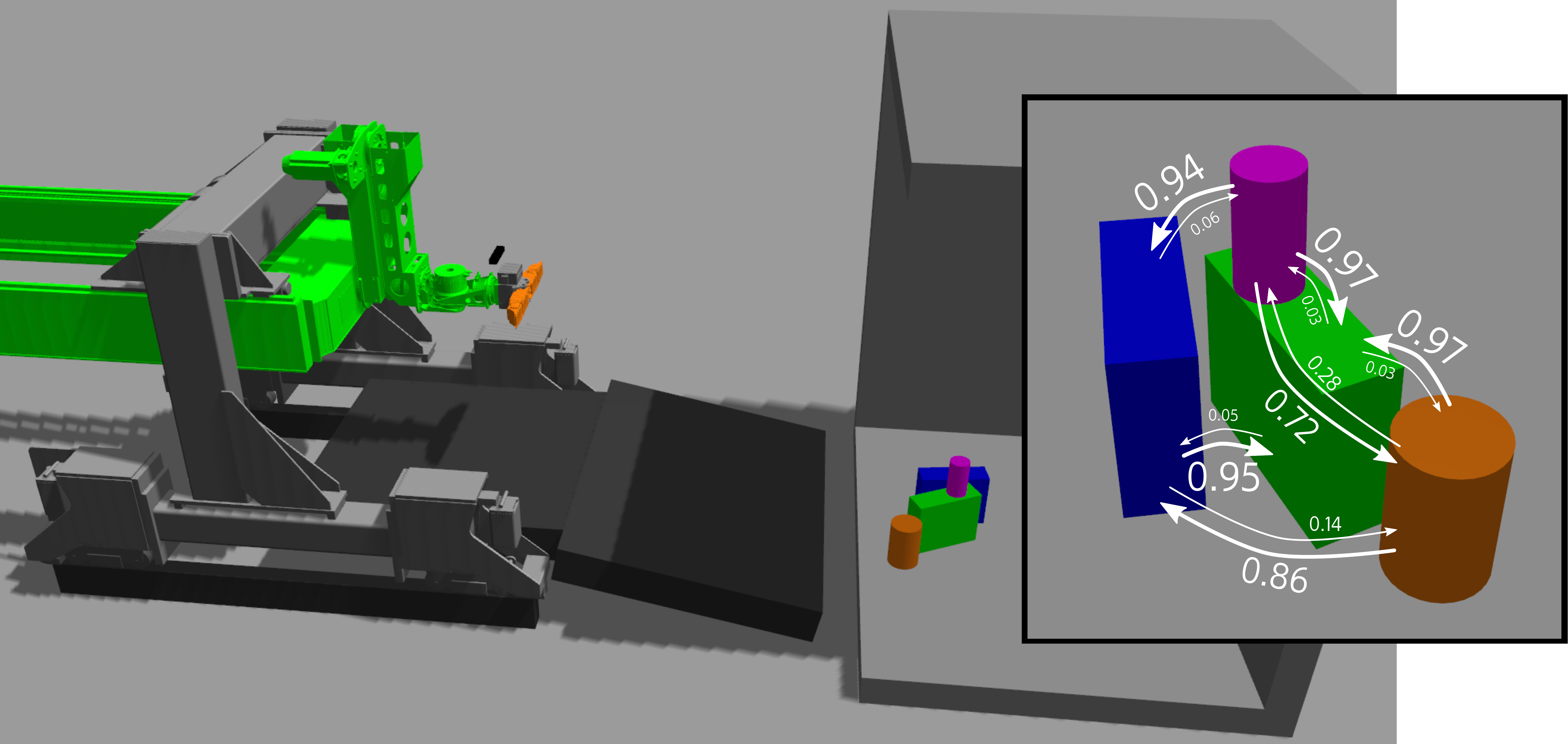} 
  \end{minipage}

  \caption{Scene 1}
  \label{fig:histograms_parcelrobot_scene1}
  \end{subfigure}

  \begin{subfigure}{\textwidth}
  \centering
  \begin{minipage}[r]{.45\textwidth}
    \pgfplotsset{every non boxed y axis/.append style={y axis line style=-}} 
    \resizebox{\linewidth}{!}{
    \begin{tikzpicture}[font=\footnotesize]
    \begin{axis}[
          xbar,
          bar width=0.03\textwidth,
	      xmajorgrids=true,
          xlabel={frequency},
          xmin=0,
          xmax=65,
          xtick={0,10,...,60},
          ytick=data,
          ytick style={draw=none},
          y dir=reverse,
          yticklabels from table={\parcelrobotscenetwofrequencies}{Ranking},
          ylabel={first-ranked sequence $\pi$},
          axis x line=bottom,
          axis y line=left,
	      ylabel near ticks,
          enlarge y limits=0.06,
	      width=0.8\linewidth,
          height=0.25\textheight,
          tick label style={font=\sffamily},
        ]
        \addplot[draw=none, fill=blue!30!white] table[y expr=\coordindex, x=Frequency] {\parcelrobotscenetwofrequencies};
    \end{axis}
    \node at (-2.25,0.63\linewidth) {\footnotesize{$\tau_{w}~~~\tau$}};
    \end{tikzpicture}
    }
  \end{minipage}
  \begin{minipage}[l]{.53\textwidth}
    \includegraphics[width=\textwidth]{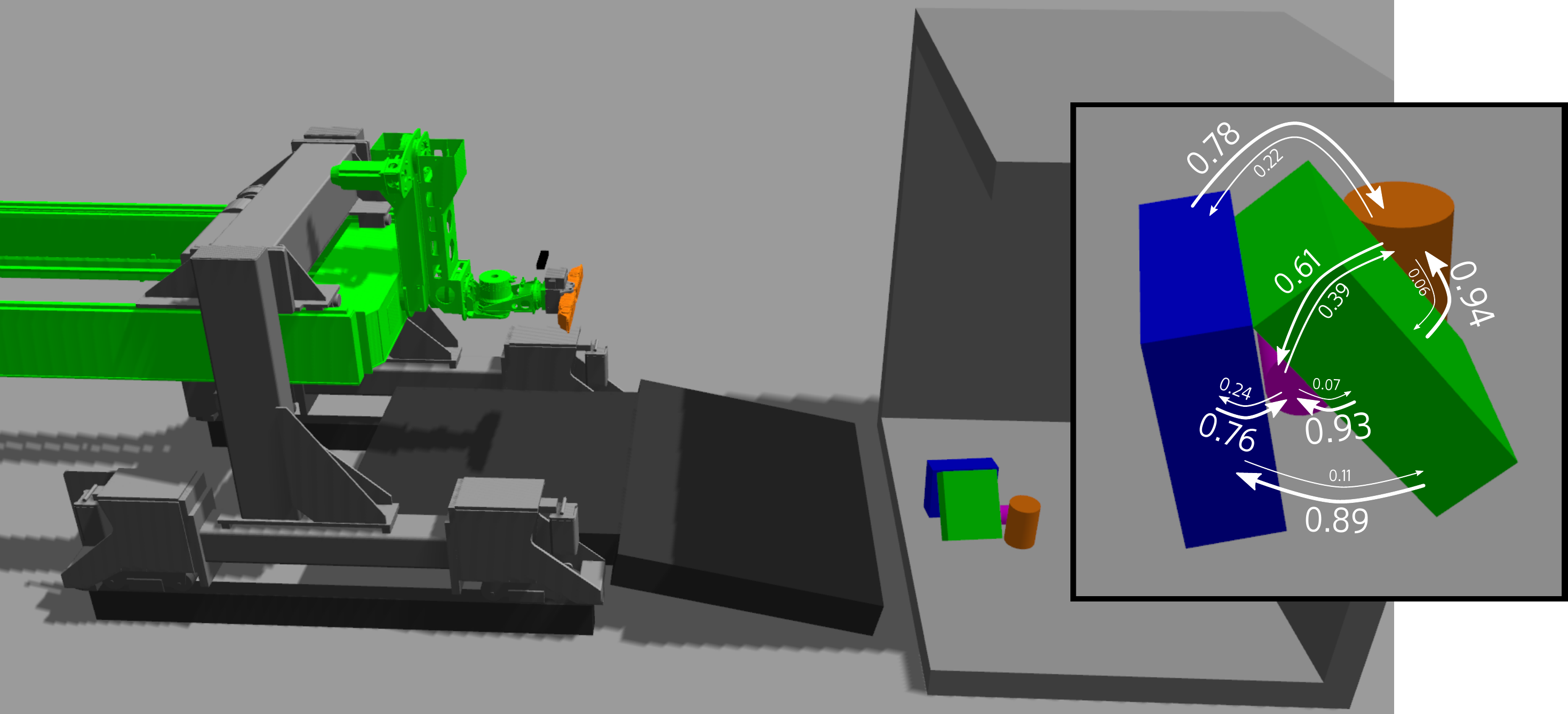} 
  \end{minipage}

  \caption{Scene 2}
  \label{fig:histograms_parcelrobot_scene2}
  \end{subfigure}

\caption{\emph{Logistics scenario}: frequencies of first-ranked sequences $\pi$ (center), $\tau_{w}$ and $\tau$ with respect to the first-ranked sequence (left) and preference weights $w_{ij}$ for pairwise preferences (right) from 100 planning repetitions per scene}
\label{fig:histograms_parcelrobot}
\end{figure*}
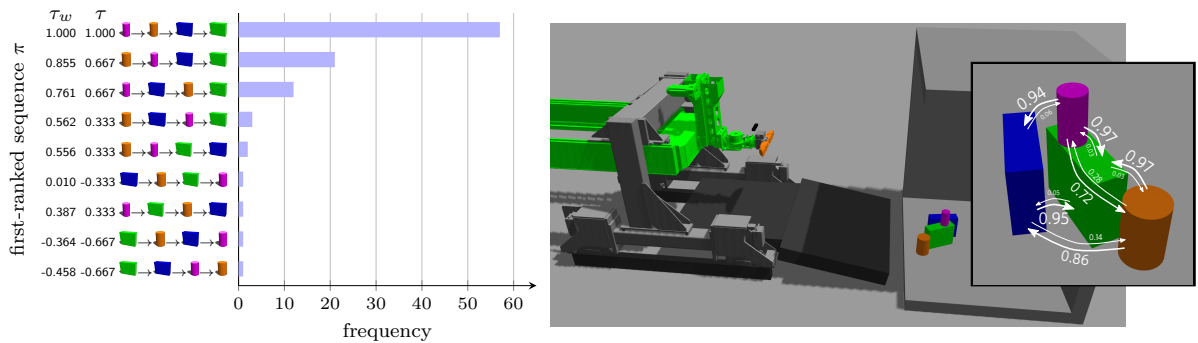
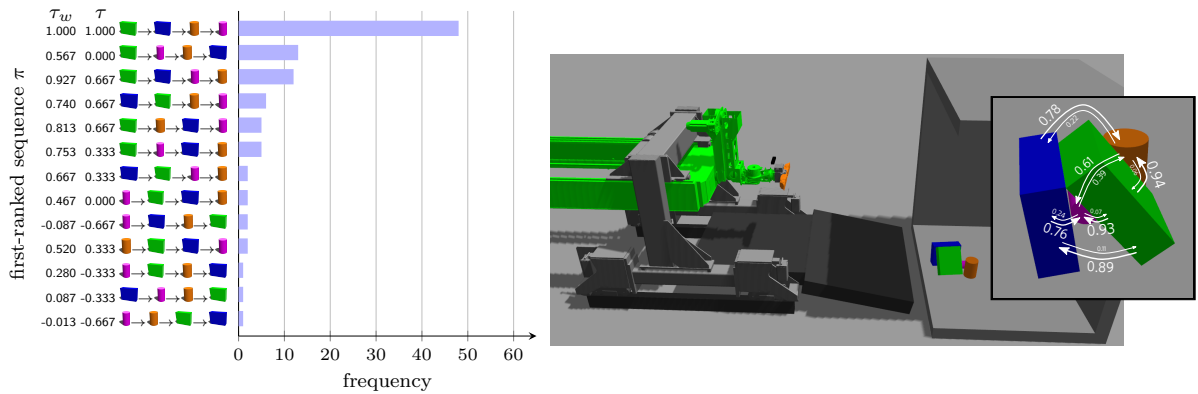

\pgfplotstableread{scene3_frequencies_graphics.csv}\prtwoscenethreefrequencies
\pgfplotstableread{scene4_frequencies_graphics.csv}\prtwoscenefourfrequencies

\begin{figure*}[btp]
  \centering
  \begin{subfigure}{\textwidth}
  \centering
  \begin{minipage}[r]{.45\textwidth}
    \pgfplotsset{every non boxed y axis/.append style={y axis line style=-}} 
    \resizebox{\linewidth}{!}{
    \begin{tikzpicture}[font=\footnotesize]
    \begin{axis}[
          xbar,
          bar width=0.03\textwidth,
	      xmajorgrids=true,
          xlabel={frequency},
          xmin=0,
          xmax=65,
          xtick={0,10,...,60},
          ytick=data,
          ytick style={draw=none},
          y dir=reverse,
          yticklabels from table={\prtwoscenethreefrequencies}{Ranking},
          ylabel={first-ranked sequence $\pi$},
          axis x line=bottom,
          axis y line=left,
	      ylabel near ticks,
          enlarge y limits=0.04,
	      width=0.8\linewidth,
          height=0.27\textheight,
          tick label style={font=\sffamily},
        ]
        \addplot[draw=none, fill=blue!30!white] table[y expr=\coordindex, x=Frequency] {\prtwoscenethreefrequencies};
    \end{axis}
    \node at (-2.22,0.705\linewidth) {\footnotesize{$\tau_{w}~~~\tau$}};
    \end{tikzpicture}
    }
  \end{minipage}
  \begin{minipage}[l]{.53\textwidth}
    \includegraphics[width=\textwidth]{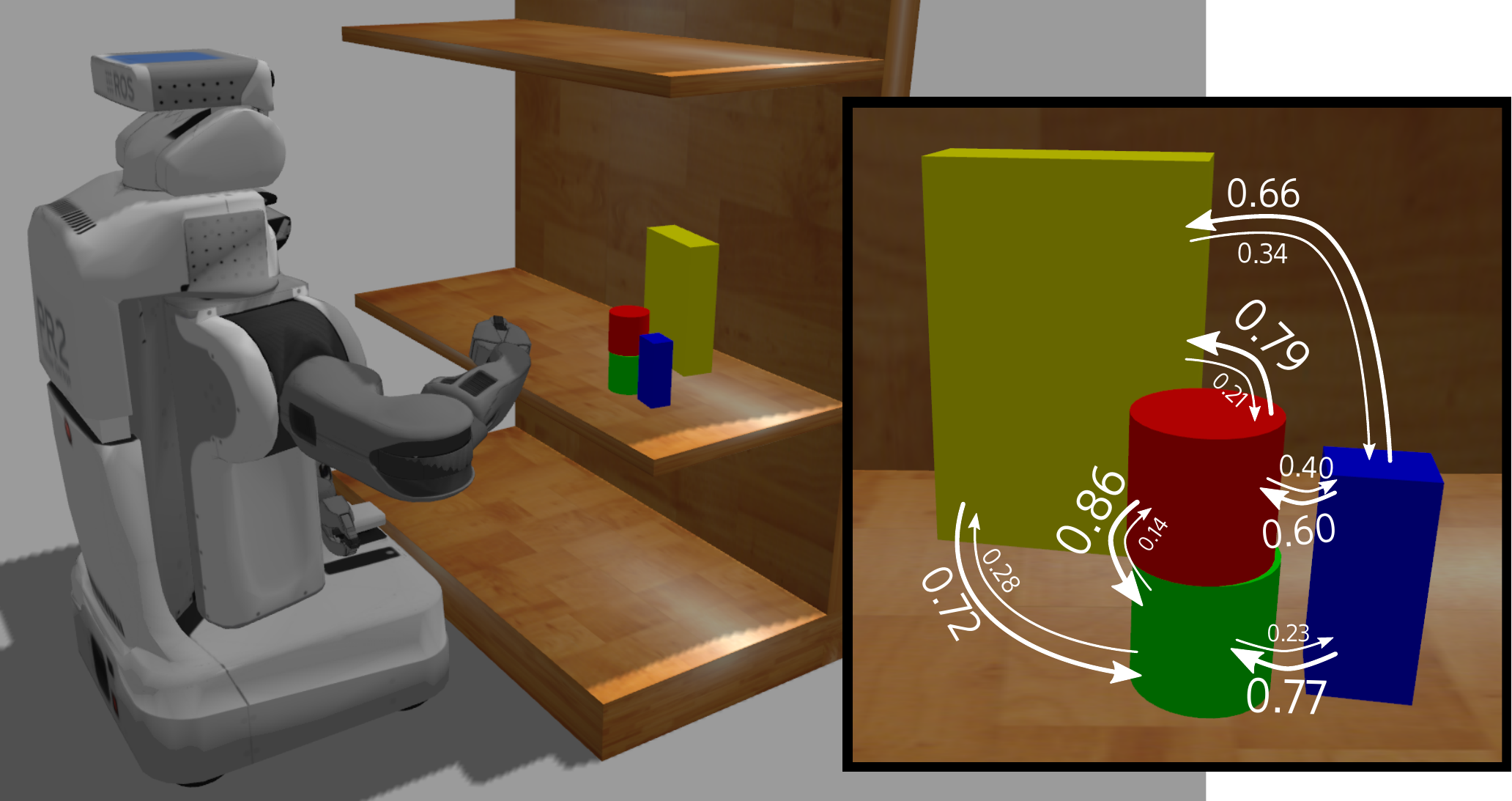} 
  \end{minipage}
  \caption{Scene 3}
  \label{fig:histograms_pr2_scene3}
  \end{subfigure}

  \begin{subfigure}{\textwidth}
  \centering
  \begin{minipage}[r]{.45\textwidth}
    \pgfplotsset{every non boxed y axis/.append style={y axis line style=-}} 
    \resizebox{\linewidth}{!}{
    \begin{tikzpicture}[font=\footnotesize]
    \begin{axis}[
          xbar,
          bar width=0.03\textwidth,
	      xmajorgrids=true,
          xlabel={frequency},
          xmin=0,
          xmax=65,
          xtick={0,10,...,60},
          ytick=data,
          ytick style={draw=none},
          y dir=reverse,
          yticklabels from table={\prtwoscenefourfrequencies}{Ranking},
          ylabel={first-ranked sequence $\pi$},
          axis x line=bottom,
          axis y line=left,
	      ylabel near ticks,
          enlarge y limits=0.04,
	      width=0.8\linewidth,
          height=0.28\textheight,
          tick label style={font=\sffamily},
        ]
        \addplot[draw=none, fill=blue!30!white] table[y expr=\coordindex, x=Frequency] {\prtwoscenefourfrequencies};
    \end{axis}
    \node at (-2.22,0.735\linewidth) {\footnotesize{$\tau_{w}~~~\tau$}};
    \end{tikzpicture}
    }
  \end{minipage}
  \begin{minipage}[l]{.53\textwidth}
    \includegraphics[width=\textwidth]{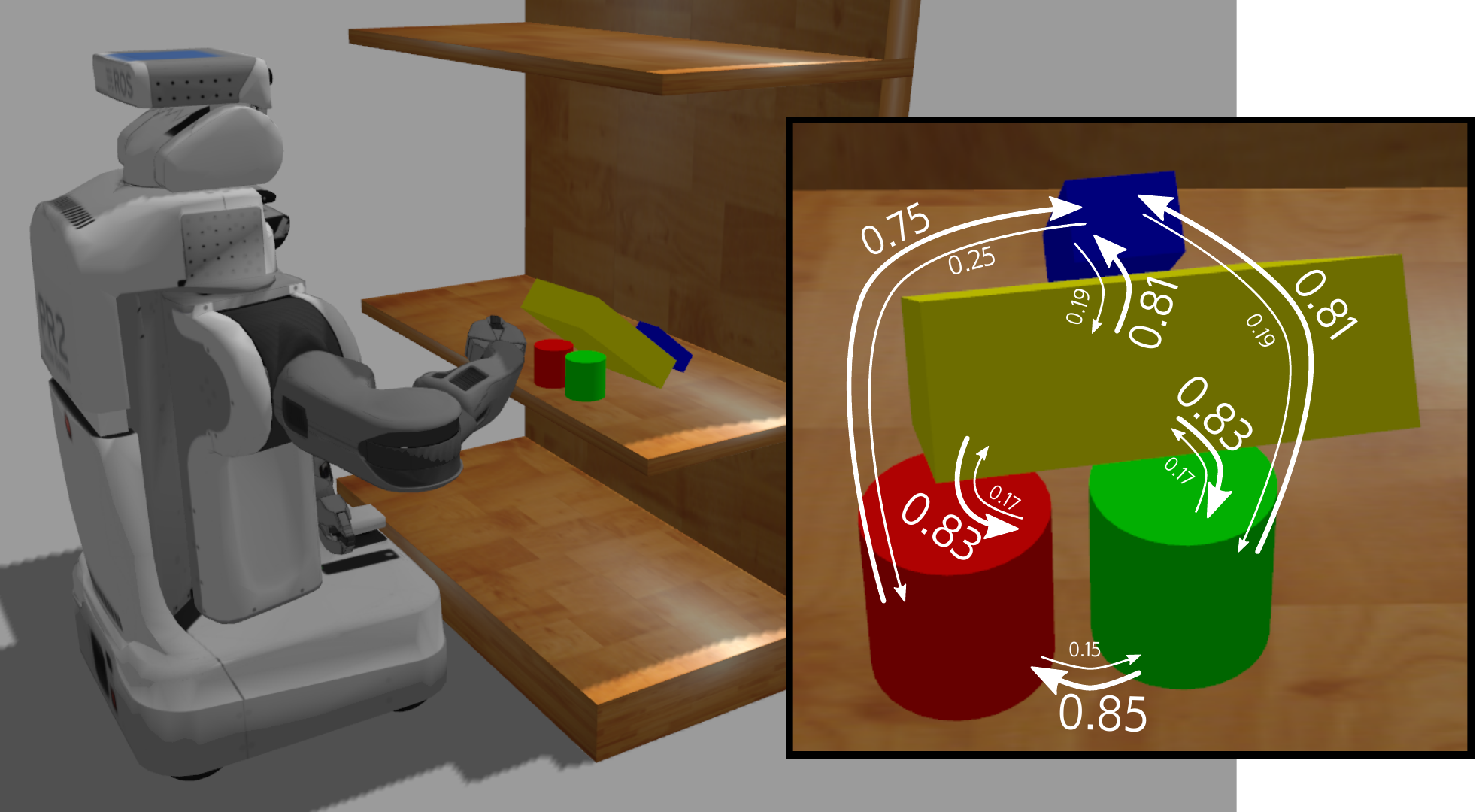} 
  \end{minipage}
  \caption{Scene 4}
  \label{fig:histograms_pr2_scene4}
  \end{subfigure}

\caption{\emph{Supermarket scenario}: frequencies of first-ranked sequences $\pi$ (center), $\tau_{w}$ and $\tau$ with respect to the first-ranked sequence (left) and preference weights $w_{ij}$ for pairwise preferences (right) from 100 planning repetitions per scene}
\label{fig:histograms_pr2}
\end{figure*}

The main requirements for the proposed approach when it is used for high-level planning as an add-on to classical motion planning is the setup of a simulation environment including object and robot models along with the respective controllers as described in Section~\ref{prerequisites}. Although this means a one-time setup per application scenario whose effort strongly decreases over multiple scenarios with the reuse of software components, this required effort may not be neglected.

For the presented scenarios, the setup of a simulation environment as in the Supplementary Video had been demanded for in the RobLog research project's requirements. The respective work package comprised of a dynamics anticipation approach including the from-scratch setup of all prerequisites described above and was accounted for with a workload of 23 person months. However, since this number includes a high amount of research which stretches beyond what is required per scenario utilizing a working software solution, the actual anticipated workload in a productive environment is a lot lower. 
Our experience showed that, after establishing and optimizing the presented method in the logistics scenario, setting up the supermarket scenario subsequently required a workload of only about 3-4 person months.
This might still seem a lot if traditional approaches possibly are able to deal with the same sort of challenges and scenarios, however, the prior-free damage avoidance of the proposed approach is a valuable asset especially in use cases dealing with expensive or high-throughput goods.

In terms of computational effort required for manipulation strategy generation and deployment, typical cycle times of the same research-based scenario circulate in the region of 3.5-\SI{5}{minutes} per item \cite[p.~11]{Stoyanov2016}. In this context, the classification times for manipulation strategies in a sub-second magnitude do not carry any weight during productive use. Additional computational effort of the proposed approach accrues mostly in the manipulation strategy optimization loop which can be run offline, during load-free robot runtime. Therefore, the proposed method imposes no additional requirements in terms of computing hardware compared to traditional methods.

\subsubsection{Mental simulation}\label{evaluation_planning}
Mental simulation (Algorithm~\ref{alg:planning_anticipation}) was run 100 times for each of the two exemplary scenes per domain.
The histograms in Figures~\ref{fig:histograms_parcelrobot} and \ref{fig:histograms_pr2} show how many times a particular manipulation sequence $\pi$ was chosen as the most optimal one. A clear preference for a specific sequence is visible for all of  scenes, however, in Scene 3, the distance in frequency between the first and second-ranked sequence is less distinct. This stems from the fact that, in the given object configuration, it does not seem to make a big difference regarding possible damage whether, for instance, the \obj{3}{red} or the \obj{3}{blue} is manipulated first. In the other scenes, however, the objects bear a higher spatial dependency on each other and it is more likely to distort the setup when manipulating them in any other than the top-ranked order.
Summarized, these results show that mental simulation within the presented method is effective per se on different object and robot configurations.

Regarding the efficiency of the proposed approach, several generic means of optimization have been explained in Section~\ref{efficiency}. With respect to these, the following concrete criteria have been evaluated in the numerical results shown in Table~\ref{table:statistics}:
\begin{itemize}
	\item \emph{mean first-ranked costs per node}: mean over all planning repetitions of the mean costs per object imposed on the manipulation sequence selected by Algorithm~\ref{alg:planning_anticipation}
	\item \emph{mean second-ranked costs per node}: mean over all planning repetitions of the mean costs per object imposed on the second-ranked manipulation sequence
	\item \emph{pruned tree nodes}: percentage of nodes pruned from the tree, resulting in a similar reduction in runtime, and composed of the following sub-criteria:
	\begin{itemize}
		\item \emph{known subtree}: a node shares its object configuration and poses with a previously processed node, thus the costs were copied without re-si\-mu\-la\-ting
		\item \emph{costs exceed existing sequence}: a solution is existing already which has lower costs than the costs accumulated so far in this branch
		\item \emph{active object moved}: the active object was pushed away during approach, ending up unreachable
		\item \emph{object out of workspace}: an object fell out of the workspace (container/shelf), causing maximum damage and ending up unreachable for the robot
		\item \emph{planning failure}: no motion plan was found for the active object, e.g.\ because it was pushed away too far in a parent node
	\end{itemize}
	\item \emph{nodes with significant movement}: percentage of nodes which were not pruned from the tree and reported significant costs above a manually defined threshold 
\end{itemize}

\begin{table*}[b!]
\centering
\scriptsize
\setlength{\tabcolsep}{.4em} 

\begin{tabular}{lccccc}
\textbf{\normalsize{Scene}} & \textbf{\normalsize{1}} & \textbf{\normalsize{2}} & \textbf{\normalsize{3}} & \textbf{\normalsize{4}} & \textbf{\normalsize{Total}} \\
\rule{0pt}{3ex}\textbf{costs per node} & & & & & \\
 - first-ranked sequence       &  1.008 \tiny{$\pm$0.008} &  1.208 \tiny{$\pm$0.074} &  1.349 \tiny{$\pm$0.359} &  1.975 \tiny{$\pm$0.321} &  1.382 \tiny{$\pm$0.435} \\
 - second-ranked sequence      &  1.337 \tiny{$\pm$0.868} &  1.429 \tiny{$\pm$0.326} &  1.700 \tiny{$\pm$1.728} &  3.764 \tiny{$\pm$7.393} &  2.051 \tiny{$\pm$3.878} \\
\rule{0pt}{3ex}\textbf{pruned tree nodes}             & \meanplotvalue{34.8}{0} & \meanplotvalue{50.4}{0} & \meanplotvalue{58.3}{0} & \meanplotvalue{61.9}{0} & \meanplotvalue{51.3}{0} \\
 - known subtree                       & \meanplotvalue{7.5}{4.8} & \meanplotvalue{2.6}{2.5} & \meanplotvalue{1.2}{2.2} & \meanplotvalue{0.0}{0.0} & \meanplotvalue{2.8}{4.1} \\
 - costs exceed existing seq.          & \meanplotvalue{16.5}{8.3} & \meanplotvalue{18.3}{5.3} & \meanplotvalue{19.3}{14.7} & \meanplotvalue{41.0}{18.1} & \meanplotvalue{23.8}{16.0} \\
 - active object moved                 & \meanplotvalue{5.8}{5.6} & \meanplotvalue{25.2}{13.5} & \meanplotvalue{4.2}{11.6} & \meanplotvalue{8.1}{9.2} & \meanplotvalue{10.9}{13.2} \\
 - object out of workspace             &  \meanplotvalue{0.0}{0.0} & \meanplotvalue{0.0}{0.0} & \meanplotvalue{31.9}{15.7} & \meanplotvalue{7.0}{9.6} & \meanplotvalue{9.6}{15.9} \\
 - planning failure                    &  \meanplotvalue{5.0}{9.6} & \meanplotvalue{4.3}{8.4} & \meanplotvalue{1.7}{3.1} & \meanplotvalue{5.8}{10.9} & \meanplotvalue{4.2}{8.6} \\
nodes with significant & \meanplotvalue{13.8}{3.0} & \meanplotvalue{11.9}{4.9} & \meanplotvalue{14.8}{4.3} & \meanplotvalue{18.9}{5.1} & \meanplotvalue{14.9}{5.0} \\
movement (costs $>2.0$)    &          &                \\
\end{tabular}
\caption{Mental simulation numerical results: means and standard deviations of 100 planning repetitions}
\label{table:statistics}
\end{table*}

Table~\ref{table:statistics} shows that pruning the tree is generally effective, eliminating a mean of \SI{51.3}{\%} of nodes from the tree during the planning process. Since the planning time behaves linear with respect to the number of nodes to cover, this means that one planning run can be completed in average within less than half the time compared to using an unpruned tree.

In Scene 2, a high number of nodes had to be skipped because the active object moved during approach. This is caused by the \obj{2}{blue} and \obj{2}{green} hiding the \obj{2}{orange} and \obj{2}{pink} from the robot's view. Therefore, the planner made the passive \obj{2}{blue} and/or \obj{2}{green} push the currently active \obj{2}{orange} or \obj{2}{pink} away.
Regarding Scene 3, an object has moved out of the workspace in \SI{31.9}{\%} of nodes. This was caused by the round \obj{3}{red} which, when the \obj{3}{green} is extracted from underneath, often rolls away and falls off the shelf.

However, such exceptional cases can be caught and a feasible sequence can be found nevertheless. In total, although in many cases the computed manipulation trajectories did not cause significant disturbance in the scene, in \SI{14.9}{\%} of nodes significant movement was detected which, in real-world execution, may have led to non-negligible damage. Within tree search, however, these nodes have generally been avoided as shown in the first-ranked per-node costs. These are low enough to ensure the provided manipulation sequences are damage-minimizing.

\subsubsection{Subconscious preference patterns}\label{results_patterns}
When looking at the initial scene configurations, most of the selected sequences in the left column of Figures~\ref{fig:histograms_parcelrobot} and \ref{fig:histograms_pr2} intuitively make sense, however, certain preferences in a single run may look counter-intuitive.
One example for this is the \obj{3}{blue} which surprisingly often turns up as a non-preferred object. This particular case can be explained with the fact that the robot has to extend its manipulator quite far to grasp this object, hence pushing other objects off the shelf if they are still present.

Nevertheless, such preference patterns may evolve from many repetitions on the same scene unexpectedly, depending on the object constellation, robot location and other factors.
The images on the right of Figures~\ref{fig:histograms_parcelrobot} and \ref{fig:histograms_pr2} show the respective preference weights $w_{ij}$ for the pairwise preferences of each object over all others.

The results show that, consistent with human intuition, generally objects whose centroid is higher than the one of other objects or which are resting on other objects are preferred. Other than that, objects obstructing a direct manipulation trajectory between gripper and objects in the back will be manipulated first in the most cases.
In Scenes 3 and 4, where the PR2 was used, there is an additional clear preference of objects which are situated at the right of other objects. This is not surprising since the right arm of the PR2 was used for manipulation, hence granting a bigger workspace to the right side as can be seen in Fig.~\ref{fig:pr2_workspace}.

Several other preferences, like between the \obj{3}{red} and \obj{3}{blue} in Scene 3, however, are less expressive in the way that the preference weights lie close to 0.5. This implies a minimum of information content because each direction of the preference is equally likely. This phenomenon is caused by the fact that a large number of viable grasping configurations can be generated easily for these objects in this configuration given the robot workspace and dexterity.

Summarized, even though it is very difficult to denote preferences patterns as encountered in the experiments manually in an extensive way, they can be discovered by the presented approach.

\subsubsection{Preference-weighted Kendall's tau $\tau_{w}$}
The next subsection shows the behavior of the classifier for different settings with the help of the exposed preference patterns during classifier training and prediction. The preferences are embedded into the $\tau_{w}$ measure which gives very distinct results for sequences with identical $\tau$. This results in an improved utility of the continuous $\tau_{w}$ as opposed to $\tau$ which gives coarse discrete values, see Figs.~\ref{fig:histograms_parcelrobot} and \ref{fig:histograms_pr2}.

For instance, the \obj{1}{green} in the sequence \seq{1}{pink}{green}{orange}{blue} ($\tau_{w}=0.387$) in Fig.~\ref{fig:histograms_parcelrobot_scene1} which gets moved first is obviously prone to accidently moving \obj{1}{orange} or \obj{1}{blue}. However, although they all produce the same $\tau$ of $0.333$, in \seq{1}{orange}{blue}{pink}{green} ($\tau_{w}=0.562$) and \seq{1}{orange}{pink}{green}{blue} ($\tau_{w}=0.556$) it is less likely that a passive object is pushed away. Hence, in contrary to using $\tau$ as a suitability measure, these sequences are better distinguishable using their $\tau_{w}$ measure which takes the pairwise preferences into account. 

\subsubsection{Ranking by pairwise comparison}\label{results_rpc}
In order to show the performance of the overall classifier, a series of 80 training scenes was generated with five variants each which carry stochastic object pose noise. For each scene variant 100 samples of manipulation sequences were collected by running the mental simulation method on them in order to account for noise in the simulation itself, like described in Section~\ref{evaluation_planning}. In total, this amounts to $80 \cdot 5 \cdot 100 = 40000$ samples, each containing a feature vector and the manipulation sequence determined by the planning algorithm. 
With the used proof-of-concept implementation, data collection takes around \SI{94}{minutes} per 100 samples using four parallel containers on a \SI{3.4}{GHz} desktop CPU with \SI{16}{GB} RAM while running with a real-time factor of \SI{1.75 \pm 0.80}{}.

\begin{figure*}[p]
  \begin{flushright}
    \pgfplotstabletranspose\classifiertraining{classifier_training.csv}
    \pgfplotstablegetcolsof{\classifiertraining}
    \pgfmathsetmacro{\numrowsclassifiertraining}{\pgfplotsretval-1}
    \pgfplotstableread{classifier_training_results.csv}\classifiertrainingresults
    \pgfplotstablegetrowsof{\classifiertrainingresults}
    \pgfmathsetmacro{\numrowsclassifiertrainingresults}{\pgfplotsretval-1}
    \begin{tikzpicture}[font=\footnotesize]
        \begin{axis}[
              xmin=-1,
              xmax=1,
              xtick={-1.0,-0.8,...,1.0},
              xtick style={draw opacity=0.5},
              xtick pos=left,
              x tick label style={/pgf/number format/.cd,fixed,fixed zerofill,precision=1,/tikz/.cd,font=\scriptsize},
              every node near coord/.style={/pgf/number format/.cd,fixed,fixed zerofill,precision=3,/tikz/.cd,font=\tiny\sffamily},
              xlabel={$\tau{_w}$},
              ymin=-0.5,
              ymax=\numrowsclassifiertraining+0.8, 
              ytick style={draw=none},
              y dir=reverse,
              ytick={1,2,...,\numrowsclassifiertraining},
              yticklabels from table={\classifiertrainingresults}{Description},
              width=0.7\textwidth,
              height=0.03\textheight*\numrowsclassifiertraining+0.1\textheight,
              axis line style={draw opacity=0},
              cycle list={{blue}},
              boxplot/draw direction=x,
              every boxplot/.style={color=black!40!blue,mark=x,every mark/.append style={mark size=0.8pt},fill=blue!10!white}
            ]
            \foreach \i in {1,...,\numrowsclassifiertraining} {
                \addplot+ [boxplot] table [y index=\i] {\classifiertraining};
            }
        \end{axis}
    \end{tikzpicture}
    \end{flushright}
    \caption{Results for manipulation strategy generation: $\tau_{w}$ for different numbers of scenes and samples per scene, binary/soft voting, with/without preference weights, each scene sampled five times with stochastic object pose noise.}
    \floatfoot{Median, lower and upper quartile ($Q_1$/$Q_3$) over 100 classifier training repetitions. Lower and upper fences were calculated using $Q_1-1.5 \cdot (Q_3-Q_1)$ and $Q_3+1.5 \cdot (Q_3-Q_1)$, respectively.}
    \label{fig:results_learning}
\end{figure*}
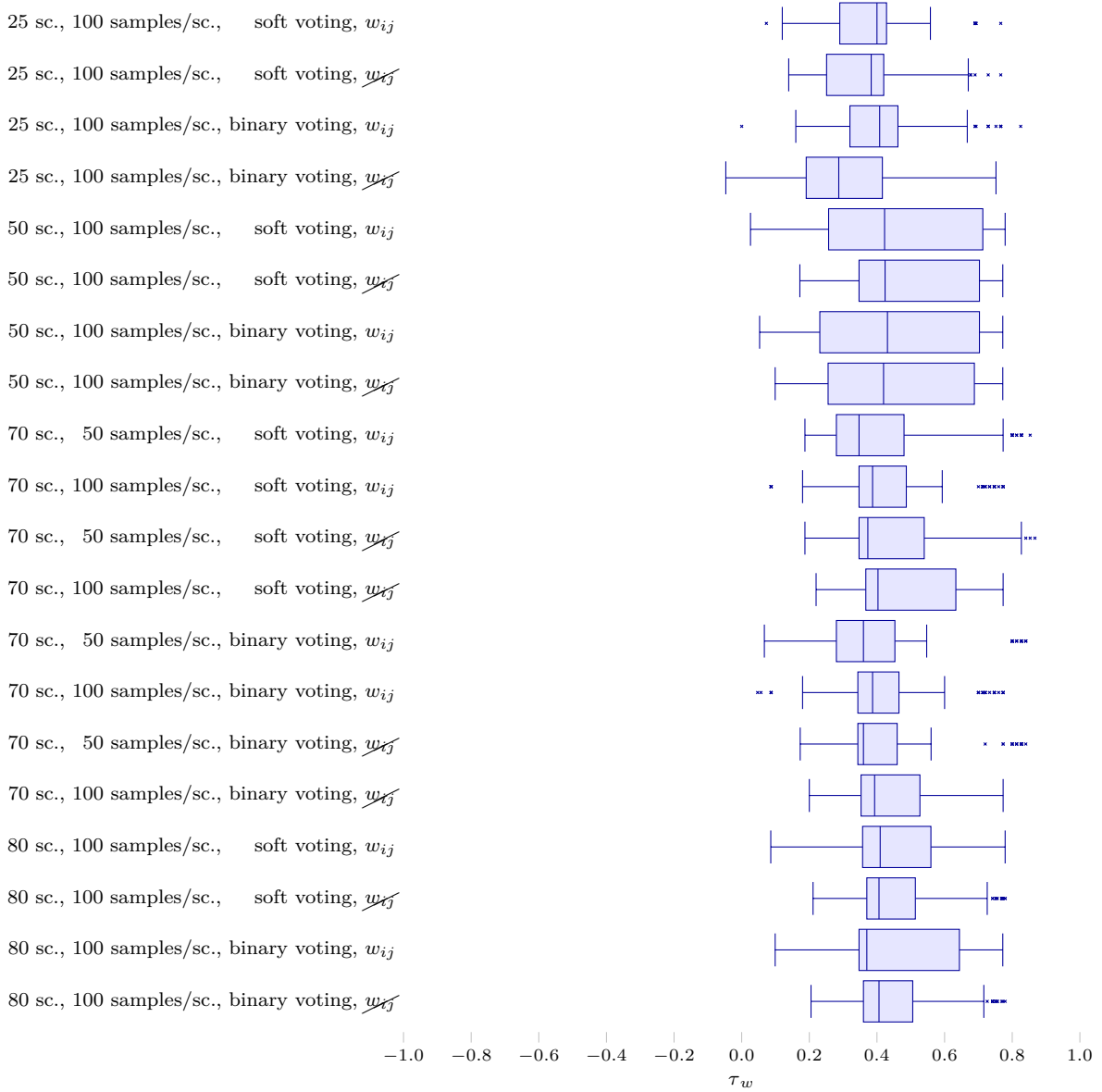

The generated data set was fed into classifier training in different configurations of the voting method and with/without integrating preference weights. Every run, the classifier was trained with a random 2:1 training/testing split on the respectively indicated number of scenes and evaluated on the testing split afterwards. 100-fold repetition of this training/testing loop yielded the distributions of results shown in Fig.~\ref{fig:results_learning}.
With respect to classifier learning time, an average of \SI{27.7 \pm 11.7}{s} has been achieved with the number of scenes rising from 10 to 80. The learning time rose from 10 scenes taking less than one second to about 45 seconds for 80 scenes.

As shown in more detail in the next subsection, the results generally improve with the number of scenes. However, Fig.~\ref{fig:results_learning} shows a difference in performance for many of the considered configurations, especially regarding the variance of the distributions. Generally, it is more desirable to obtain low-variance distributions rather than high-variance ones with a similar median because the former mean higher precision and hence show to be more resilient against noise.
Nevertheless, any bias in the distribution harms the precision, but positive bias w.r.t.\ $\tau_{w}$ means an increase in accuracy. Therefore, gauging accuracy versus precision of these results is important for the overall assessment.

Concretely, even though the obtained results show no significant difference between configurations with or without injected preference weights, the weights themselves cohere with intuitive behavior. Hence this enables a deeper understanding of the classification results by providing the $\tau_{w}$ metric which allows for measuring this coherency. Without this $\tau_{w}$ metric which incorporates the preference weights, we would not be able to see that the learned manipulation strategy adheres to the subconscious preference patterns discovered using the proposed method.

As for the used voting method, the results in Fig.~\ref{fig:results_learning} show a slight improvement of soft over binary voting in terms of mean and variance/positive bias. Therefore the author recommends to use soft voting when applying the method.
Some improvement is shown also for the number of samples per scene, although this is not surprising since a higher number of samples per individual scene statistically rules out a larger amount of noise. 

Summarized, for the following experiment which shows the effectiveness of the explained self-supervised optimization cycle, it was therefore chosen to use a configuration of \emph{100 samples per scene, soft voting and no injection of preference weights}.

\subsubsection{Self-supervised free-time manipulation strategy optimization}
With the classifier configuration determined in the last experiment, the whole manipulation strategy optimization cycle was run in an iterative fashion like in Alg.~\ref{alg:optimization}. It was initialized with a set of 10 training scenes taken from the total training set as described above. After establishing an initial classifier, one more training scene was generated, a damage-avoiding manipulation sequence planned and the classifier retrained. This was repeated up to a maximum of 80 scenes. Note that, for evaluation purposes, classifier training was repeated 10 times each with a randomized 2:1 training/testing split on the scenes so as to generate a more noise-resilient evaluation.

\begin{figure}[btp]
    \pgfplotstabletranspose\classifierimprovement{classifier_improvement_training.csv}
    \pgfplotstableread{classifier_improvement_training_results.csv}\classifierimprovementresults
    \pgfplotstableread{classifier_improvement_training_linearized.csv}\classifierimprovementlinearized
    \pgfplotstablegetrowsof{\classifierimprovementresults}
    \pgfmathsetmacro{\numrowsclassifierimprovementresults}{\pgfplotsretval+10}
    \begin{tikzpicture}[font=\footnotesize]
        \begin{axis}[
              ymin=-0.2,
              ymax=1,
              ytick={-0.2,-0.0,...,1.0},
              ytick style={draw opacity=0.5},
              ytick pos=left,
              y tick label style={/pgf/number format/.cd,fixed,fixed zerofill,precision=1,/tikz/.cd,font=\scriptsize},
              every node near coord/.style={/pgf/number format/.cd,fixed,fixed zerofill,precision=3,/tikz/.cd,font=\tiny\sffamily},
              ylabel={$\tau_{w}$},
              ymajorgrids,
              xmin=9,
              xmax=\numrowsclassifierimprovementresults,
              xtick pos=left,
              xtick style={draw opacity=0.5},
              xtick={10,15,...,\numrowsclassifierimprovementresults},
              xlabel={\#scenes},
              width=0.7\textwidth,
              height=0.3\textheight,
              axis line style={draw opacity=0},
              cycle list={{blue}}
            ]
            \addplot+ [smooth, blue!10!white, name path=UQ] table [x=Scenes, y=TauThirdQuartile] {\classifierimprovementresults};
            \addplot+ [thick, black, name path=M, mark=x, mark size=1.3pt, only marks] table [x=Scenes, y=TauMedian] {\classifierimprovementresults};
            \addplot+ [smooth, blue!10!white, name path=LQ] table [x=Scenes, y=TauFirstQuartile] {\classifierimprovementresults};
            \addplot[blue!30!white] fill between [of=UQ and LQ];
            \addplot+ [thick, orange] table [x=x, y=y] {\classifierimprovementlinearized};
        \end{axis}
    \end{tikzpicture}
    \caption{Results for manipulation strategy optimization: $\tau_{w}$ for different numbers of scenes using 100 samples per scene, soft voting, no preference weights, each scene sampled five times with stochastic object pose noise.}
    \floatfoot{Median ($\tilde{\tau_{w}}$), \textcolor{blue!70}{lower/upper quartile envelope} and \textcolor{orange}{least-squares best-fit line} for 10 training runs.}
    \label{fig:results_improvement}
\end{figure}
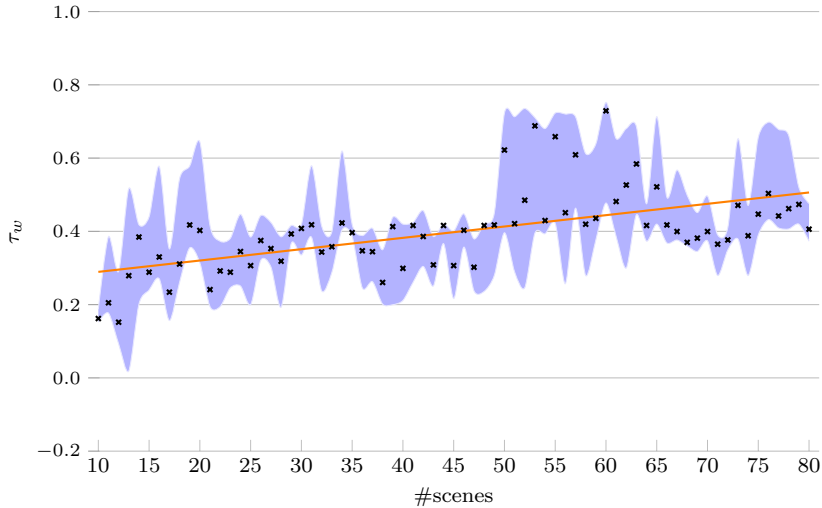

The emerging results are plotted in Fig.~\ref{fig:results_improvement} with the median of $\tau_{w}$ over these 10 classifier training runs per scene, further on named $\tilde{\tau_{w}}$. A \textcolor{orange}{least-squares best-fit line} over $\tilde{\tau_{w}}$ for each scene shows a positive slope, hence generally more training data gives better classification results. However, the \textcolor{blue!70}{lower/upper quartile envelope}, which initially becomes more narrow, widens again after reaching 50 individual training scenes. Eventually, after 63 scenes, $\tilde{\tau_{w}}$ drops again well below the best-fit line which indicates overfitting, but recovers for 75+ scenes.

With respect to the method's overall runtime, Alg.~\ref{alg:optimization} in a worst-case, non-parallelized implementation performs depending on its generate/update loop (Lines~\ref{alg:optimization:loop_begin}--\ref{alg:optimization:loop_end}, also depicted in Fig.~\ref{fig:components}). Taking into account the temporal performance described in Section~\ref{results_rpc}, this loop performs with an average cycle time of around $T=\SI[parse-numbers=false]{(n_{\mathrm{samples}}\cdot60\cdot0.94+27.7)}{s}$ in the worst case, i.e.\ without any parallelization, for a training sample batch size of $n_{\mathrm{samples}}$. For $n_{\mathrm{samples}}=1$, i.e.\ retraining after each generated sample, this amounts to $T_{1} = \SI{84.1}{s}$; for $n_{\mathrm{samples}}=100$ like used in the experiments, this amounts to $T_{100} = \SI{5667.7}{s} \approx \SI{94.5}{min}$. Through parallelization, a theoretical speedup of $S_{dp} = p_d \cdot p_p$ can be achieved where $p_d$ is the number of Docker containers and $p_p$ is the number of mental simulation processes running in parallel.

The given results prove the initial claim that, while generating more and more training scenes and optimizing the manipulation strategy in a self-supervised manner, there is an improvement in classification accuracy with an increasing number of considered unique scenes.
However, it is hard to provide a convergence measure which determines the globally optimal number of scenes after which optimization should be stopped for the scene/object configuration at hand. This strongly depends on the application scenario and the ground truth data for the presented method has been generated by the method itself. That data includes a certain amount of noise propagating from the low-level motion planning and physics simulation towards the final manipulation strategy which cannot completely be ruled out by the proposed approach.
Therefore, for a specific application scenario, it is necessary to define some heuristics to determine when exactly to stop the optimization loop and to proceed towards optimizing different scene/object configurations.

In the presented scenario, the $\tau_{w}$ grow sufficiently on their $[-1,1]$ scale with an increasing number of scenes. Given that $-1.0$ means full discordance (i.e.\ the preferences appear in entirely inverse order) and $1.0$ means full concordance (i.e.\ the preference appear in entirely correct order), the $\tau_{w}$ from the classifier predictions are close to the optimal manipulation sequences as determined by mental simulation. This means that not only are the results improving as desired during the optimization process, but they also reflect realistic human-like behavior concerning the respective scene. Hence, this method is able to increase the cognition and reasoning skills of the robot with respect to manipulation preferences.

\section{Conclusions}\label{conclusions}
The first part of this work presented a mental simulation method to plan manipulation sequences while minimizing potential damage with respect to objects in the scene. This is achieved via anticipating the scene's dynamics during the interaction process. Moreover, several measures allow for a significant improvement in efficiency.
In the second part, a classifier was trained that can predict an optimal manipulation sequence from new, unknown scenes. This happens iteratively within a self-supervised manipulation strategy optimization cycle and allows the robot to continously acquire skills during load-free times. The resulting strategies can be deployed anytime and executed in near-real time.

The presented work merely scratches the surface of what may be possible to infer from physics simulation to create long-term knowledge. However, it was shown that preference patterns humans acquire subconsciously get shaped as well in robotic behavior learned via mental simulation.
Regarding the fidelity of simulation-based planning with respect to the real world, it is a hard problem to design a feature-based physical scene understanding approach based on complex human inferences. It would be necessary to find a set of features which is capable of depicting all inferences humans draw from some input scene using their world knowledge.
However, the featureless manipulation sequence planning approach by itself is very resource-demanding during productive use. Hence, the feature-enabled manipulation strategy learning part of the method allows for reducing execution times to a minimum, outsourcing expensive computation into low-load time slots. Strictly speaking, the overall approach proposed herein does not rationalize any mental simulation capabilities, but minorly softens their expressiveness in favor of reasonable use under real-life conditions.

Vice versa, humans also use certain features as a fallback when scene dynamics are too complex to be anticipated. Hence, the utility of features as a mean of generalization cannot be neglected per se.
Nevertheless, it may be possible to increase the overall performance, explicitely precision, when the feature set is iteratively improved based on experimental classification outcomes. The proposed feature set serves as a first attempt to prove the general feasibility and applicability of this approach, although more investigation is required into adapting the features to the respective use case in productive use. With increased precision, the answer about how to measure converge of the optimization loop may potentially be perfectly obvious.

In any case, in order to avoid permanent retraining of the classifier from scratch, using an online algorithm would improve the temporal performance. With the current setup, retraining the classifier is necessary every time a new training sample has been generated. Although this can happen in parallel to generating the next training sample via mental simulation, updating the classifier in an online fashion removes some computational burden from the approach and allows for faster overall processing, hence shorter time to convergence. Further investigation in terms of speed improvement may prove beneficial for the applicability of the presented method in real-world applications.
Future research potentially may also point into the direction of deep learning which generally does not require the developer to define the feature set prior to classifier design. Using this technique, certainly major effort needs to be invested into designing the classifier pipeline. This exceeds the scope of this initial approach to solve the present complex problem and should be investigated in future work.

Nevertheless, a major advantage of the proposed approach is that, except for the feature set, it is almost parameter-free and hence does not require extensive tuning based on the application scenario.
The genericness of the proposed approach allows for deployment in many different scenarios for which a simulation of the robot and scene dynamics can be provided. Therefore, the method can easily be integrated into existing applications as a mean of high-level task planning including the capability of self-supervised strategy optimization.

\bibliographystyle{spmpsci_unsorted}
\bibliography{bibliography}

\end{document}